\documentclass[reprint,amsmath,amssymb,aps,prx,floatfix]{revtex4-2}

\usepackage{graphicx}
\usepackage{dcolumn}
\usepackage{bm}
\usepackage{hyperref}
\usepackage{soul,color} 
\usepackage{pgfplots}
\usepackage{enumitem}

\DeclareMathOperator{\Tr}{Tr}

\begin{document}

\preprint{APS/123-QED}

\title{Disentangling representations in Restricted Boltzmann Machines without adversaries}

\author{Jorge Fernandez-de-Cossio-Diaz}%
\email{j.cossio.diaz@gmail.com}
\author{Simona Cocco}%
\author{R\'emi Monasson}
\affiliation{Laboratory of Physics of the Ecole Normale Supérieure, CNRS UMR 8023 \& PSL Research, Sorbonne Universit\'e, Paris, France}%

\date{\today}

\begin{abstract}
A goal of unsupervised machine learning is to build representations of complex high-dimensional data, with simple relations to their properties. Such disentangled representations make easier to interpret the significant latent factors of variation in the data, as well as to generate new data with desirable features. Methods for disentangling representations often rely on an adversarial scheme, in which representations are tuned to avoid discriminators from being able to reconstruct information about the data properties (labels). Unfortunately adversarial training is generally difficult to implement in practice. Here we propose a simple, effective way of disentangling representations without any need to train adversarial discriminators, and apply our approach to Restricted Boltzmann Machines (RBM), one of the simplest representation-based generative models. Our approach relies on the introduction of adequate constraints on the weights during training, which allows us to concentrate information about labels on a small subset of latent variables. The effectiveness of the approach is illustrated with four examples: the CelebA dataset of facial images, the two-dimensional Ising model, the MNIST dataset of handwritten digits, and the taxonomy of protein families. In addition, we show how our framework allows for analytically computing the cost, in terms of log-likelihood of the data, associated to the disentanglement of their representations.
\end{abstract}

\maketitle

\section{Introduction}

\begin{figure}[t]
\includegraphics[width=\linewidth]{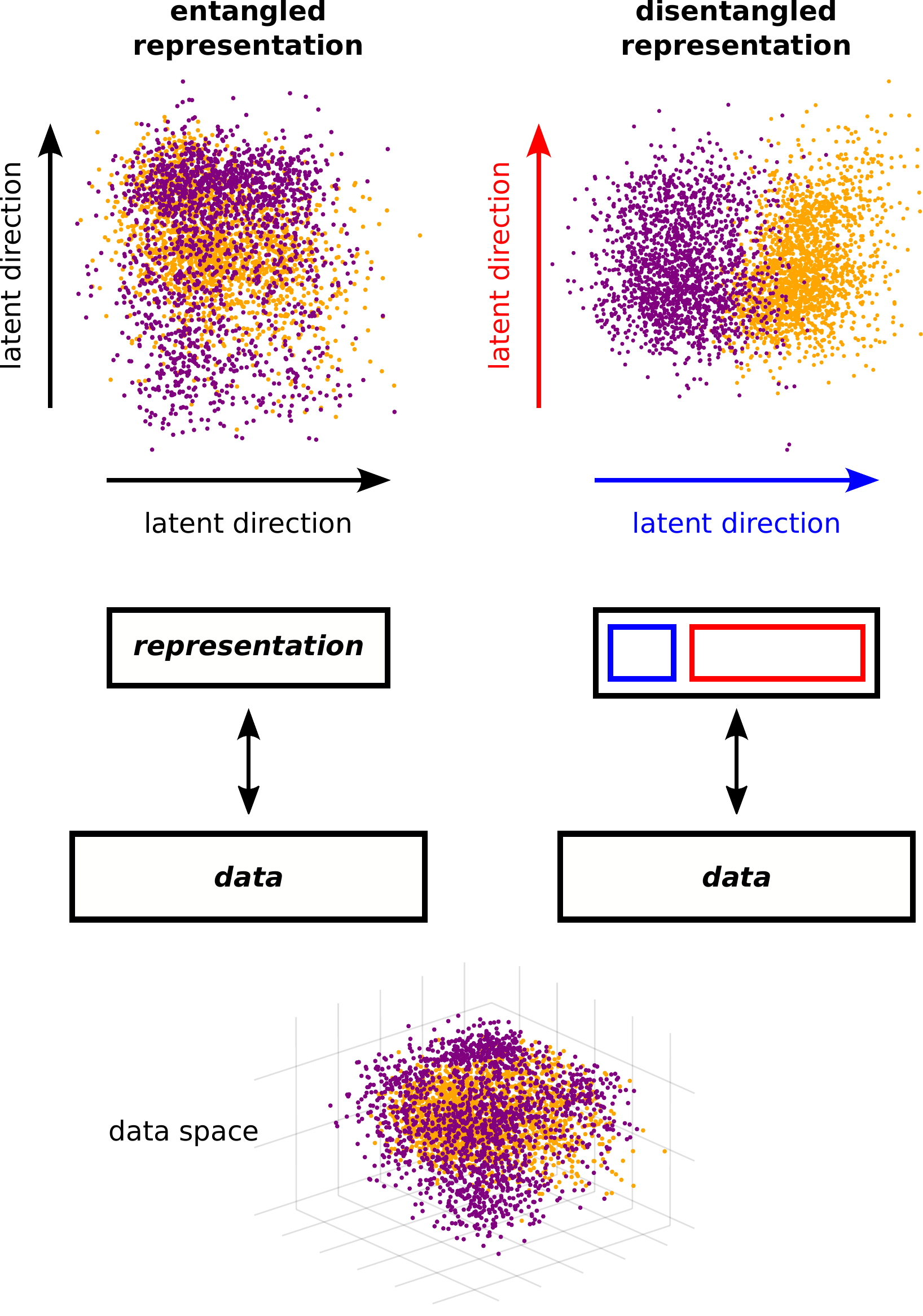}
\caption{\label{fig:intro-entanglement}\textbf{Entangled vs. disentangled representations.} A set of high-dimensional data points (bottom) is mapped, through unsupervised learning, onto a latent representation (top). Data are colored in purple and orange according to a binary-valued attribute, \emph{e.g.} being an odd or even number for MNIST images of handwritten digits. Left: When representations are \emph{entangled}, the separation of data classes is not aligned with a single latent direction. Right: When representations are \emph{disentangled}, one or few directions in latent space (blue) separate the labeled classes, while other directions are not correlated with the label (red).}
\end{figure}

Unsupervised learning involves mapping data points to adequate representations, where the statistical features relevant to the data distribution are encoded by latent variables \cite{bengio2012deep}. Examples of unsupervised architectures include restricted Boltzmann machines \cite{salakhutdinov2009deep}, variational auto-encoders \cite{kingma2013auto}, and generative adversarial networks \cite{goodfellow2014generative}, among others. However, the mapping between latent-variable activities and the relevant properties of the data is generally complex and not easily interpretable (Figure \ref{fig:intro-entanglement}), a phenomenon referred to as {\em entanglement} of representations in machine learning, or mixed sensitivity in computational neuroscience \cite{johnston2020nonlinear}. Entangled representations are hard to interpret and to manipulate, \emph{e.g.} for generating new data with desired properties \cite{bengio2012deep,locatello2019challenging}. 

A stream of literature has recently focused on how to train unsupervised models to obtain disentangled representations, where information about certain properties is concentrated in some latent variables and excluded from others \cite{lample2017fader,kim2018disentangling,hu2018disentangling,esmaeili2019structured,he2019attgan,shen2020interpreting,zaidi2020measuring}, or absent altogether from representations \cite{feutry2018learning,zemel2013learning}. Concentration of information makes, in turn, possible to change the values of few variables and generate data points with controlled properties \cite{lample2017fader}. In practice, learning of disentangled representations is often done in an adversarial framework through optimization of variational bounds to quantities hard to estimate, such as mutual information between the data features and some part of the representations. While conceptually appealing, this approach may be tricky to adopt from a numerical point of view, due to well-known difficulties in adversarial-based learning \cite{arjovsky2017towards}. In addition, its complexity has prevented theoretical analysis so far, leaving important questions, such as the cost of disentangling representations unanswered.

As a concrete illustration, which we consider later on in this work, imagine training an unsupervised model from a set of face images. Once learning is complete the model can be used to generate many new faces, generalizing from the features in the training data. Generated images will show smiling faces, wearing eye-glasses, with bald heads, ... {\em i.e.} will be characterized by a collection of attributes. From a practical point of view, disentangling the representations of those data would make possible, in the generation process, to control and modify one of these attributes, such as smiling vs. not smiling while leaving the remaining ones (the overall shape of the face) unchanged. From a conceptual point of view, the coordinates of the representation space are explicitly related to the different attributes. Moving from one face with eyeglasses to the `same' face without corresponds to a translation of the representation vector of the face in the low-dimensional space defined by the few coordinates associated to the eye-glasses attribute, a property bearing some analogy with Word2Vec encodings \cite{word2vec}.

The purpose of the present work is to propose a method for disentanglement of representations, which is both effective on real data and amenable to mathematical analysis. We consider Restricted Boltzmann Machines (RBM), a simple unsupervised generative model implementing a data/representation duality \cite{hinton2012practical}. RBMs are used as building bricks of deeper networks \cite{salakhutdinov2009deep}, and are competitive with more complex models in various relevant situations \cite{tubiana2019learning,bravi2021rbm,salakhutdinov2007restricted}. We derive conditions on the RBM parameters, which deprive all or part of the representation from information about data labels. This procedure allows us to concentrate the information about labels into a subset of latent units. Manipulation of these units then allows us to generate high-quality data with prescribed label values. Furthermore, the simplicity of our framework allows us to estimate the loss in log-likelihood resulting from the disentanglement requirement, with a deep connection with Poincar\'e separation theorem \cite{abadir2005matrix}. Informally speaking, this loss is the cost to be paid for enhanced interpretability of the machine.

Our paper is organized as follows. We first show that standard learning with RBM generically produces entangled representations on four applications, chosen for their diversity and interest: (1) the CelebA dataset of face images \cite{liu2015faceattributes} annotated with several binary attributes; (2) the two-dimensional Ising model, where configurations are annotated by the sign of their magnetizations; (3) the MNIST dataset of handwritten digits \cite{deng2012mnist}, where the digits represented in each image are the labels; and (4) protein sequence families from the PFAM database \cite{el2019pfam} annotated based on their taxonomic origins. We then present how our approach learns disentangled representations, and demonstrate its effectiveness when applied to the three data distributions listed above. Special emphasis is brought to the physical meaning of the unsupervised models corresponding to the Ising model case. We then calculate the costs associated to representation disentanglement.

\section{Representations of complex data with Restricted Boltzmann Machines are generally entangled}

\subsection{Unsupervised learning with RBM}

\begin{figure*}[t]
\includegraphics[width=0.8\textwidth]{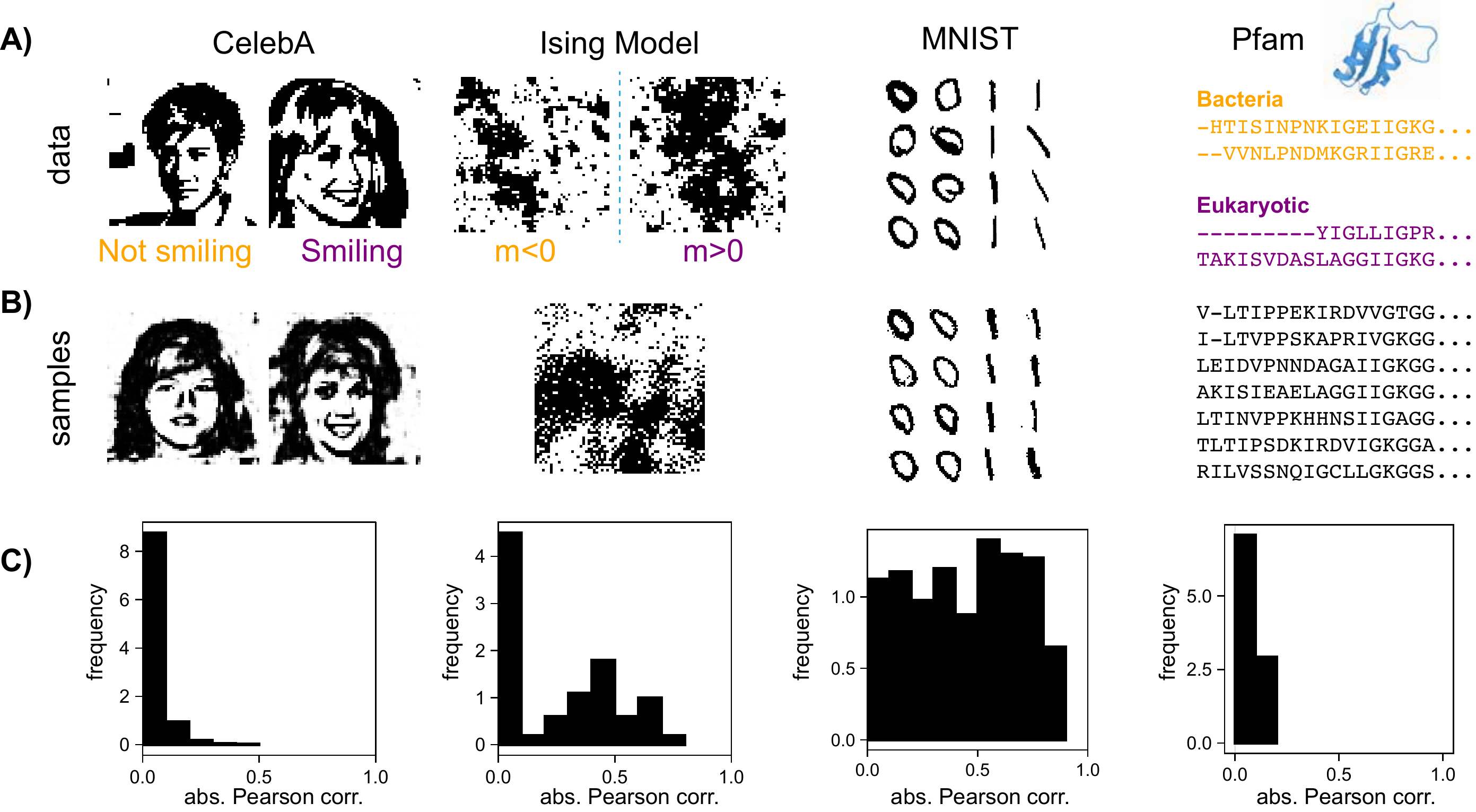}
\caption{\label{fig:datasets} \textbf{Datasets considered in the paper and entanglement of representations.} \textbf{A)} CelebA dataset of face images \cite{liu2015faceattributes}; 2-dimensional Ising model; MNIST0/1 database of handwritten digits \cite{deng2012mnist}; multiple sequence alignments from the PFAM PF00013 family of the KH domain. \textbf{B)} Samples generated by different RBMs trained on each dataset. See SM Appendix \ref{SI:sec:RBM-architecture} for the architectures of the RBMs used in each case. \textbf{C)} Histogram of the absolute value of the Pearson correlations between hidden unit inputs and the chosen label, see Eq.~\eqref{eq:pearsonlin}: Smiling or not smiling for CelebA, sign of the magnetization for the Ising model, whether the digit is a 0 or 1 for MNIST0/1, and whether the KH sequence is from bacterial or eukaryotic origin.}
\end{figure*}

Restricted Boltzmann Machines (RBM) are bipartite graphical models over $N$ visible variables $\mathbf{v}=\{v_1,v_2,...,v_N\}$ and $M$ hidden (or latent) variables $\mathbf{h}=\{h_1,h_2,...,h_M\}$, see Figure~\ref{fig:datasets}A. Both visible and hidden variables are assumed to be Bernoulli, \emph{i.e.} to take 0 or 1 values. The two layers are connected through the interaction weights $w_{i\mu}$. An RBM defines a joint probability distribution over $\mathbf{v}$ and $\mathbf{h}$ through
\begin{equation}\label{eq:Prbm}
P(\mathbf v,\mathbf h) = \frac{1}{Z}\mathrm e^{-E(\mathbf v, \mathbf h)},
\end{equation}
where $Z$ is a normalizing factor and the energy $E$ is given by
\begin{equation}\label{eq:rbm-energy}
E(\mathbf v, \mathbf h) = -\sum_{i=1}^N g_i v_i - \sum_{\mu=1}^M \theta_\mu \, h_\mu - \sum_{\mu=1}^M I_{\mu} (\mathbf{v}) \, h_\mu
\end{equation}
The parameters $g_i$ and $\theta_\mu$ are local fields biasing the distributions of single units, and
\begin{equation}\label{eq:input}
I_\mu(\mathbf{v}) = \sum_{i=1}^N w_{i\mu} v_i 
\end{equation}
is the input received by hidden unit $\mu$ given the visible configuration.

Marginalizing over the states of the hidden units results in the likelihood $P(\mathbf{v}) =\frac{1}{Z}\sum_\mathbf{h} \mathrm{e}^{-E(\mathbf{v},\mathbf{h})}$ of visible configurations that can be fit to data. Given a set of data points ${\cal D}$, the weights and potential-defining parameters of the RBM are learned through gradient ascent of the dataset log-likelihood,
\begin{equation}\label{eq:Lrbm}
\mathcal{L} = \langle \log P(\mathbf{v})\rangle _{\cal D}\ ,
\end{equation}
where the average $\langle\cdot\rangle_{\cal D}$ is taken over the data points. In practice computing the gradient of $\mathcal{L}$ requires to estimate the moments of visible and/or hidden variables with respect to the model distribution \cite{hinton2012practical}. Regularization of the weights can also be easily included in this approach. Details about the computation of the gradient and the training procedure implemented in this work can be found in SM Appendix \ref{SI:sec:impl}.

\subsection{Datasets}

We train the RBM on four datasets, illustrated by the four columns in Figure~\ref{fig:datasets}:

\subsubsection{CelebA face images dataset}

The CelebA dataset consists of a collection of 202,599 color images of celebrity faces, each annotated with 40 binary attributes, including whether the person is smiling, wearing glasses, has a beard, and others \cite{liu2015faceattributes}. The images in this dataset cover large pose variations and background clutter. Figure \ref{fig:datasets}A shows a pair of black-and-white versions of CelebA examples, see SM Fig. \ref{SI:fig:celeba_sauvola} for more examples and SM Appendix \ref{SI:sec:celeba-processing} for processing details.

\subsubsection{Two-dimensional Ising model}

We next consider the Ising model \cite{baxter2016exactly} on a two-dimensional regular $L\times L$ square grid ($L=32$ or 64), with uniform positive interactions between nearest-neighbour spins. The values of the interaction, or, equivalently, of the inverse temperature are varied to explore both paramagnetic (weak interations) and ferromagnetic (strong interactions) regimes. Data are configurations of the Ising model generated by Monte Carlo, and labeled according to the sign $u$ of its magnetization $m$, \emph{i.e.} the differences between the numbers of $+$ (black dots in Figure~\ref{fig:datasets}A) and $-$ spins (white dots). 

\subsubsection{MNIST handwritten digits}

The MNIST dataset \cite{deng2012mnist} consists of a collection of 70,000 images of $28\times28$ pixels each, labeled by the identity of the 0-9 handwritten digit they represent. We show 16 of them in Figure~\ref{fig:datasets}A. We hereafter consider in particular (1) \emph{MNIST0/1}, a simplified version of MNIST consisting only of images of the digits 0 and 1, with binary labels $u=0,1$; and (2) \emph{MNIST0/1/2/3}, the set of all images of digits from 0 to 3, with 4-state labels $u$. In SM Fig. \ref{SI:fig:mnist_black_white} we have also considered an additional example consisting of zero digits only, in black or white backgrounds (see \ref{sec:mnist_bw_generative}).

\subsubsection{PFAM database of protein family sequences}

Last of all, we consider protein families in the PFAM sequence database \cite{el2019pfam}. A protein family consists of a collection of homologous protein sequences from different organisms, \emph{i.e.} sharing common evolutionary origins and common functional or structural features. As an illustration Figure~\ref{fig:datasets}A sketches some sequences of the K Homology (KH) domain found in nucleic-acid binding proteins. Many families include sequences issued from prokaryotic and eukaryotic organisms, and we use this classification as the label $u$ for sequences in the dataset. 

\subsection{RBMs generically learn entangled representations}

We trained RBMs with 200 - 400 binary hidden units on CelebA images, 2-dimensional Ising model configurations, MNIST0/1 digits, and KH domain protein sequences (see SM Appendix \ref{SI:sec:RBM-architecture} for details). Consistent with previous results on similar datasets \cite{tubiana2019learning,bravi2021rbm,yevick2021accuracy,harsh2020place}, RBM accurately fit the data, and generate high-quality samples in the four cases, see Figure \ref{fig:datasets}B. In addition, training simple classifiers to predict the label from the hidden inputs of the models, gives areas under the curve (AUC) $> 0.9$ for all cases, see SM Appendix \ref{SI:sec:classifiers} for details and SM Fig. \ref{SI:fig:inputs_classifier}. These results demonstrate that the RBM automatically captures information relevant to the labels of interest. We emphasize that in all cases the RBM did not have access to the labels during training. 

We plot in Figure \ref{fig:datasets}C the histogram of Pearson correlations between the label and hidden unit inputs,
\begin{equation}\label{eq:pearsonlin}
\rho_\mu = \frac{\langle u(\mathbf{v})\, I_\mu(\mathbf{v})\rangle_{\cal D} - \langle u(\mathbf{v})\rangle_{\cal D} \, \langle I_\mu(\mathbf{v})\rangle_{\cal D} }{\sqrt{\langle I_\mu(\mathbf{v})^2\rangle_{\cal D} - \langle I_\mu(\mathbf{v})\rangle_{\cal D}^2}\sqrt{\langle u(\mathbf{v})^2\rangle_{\cal D} - \langle u(\mathbf{v})\rangle_{\cal D}^2} }\ .
\end{equation}
For some datasets ({\em e.g.} KH sequences), hidden units have low correlations to the label. Changing the label identity of generated data would require to act on the states of all these hidden units in a concerted manner. In other cases, such as Ising model and MNIST a number of units exhibit higher correlations with the labels, see right tails of distributions in Figure~\ref{fig:datasets}C. However, as the label information captured by the RBM is distributed among these units, manipulating the few most correlated units is not sufficient to define the label of generated data, see SM Fig. \ref{SI:fig:ising_entangled}.

Although a precise definition of disentangled representation learning may be debated \cite{zaidi2020measuring,locatello2019challenging}, it is generally agreed upon that interesting features should map to single, or few dimensions in latent space, see Figure \ref{fig:intro-entanglement} \cite{bengio2012deep}. As shown above standard training of RBM fails to produce disentangled representations.

\section{Learning of disentangled representations}

\begin{figure}
\includegraphics[width=\linewidth]{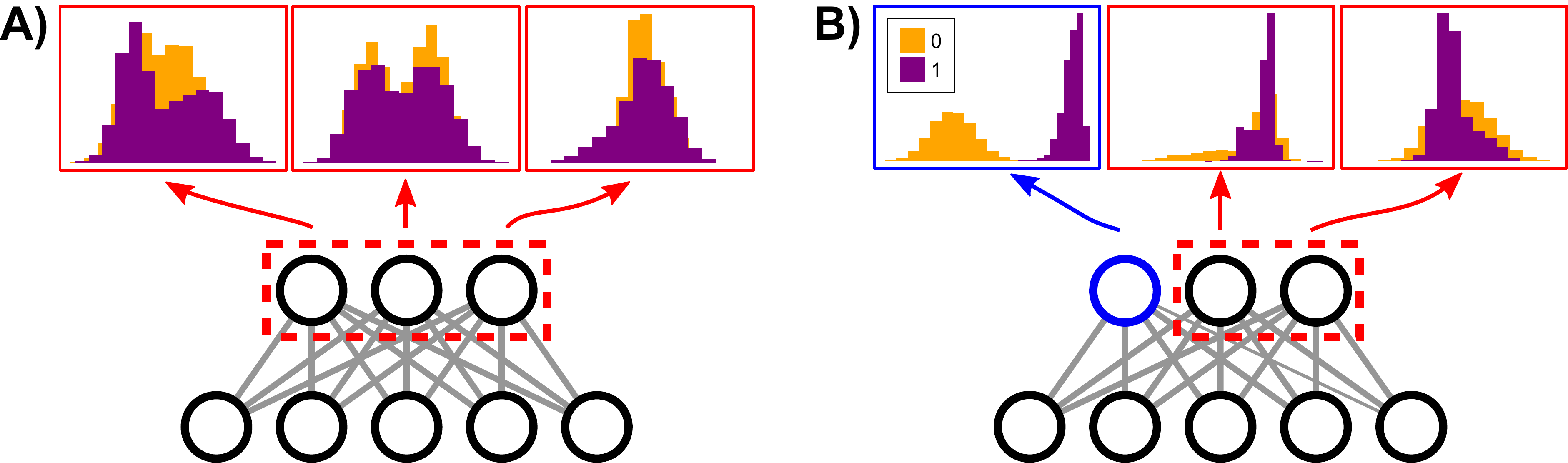}
\caption{\label{fig:architectures}\textbf{Model schema.} \textbf{A)} Constraints imposed on all hidden units, promote overlapping hidden input distributions of the two classes. \textbf{B)} Constraints imposed on a subset of hidden units (red), promotes class separation on the remaining hidden units (blue).}
\end{figure}

Our strategy for disentangling and manipulating representations is to drastically alter the distribution of correlations between hidden units and labels (Figure~\ref{fig:datasets}C) by imposing appropriate constraints on the interaction weights throughout the learning process. 

Ideally, constraints should impose that mutual information, rather than correlations, vanishes. Due to the difficulty in computing mutual information we focus on correlations, at different orders in the hidden inputs, as they offer a good compromise between computational efficiency and performance. Focusing on inputs $I_{\mu}$ rather than on latent variables $h_{\mu}$ follows a two-fold motivation. First, the constraints on the weights $w_{i\mu}$ resulting from the vanishing requirements on the correlations are simpler to interpret and to fulfill from a computational point of view. Second, given a data configuration $\mathbf{v}$, $h_{\mu}$ is a stochastic variable conditioned to $I_{\mu}$; by virtue of the data processing inequality \cite{cover1999elements} the mutual information between labels $u$ and inputs $I_{\mu}$ upper bounds its counterpart between $u$ and $h_{\mu}$; enforcing low mutual information between labels and inputs therefore immediately imply that latent variables are not informative about labels.

Two objectives can be pursued:
\begin{enumerate}[label=\Alph*.]
\item {\em Approximating as best as possible the data distribution, while removing as much information as possible about their labels}. This can be achieved by an architecture in which all hidden units are under strong constraints, see Figure~\ref{fig:architectures}A. Objective A leads to a generic model distribution in which label-associated features are blurred, {\em i.e.} it is hard to tell whether they are present or absent. Conversely, the other `orthogonal' features are well captured by this RBM model.

\item {\em Reproducing as best as possible the data distribution, while concentrating as much information as possible about their labels on one (or few) hidden units}. This can be achieved by an architecture in which a few hidden units are left unconstrained and are referred to as {\em released}, while all the other ones are under strong constraints, see Figure \ref{fig:architectures}B. Objective B defines a model distribution, in which label-associated features are either present or absent, as in the training data. In addition the representations can be easily manipulated to bias data generation, {\em e.g.} to morph one configuration into another one in which the label value has changed but other features have not.
\end{enumerate}

For the sake of simplicity we present the approach in the case of binary labels $u=0,1$ (equivalently, $u=\pm 1$). An extension to labels with more than two values is immediate, and will be discussed in the applications.
 
\subsection{Fully constrained RBM}

\begin{figure}
\includegraphics[width=\linewidth]{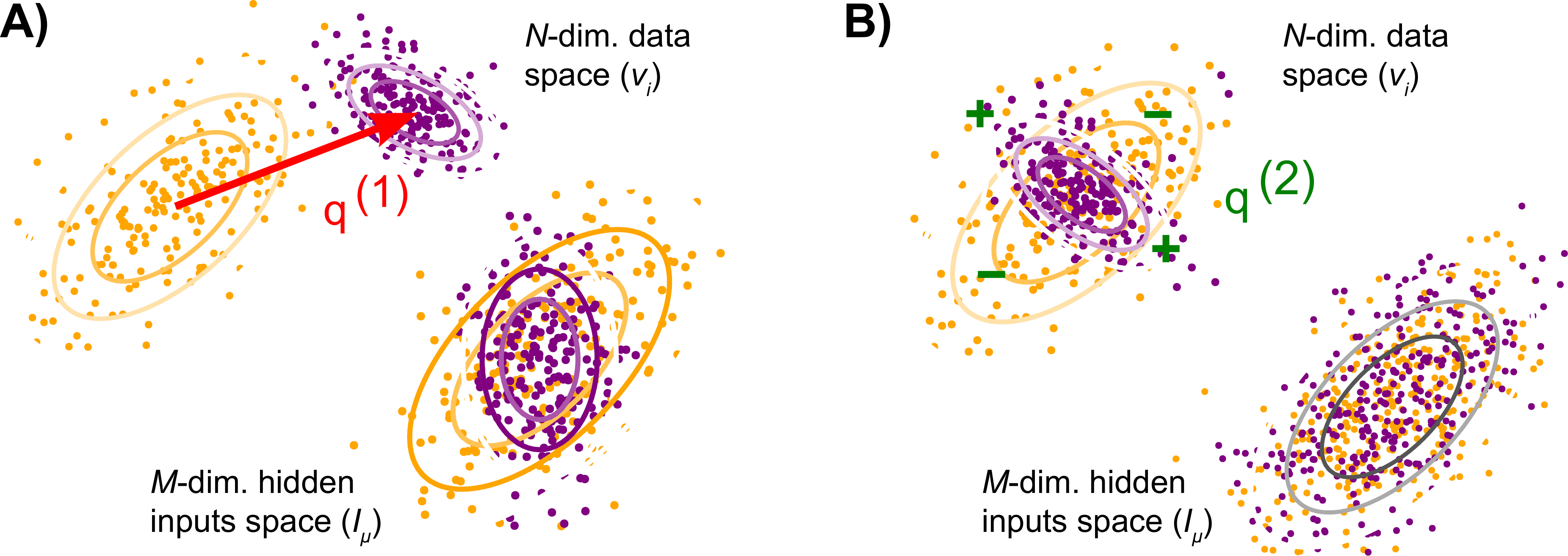}
\caption{\label{fig:constraints}\textbf{First and second-order constraints.} \textbf{A)} The first-order constraint \eqref{eq:Worth} ensures that the classes have the same means in input space, by imposing orthogonality of the weights to the vector separating their centers of masses in data space (red). \textbf{B)} Second-order constraints \eqref{eq:Worth2} ensure that the two classes have the same covariance in input space.}
\end{figure}

Following objective A we demand that all hidden-unit inputs $I_\mu$ are uncorrelated with the labels $u$ across the data. The corresponding architecture is sketched in Figure~\ref{fig:architectures}A. A RBM trained under these constraints defines a distribution, in which information about the label has been degraded, if not fully erased, but the other data-defining features are affected as little as possible.

\subsubsection{Linear constraints}

In its simplest formulation the approach only considers linear correlations in the inputs. The constraint $\rho_\mu = 0$, see Eq.~\ref{eq:pearsonlin}, can be rewritten as
\begin{equation}\label{eq:Worth}
\sum_{i=1}^N q^{(1)}_i\, w_{i\mu} = 0\ ,
\end{equation}
with
\begin{equation}\label{eq:q1}
q_i^{(1)}=\langle u(\mathbf{v})\, v_i\rangle_{\cal D} - \langle u (\mathbf{v})\rangle_{\cal D} \langle v_i \rangle_{\cal D}\ .
\end{equation}
The $N$-dimensional vector $\mathbf q^{(1)}$ is parallel to the line joining the centers of mass of the clouds of data points associated to, respectively, $u=0$ and $u=1$, see Figure~\ref{fig:constraints}A. Imposing $\rho_{\mu}=0$ for all $\mu=1,..,M$ is thus equivalent to looking for the RBM maximizing the log-likelihood ${\cal L}$ in Eq.~\ref{eq:Lrbm} under the constraints that all $M$ weight vectors $\mathbf w_\mu$ are orthogonal to $\mathbf q^{(1)}$; this can be easily done by projecting the gradient of ${\cal L}$ onto the space orthogonal to $\mathbf{q}^{(1)}$ after each update of the weights (see SM Appendix \ref{SI:sec:impl} for details). In other words, the RBM is blind to the direction $\mathbf q^{(1)}$ separating the clouds and is modeling only the statistical features of the data in the $N-1$-dimensional space orthogonal to $\mathbf q^{(1)}$.

The consequences of $\mathbf w_\mu \perp \mathbf q^{(1)}$ can be phrased in an adversarial context. Imagine a linear discriminator is trying to predict the labels $u(\mathbf v)$ of data configurations $\bf v$ based on the $M$-dimensional sets of inputs $I_\mu(\mathbf v)$. In practice a linear discriminator is parameterized by $M$ weights $a_\mu$, and assigns a probability $\pi \big(\sum_\mu a_\mu \,I_\mu(\mathbf v)\big)$ to, say, label $u=1$ (and probability $1-\pi$ to $u=0$) given $\bf v$, where $\pi$ is some sigmoid function comprised between 0 and 1. The parameters $a_\mu$ are fitted to maximize the probability that the discriminator makes the correct prediction. In geometrical terms, this is equivalent to finding the hyperplane (orthogonal to $\mathbf a$ in $M$ dimensions) separating the classes of data points $\mathbf I$ associated to $u=0$ and to $u=1$ with the largest margin \cite{engel2001statistical}. We show in SM Appendix \ref{SI:sec:adv} that, under the conditions expressed in Eq.~(\ref{eq:Worth}), the best linear discriminator cannot do better than random guessing of the labels. In other words, imposing constraints \eqref{eq:Worth} is equivalent to demanding that no adversarial linear discriminator looking at hidden-unit inputs is able to predict the labels associated to configurations.

\subsubsection{Quadratic constraints}

Even if no linear discriminator can recover the label from the inputs $I_\mu$, more complex machines, such as deep neural networks, could still be able to predict the label \cite{brenner2000adaptive} if the mutual information between $u$ and $\mathbf I=(I_1,I_2,...,I_M)$ is non-zero. Imposing $\rho_\mu=0$ can be seen as a first-order approximation to the stronger condition that the mutual information between the label and the inputs vanishes, $\mathrm{MI}(u,\mathbf I) = 0$. The later implies that not only the linear correlations but also all higher-order connected moments between $u$ and $\mathbf I$ vanish. In particular, the second-order correlations
\begin{equation}\label{eq:constraint2nd_h}
C_{\mu,\nu} = \langle u(\mathbf{v})\, I_\mu(\mathbf{v}) I_\nu(\mathbf{v}) \rangle_{\cal D} - \langle u(\mathbf{v})\rangle_{\cal D} \, \langle I_\mu(\mathbf{v})I_\nu(\mathbf{v})\rangle_{\cal D} 
\end{equation}
should also vanish. Setting $C_{\mu,\nu}=0$ for all pairs $\mu,\nu$ in Eq.~\eqref{eq:constraint2nd_h} forces the two classes of data attached to $u=0$ and $u=1$ to have identical covariance matrices in the input space. These constraints imply that no kernel-based adversarial discriminator, where the kernel is a quadratic function of the inputs, would be able to predict the label values (see SM Appendix \ref{SI:sec:adv} for a proof). More generally, higher-order constraints would rule out the possibility for discriminator adversaries with polynomial kernels of higher degrees to successfully classify the data \cite{scholkopf2018learning} (see SM Appendix \ref{SI:sec:adv})).

In practice, setting $C_{\mu,\nu}=0$ amounts to imposing a quadratic constraint over the weight vectors:
\begin{equation}\label{eq:Worth2}
\sum_{i,j=1}^N q^{(2)}_{i,j}\, w_{i\mu} \, w_{j\nu}=0,
\end{equation}
where the mean difference between the covariance matrices of the two classes of data is defined through
\begin{equation}\label{eq:Q2}
q^{(2)}_{i,j} = \langle u(\mathbf{v})\, v_i v_j\rangle_{\cal D} - \langle u (\mathbf{v})\rangle_{\cal D} \langle v_i v_j\rangle_{\cal D}\ ,
\end{equation}
see illustration in Figure~\ref{fig:constraints}B. To draw a physical analogy, the $\mathbf q^{(2)}$ matrix looks like the quadrupole tensor separating positive and negative charge distributions in electrostatics, while $\mathbf q^{(1)}$ is analogous to a dipole moment.

To implement constraints \eqref{eq:Worth2} in practice, we square the left-hand side of \eqref{eq:Worth2} and add it to the optimization objective during learning times a large penalty term, see SM Appendix \ref{SI:sec:impl} for details.

The matrix $\mathbf{q}^{(2)}$ defined in \eqref{eq:Q2} is estimated on empirical data and is subject to sampling noise. In practice, from finite datasets one can extract reliable estimates only of the top components of $\mathbf{q}^{(2)}$, while the empirically observed lower components will be dominated by noise. The Marchenko-Pastur (MP) law \cite{marchenko1967distribution}, describing the spectrum of correlation matrices in the null model case of independent variables, can be used to estimate the thresholds between eigenvalues dominated by noise and eigenvalues reflecting the presence of structure in the data. The MP spectrum predicts that all eigenvalues $\lambda$ located in the range $[\lambda_-;\lambda_+]$ have to be discarded, with $\lambda_\pm = (1 \pm \sqrt{r})^2$, where $r$ is the ratio of the numbers of variables and of samples. As an example, for the MNIST0/1 dataset, we estimate $\lambda_+\simeq 1.6$ for both 0 and 1 digits. Out of the 784 eigenvalues of $\mathbf{q}^{(2)}$, only 60 (61) are larger than this bound for the 0's dataset (1's). The above discussion suggests replacing the full matrix $\mathbf{q}^{(2)}$ with a low-rank approximation focusing on the top components only. A lower-rank version of $\mathbf{q}^{(2)}$ also implies that the weights have more degrees of freedom, since \eqref{eq:Worth2} does not affect the weights components belonging to the kernel of $\mathbf{q}^{(2)}$. In practice, penalizing the squared norm of the left-hand side of \eqref{eq:Worth2} during training, automatically places more weight on constraints associated to the top components of $\mathbf{q}^{(2)}$, and neglects lower components.

\subsection{Partially constrained RBM}

We now consider Goal B. Our objective is to concentrate the information about the labels on one of few released hidden units. For this purpose we consider the architecture of Figure \ref{fig:architectures}B. The weights attached to these released hidden units are unconstrained during training, while the other weights are subject the to linear or quadratic constraints in Eqs.~\eqref{eq:Worth} \& \eqref{eq:Worth2}, as in Goal A. Informally speaking, this strategy will turn the large number of  weak input-label correlations found in standard RBM representations (Figure \ref{fig:datasets}C) into a small number of large correlations ($\propto M$) present on the released hidden units only.

\subsubsection{Manipulation of label-determining hidden units}

As a consequence,  the values of the released hidden units strongly affect the conditional distribution of visible configurations, and act as knobs that can be manipulated to generate data with desired labels. Manipulation is carried out as follows; to lighten notations we assume that a single hidden unit, say, $\mu=1$, is released. The value of this unit, $h_1$, is fixed (to 0 or 1). We then sample the remaining hidden units (attached to the constrained weights) and the visible units using alternate Gibbs sampling (SM Appendix \ref{SI:sec:impl}). The visible configurations $\mathbf{v}$ are then distributed according to a conditional probability $P(\mathbf{v}|h_1)$, and span a class of the data corresponding to a specific label value $u$. Flipping $h_1$ to $1-h_1$ allows us to change class, and quickly morph a data configuration into the closest configuration with a flipped label.

\subsubsection{Cost of disentanglement} \label{sec:introcost}

Constraining all weight vectors (Goal A) is damaging the capability of RBM to reproduce the data distribution. The loss in performance is measured by the change in log-likelihoods of test data due to the partial erasure of information about the labels,
\begin{equation}\label{eq:cost1a}
\Delta{\cal L}_\text{part. erasure} = {\cal L}_\text{unconstr.}- {\cal L}_\text{constr.} \ .
\end{equation}
In the equation above,   ${\cal L}_\text{constr.}$ denotes the log-likelihood of data estimated with the fully-constrained RBM, and ${\cal L}_\text{unconstr.}$ corresponds to the standard (unconstrained) RBM. As we shall see in subsequent applications this difference is generally large.

Once one or few hidden units are released (Goal B), the test log-likelihood increases to ${\cal L}_\text{rel.}$. We define the cost for disentangling representations through
\begin{equation}\label{eq:cost1b}
\Delta{\cal L}_\text{disent.} = {\cal L}_\text{unconstr.} - {\cal L}_\text{rel.} \ .
\end{equation}
This cost is guaranteed to be non-negative if both RBM are trained with equal hyperparameters, \emph{e.g.} have the same number of hidden units and weight regularizations.

\section{Application to face images}

\begin{figure*}
\includegraphics[width=0.8\textwidth]{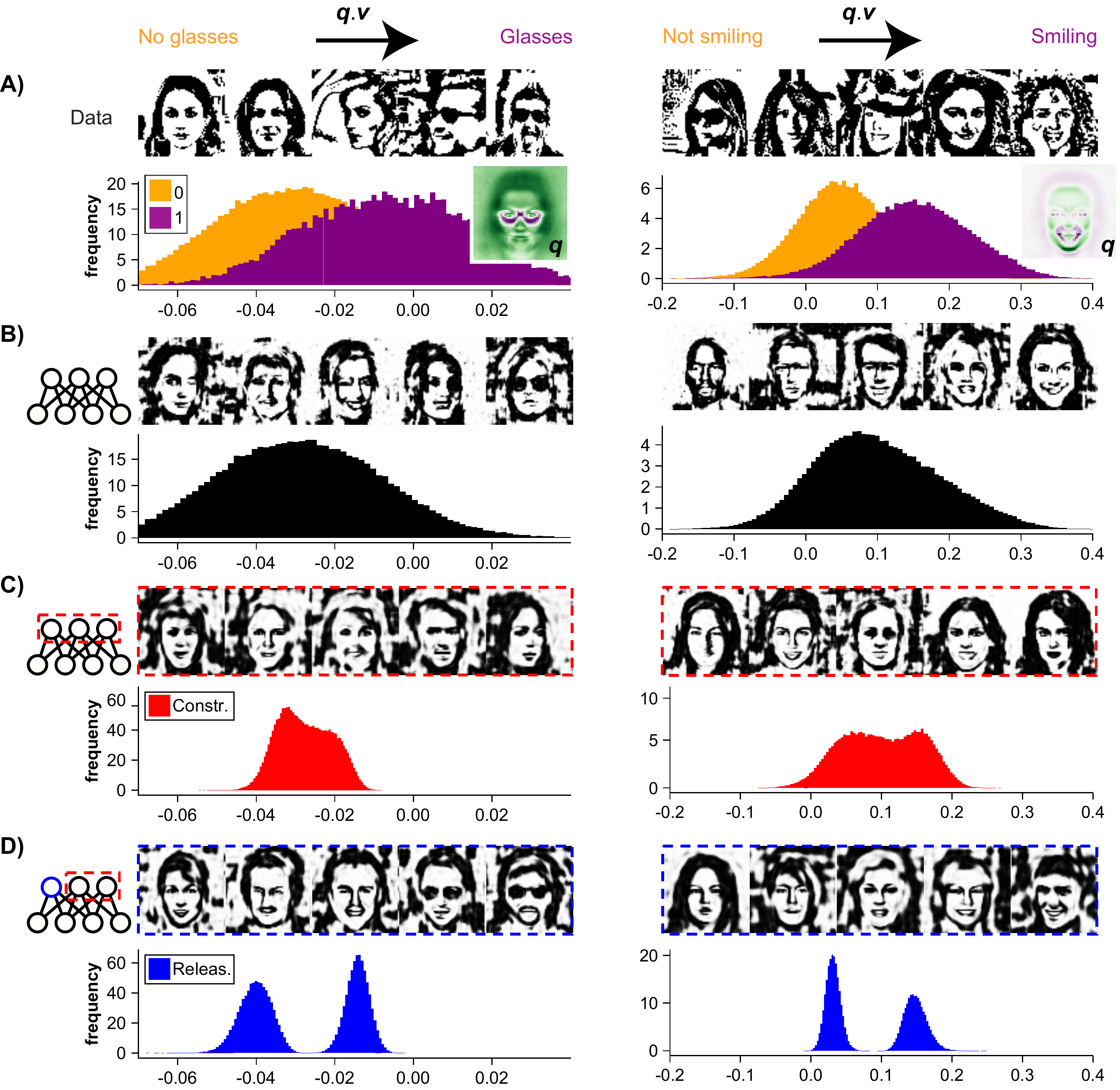}
\caption{\label{fig:celeba_sampling}\textbf{Application to CelebA dataset.} Left: label = presence/absence of eye-glasses; right: label = smiling/not smiling. \textbf{A)} Selected images from the data, arranged by the value of their projection along the vector $\mathbf{q}$ defined in Eq.~\eqref{eq:q1}. Below, the histogram of these projections computed for all images in the data. The inset shows a heatmap of the vector $\mathbf{q}$. \textbf{B)} Samples generated by an unconstrained RBM, and histogram of their projections on vector $\mathbf{q}$. \textbf{C)} Samples generated by a RBM, all the hidden units of which are subject to the constraint in Eq.~\eqref{eq:Worth} (dashed red). The histogram (red) of projections on $\mathbf{q}$ concentrates on intermediate values. \textbf{D)} Samples generated by a RBM trained under constraint \eqref{eq:Worth} acting on all but one released hidden unit (dashed blue), and histogram of projections along $\mathbf{q}$ (blue). Details about the RBMs architecture and training can be found in SM Appendix~\ref{SI:sec:RBM-architecture}.}
\end{figure*}

\subsection{Learning with standard RBM}

We first illustrate our approach on the CelebA dataset of celebrity face images \cite{liu2015faceattributes}. Since we have chosen to work with binary RBMs for simplicity, we first convert the images to binary black and white pixels of resolution $64\times64$, following a procedure similar to \cite{decelle2021equilibrium} and detailed in SM Appendix~\ref{SI:sec:celeba-processing}. Using the annotations available in the dataset, we choose the presence/absence of eyeglasses and smiling/not smiling as our labels. We compute the vector $\mathbf{q}$ defined by \eqref{eq:q1} for each one of these two labels. Figure \ref{fig:celeba_sampling}A shows sample images arranged in increasing value of their projection along this vector, as well as the histograms of these projections over the dataset for each  label.

Next, we train a standard RBM on this dataset. Following \cite{decelle2021equilibrium} we use 5,000 hidden units (SM Appendix~\ref{SI:sec:celeba-processing}). After training, we generate 10,000 samples starting from random binary configurations and running Gibbs sampling for 5,000 iterations. Some sampled configurations are shown in Figure \ref{fig:celeba_sampling}B, as well as the histogram of projections along direction $\mathbf{q}$. Samples are diverse and span the different classes present in the dataset, {\em i.e.} smiling/not smiling, wearing/not wearing eye glasses, indicating that RBM is an adequate generative model for this dataset.

\subsection{Partial erasure of information with fully constrained RBM}

We next consider a RBM with the same architecture and with constraint \eqref{eq:Worth} acting on all hidden units. Figure \ref{fig:celeba_sampling}C shows samples from such a RBM (dashed red). These samples are recognizable faces similar to the data, therefore the model is generative. In the projection on $\mathbf{q}$, they concentrate on intermediate values and seem to be ambiguous with respect to the label-associated feature: eyes seem closed or darkened in the eye-glasses case, and the mouth seems slightly open, but not entirely smiling in the second case. These findings nicely illustrate the effects of objective A. 

\subsection{Manipulating representations and face attributes with partially constrained RBM}

\begin{figure*}
\includegraphics[width=0.9\textwidth]{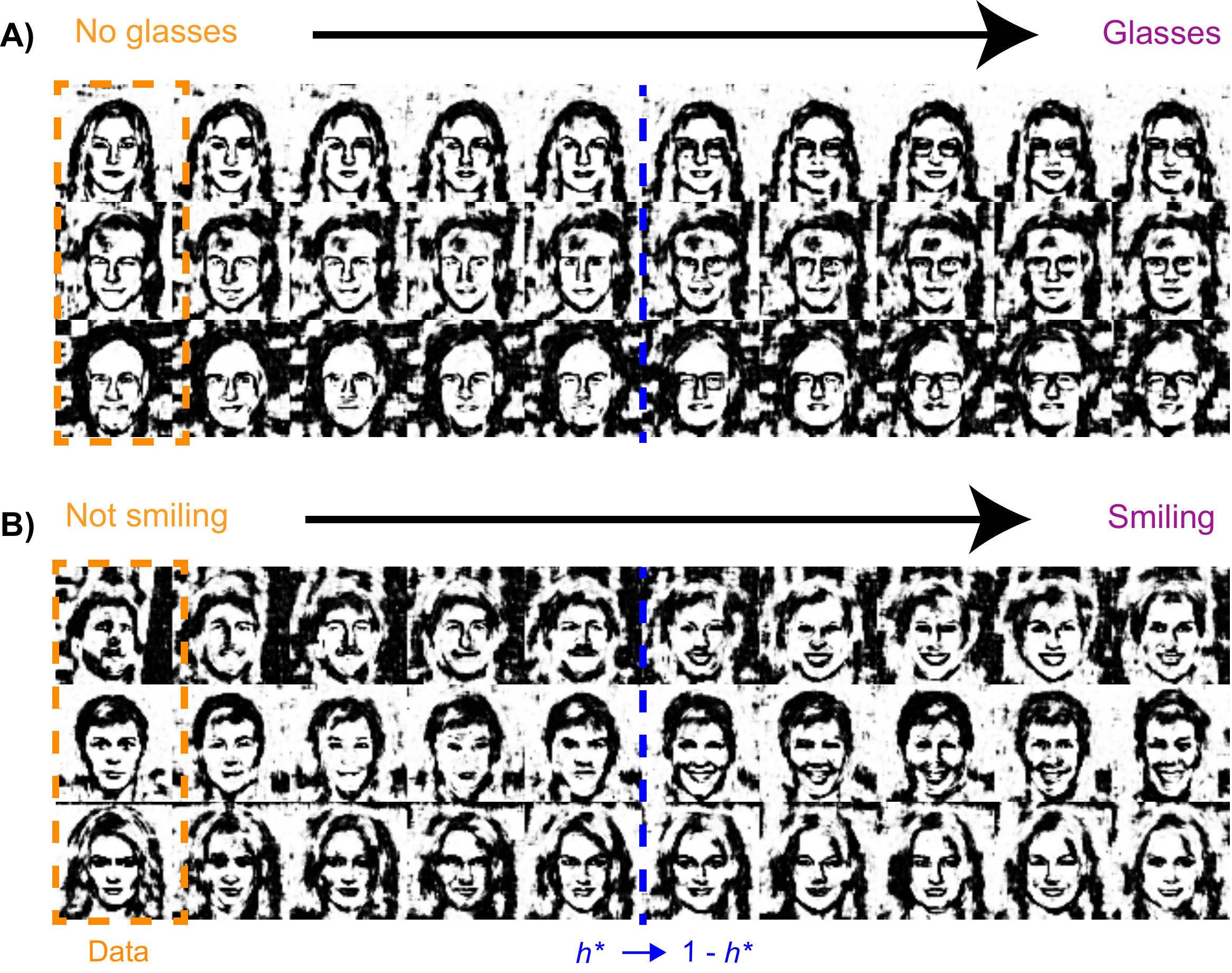}
\caption{\label{fig:celeba_transition}\textbf{Transitions between labeled classes in the CelebA dataset.} RBMs are trained subject to the linear constraint acting on all but the first hidden unit, denoted $h^\star$. Samples are generated conditioned on a frozen value of $h^\star$, which is flipped in the center of the Markov chain (indicated by the dashed blue lines). \textbf{A)} Label corresponds to the ``Eyeglasses'' attribute of CelebA. Samples are collected every 3 Gibbs iterations. \textbf{B)} Label corresponds to the ``Smiling'' attribute of CelebA. Samples are collected every 5 Gibbs iterations.}
\end{figure*}

We now train a RBM with constraint \eqref{eq:Worth} acting on all but one hidden unit, say, $h^\ast$. The weights attached to this unit are correlated with the vector $\mathbf q^{(1)}$, shown in Figure \ref{fig:celeba_sampling}A (inset). The model is generative, representative samples are shown in Figure \ref{fig:celeba_sampling}C, bottom panel. The projection of these samples along the $\mathbf{q}$ direction is bimodal, with two peaks corresponding to the two values of the released hidden unit $h^\ast$. Inspecting the samples shows that $h^\ast$ correlates with the attribute, as shown below, in full agreement with objective B.

The value of $h^\ast$ can be manipulated during sampling to drive the Markov chain toward one class or another. We illustrate this in Figure \ref{fig:celeba_transition}, where an initial sample from the data is sampled through this model and the value of $h^\ast$ is flipped at the midpoint of the sampling chain. As a result, the face images transition toward the expected label value. The transition is smooth: right after the flip of $h^\ast$ most facial features are still preserved, while the one associated to the label  has been modified (morphing effect).

\section{Application to the two-dimensional Ising model} \label{sec:ising}

\begin{figure*}
\includegraphics[width=\textwidth]{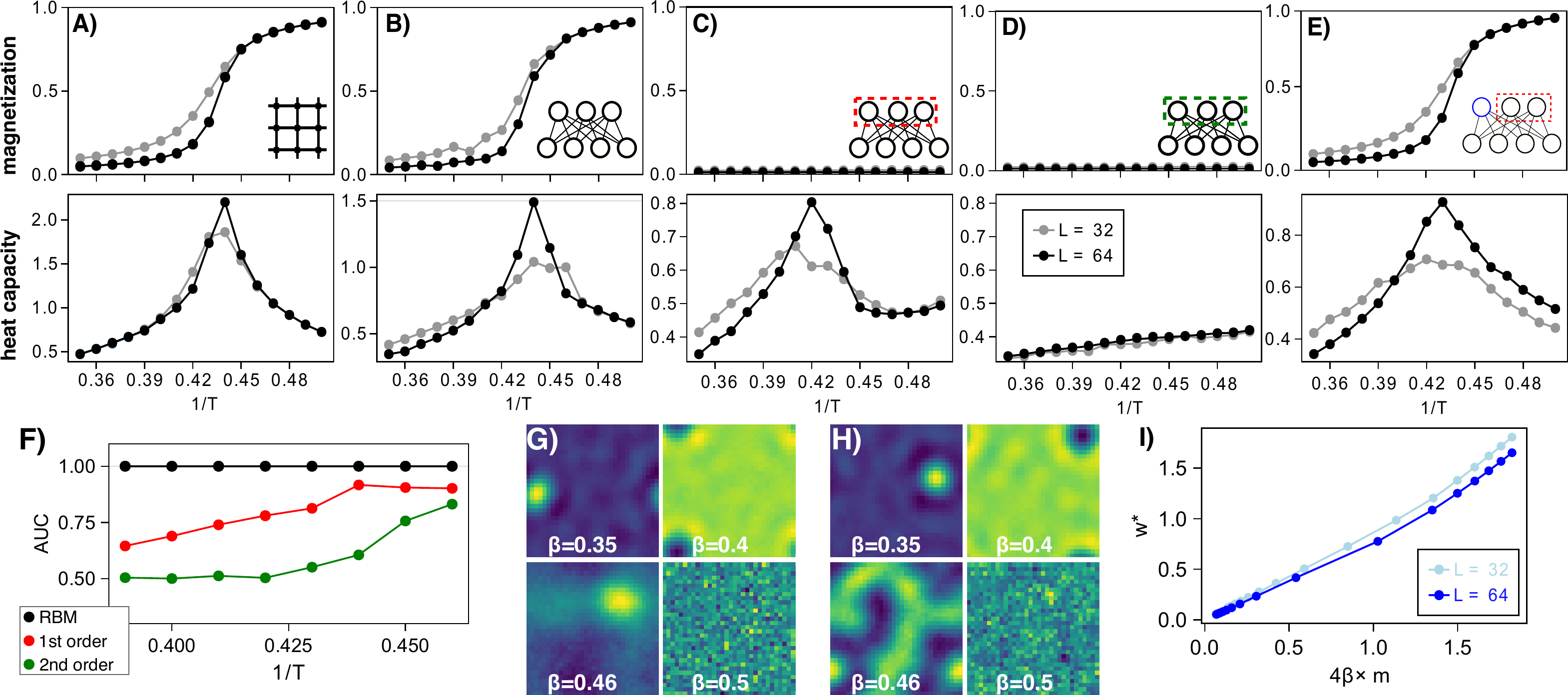}
\caption{\label{fig:ising}\textbf{Learning RBMs on two-dimensional Ising model data.} \textbf{A)} Magnetization and heat capacity as functions of temperature for the samples generated by the Ising model \eqref{eq:ising}. \textbf{B)} Magnetization and heat capacity of samples generated by an RBM trained on the Ising data. \textbf{C)} Magnetization and heat capacity of samples generated by RBM with constraint \eqref{eq:Worth} acting on all hidden units. \textbf{D)} Magnetization and heat capacity of samples generated by RBM with quadratic constraint \eqref{eq:Worth2} acting on all hidden units. \textbf{E)} Magnetization and heat capacity of samples generated by RBM with linear constraint \eqref{eq:Worth} acting on all but one hidden unit. \textbf{F)} Maximum AUC of classifiers trained to predict the sign of the sample magnetization from the RBM inputs. \textbf{G,H)} Typical weights learned by the RBM at selected temperatures ($1/T=0.35, 0.4, 0.46, 0.5$), for the unconstrained RBM, and for the RBM with the 1st-order constraint. \textbf{I)} Free weights attached to the released hidden unit compared to $4\beta$ times the magnetization of the Ising model. }
\end{figure*}

The two-dimensional Ising model is defined by the following energy function over $N=L^2$ spin configurations $\mathbf{v} = (v_1,v_2,....v_{N})$,
\begin{equation}\label{eq:ising}
E(\mathbf{v}) = -\sum_{(i,j)} v_i v_j
\end{equation}
where the sum runs over pairs $(i,j)$ of nearest neighbours on a two-dimensional  squared grid with $L\times L$ sites. Each spin $v_i$ can take $\pm 1$ values. We choose periodic boundary conditions, that is, site $(1,1)$ is interacting with sites $(1,2)$, $(2,1)$, $(L,1)$ and $(1,L)$. The model assigns probabilities given by the Boltzmann law $P_\mathrm{Ising}(\mathbf v)\propto e^{-\beta E(\mathbf{v})}$ to configurations $\mathbf{v}$, where $\beta$ is the inverse temperature; we hereafter denote the average over $P$ by $\langle\cdot\rangle$. In the infinite $L$ limit, the model undergoes a phase transition from a paramagnetic phase ($\beta< \beta_c$) in which the magnetization
\begin{equation}
m=\left\langle \left|\frac{1}{N}\sum_i v_i\right|\right\rangle
\end{equation}
vanishes, to a ferromagnetic phase ($\beta> \beta_c$) in which $m>0$ \cite{baxter2016exactly}. The transition occurs at a critical inverse temperature $\beta_c \approx 0.44$, computed exactly by L. Onsager \cite{onsager1944crystal}, see Figure \ref{fig:ising}.

\subsection{Sampling the Ising model at equilibrium}

We start by generating up to $10^6$ samples from the Ising model through Monte Carlo (MC) simulations, at different inverse temperatures in the range $0.35 \le \beta \le 0.5$. To quickly equilibrate at all temperatures the MC chain includes both local Metropolis updates and global Wolff cluster moves, known to be efficient to sample the model near $\beta_c$ \cite{newman1999monte}; details about the implementation can be found in SM Appendix \ref{SI:sec:impl}. The magnetization $M$ and the heat capacity 
\begin{equation}
C= \frac{\beta^2}{N}\left(\langle E^2\rangle-\langle E \rangle^2\right)
\end{equation}
are shown as functions of the inverse temperature in Figure \ref{fig:ising}A for two system sizes, $L=32$ and $L=64$. Additional observables, such as the susceptibility 
\begin{equation}
\chi = \frac{\beta}{N}\left[\left\langle \left(\sum_i v_i\right)^2\right\rangle - \left\langle \left|\sum_i v_i \right|\right\rangle^2\right]
\end{equation}
and the correlation length are reported in SM Fig. \ref{SI:fig:ising_obs}. A peak in the heat capacity (and in the susceptibility) signals the cross-over between the two phases, when  $\beta$ gets close to $\beta_c$, with a shift that vanishes with increasing $L$ as predicted by finite size-effects theory.

\subsection{Learning with standard RBM}

We then use the MC samples as training data for an unconstrained RBM, with visible units taking $\pm 1$ values. To enforce  the global sign symmetry of the energy, i.e. $E(-\mathbf{v})=E(\mathbf{v})$, see  Eq.~\eqref{eq:ising}, we choose  hidden units $h_\mu=\pm 1$  (instead of $0,1$ as in the MNIST case) and vanishing biases on the both visible ($g_i=0$) and hidden ($\theta_\mu=0$) units. The training phase thus consists  in inferring the RBM weights $w_{i\mu}$ only.

We verify that the log-likelihood $\log P(\mathbf{v} )$ of test MC data estimated with the trained RBM correlate with the Ising energy $E(\mathbf{v})$ (SM Fig. \ref{SI:fig:ising_energy_rbm}). The weights learned by the RBM  exhibit localization patterns (see Figure \ref{fig:ising}G) at low temperatures, in agreement with observations reported in previous works on the 1-dimensional Ising model \cite{harsh2020place}.

We generate samples from these RBMs learnt at different $\beta$'s using alternate Gibbs sampling, and evaluate the magnetization, heat capacity, and susceptibility. Results are in agreement with the same quantities computed from samples of the Ising model distribution, see Figure \ref{fig:ising}B. This observation is consistent with literature \cite{yoshioka2019transforming,yevick2021accuracy,cossu2019machine}, where RBMs were shown to be able to  accurately  fit statistical physics models such as the Ising model.

\subsection{Partial erasure of information with fully constrained RBM}

We hereafter  choose that the label $u=\pm 1$ associated to a configuration of spins $\bf v$ is the sign of its magnetization,
\begin{equation}
u(\mathbf{v})= \textrm{sign} \left(\sum_i v_i \right) \ .
\end{equation}

\subsubsection{Linear constraints}

By symmetry, the vector $\mathbf{q}^{(1)}$ in Eq.~\eqref{eq:Worth} has uniform components $q_i^{(1)}=q^{(1)}$ due to the translation invariance of the lattice resulting from periodic boundary conditions. Imposing the linear constraint in Eq.~(\ref{eq:Worth}) thus amounts to demanding that all weight vectors sum up to zero, i.e.  $\sum_i w_{i\mu} = 0$ for $\mu=1,...,M$.

We then train a RBM on the MC data under these constraints. The log-likelihoods of test Ising configurations are poorly correlated with the Ising model energies in Eq.~\eqref{eq:ising}, see SM Fig.~\ref{SI:fig:ising_energy_adv}. In addition, RBM generated samples show no magnetization at any inverse temperature, even for $\beta > \beta_c$, see Figure~\ref{fig:ising}C. Surprisingly, however, other observables such as the heat capacity (Figure~\ref{fig:ising}C) or the susceptibility (SM Fig.~\ref{SI:fig:ising_obs}) exhibit a peak at the cross-over inverse temperature. We conclude that the constrained RBM generated spin configurations with zero first moment, but a substantial part of higher-order correlations is still correctly captured and reproduced. We will come back on the interpretation of the effective energy corresponding to this fully constrained RBM in Section~\ref{sec:inter}.

\subsubsection{Quadratic constraints}

We next apply second-order constraints \eqref{eq:Worth2} to all weight vectors of the RBM. Due to the global invariance of the Ising energy under spin reversal $\mathbf{q}^{(2)} = 0$ abiding to definition \eqref{eq:Q2}. However, the reversal symmetry is lifted in the presence of an arbitrary small uniform external field $\Delta$, \emph{i.e.} $E(\mathbf{v})\to E(\mathbf{v})- \Delta \sum_i v_i$. We show in SM Appendix \ref{SI:sec:Q2} that, to first order in $\Delta$, $\mathbf{q}^{(2)}\simeq \frac 12 \,\Delta\, \mathbf{Q}^{(2)}$ with
\begin{equation}
{Q}^{(2)}_{i,j}= \left\langle \left|\sum_k v_k\right| v_i v_j \right\rangle_{\cal D} - \left\langle\left|\sum_k v_k\right|\right\rangle_{\cal D} \left\langle v_i v_j \right\rangle_{\cal D}\ .
\end{equation}
The tensor $\mathbf{Q}^{(2)}$ can be estimated numerically, and used to constrain the weight vectors through Eq.~\eqref{eq:Worth2}.

RBM learnt under these quadratic constraints generate spin configurations with zero magnetization, as in the case of linear constraints, see Figure~\ref{fig:ising}E. Remarkably, the specific heat and the susceptibility show no peak as $\beta$ is varied, suggesting that quadratic constraints on the weights have much stronger impact on the distribution of spin configurations. The heat capacity in particular, has a mild monotonic increasing tendency with $\beta$, attaining similar values to the original model at low and high temperatures.

However, inference of the magnetization sign is still possible from the hidden representation, although with degraded performance. For each inverse temperature, we trained classifiers of varying complexity, and measured their performance in predicting the labels. The resulting AUC are shown in Figure \ref{fig:ising}F, and are above chance level ($.5$) at high $\beta$. This indicates that higher-order correlations presumably present in the inputs of full-constrained RBM (such as the Binder cumulant \cite{selke2006critical}) can be used for predicting labels with some success; we will encounter a similar situation in the MNIST0/1 case.

\subsection{Manipulating representations and spin configurations with partially constrained RBM} \label{sec:isingrel}

We now apply constraint \eqref{eq:Worth} on all but one hidden unit when training the RBM on the Ising data. The released hidden unit, hereafter referred to as $h^\ast$, learns a weight vector which is approximately proportional to $\mathbf{q}^{(1)}$, that is, the weights connecting to $h^\ast$ are uniform over the visible layer, with a common value hereafter referred to as $w^\ast$. The resulting RBM then has one hidden unit that controls the sign of the magnetization of the generated samples, while the remaining hidden units capture local correlated patterns of neighboring spins. Indeed, the constrained weights display localized patterns similar to those of unconstrained RBM (Figure \ref{fig:ising}E). In addition, the RBM reproduces the behavior of all observables as the inverse temperature is varied (Figures \ref{fig:ising}E and SM Fig. \ref{SI:fig:ising_obs}). These results strongly suggest that the constraints on (all but one) weight vectors applied during learning do not impair the ability to fit the data, but only serve to reorganize the latent representations. In addition to \eqref{eq:Worth}, we can also impose constraints \eqref{eq:Worth2} on all but one hidden units, with similar results as those reported (not shown).

\subsection{Effective energy resulting from constraints} \label{sec:inter}

A heuristic argument allows us to better understand the nature of the distribution expressed by the fully-constrained RBM (linear case), in particular, why generated configurations have zero magnetization while encoding non-trivial spin-spin correlations (Figure~\ref{fig:ising}C).

Let us first notice that the general expression for the log-probability of a visible configuration $\mathbf v$ in the RBM reads, due to the absence of biases on the units,
\begin{equation}\label{eq:PrbmIsing}
\log P_\mathrm{RBM}(\mathbf{v}) = \sum_{\mu=1}^M \log \cosh \left(\sum_i w_{i\mu}v_i\right) \ ,
\end{equation}
up to an irrelevant additive constant. This formula applies in particular to the released RBM of Section~\ref{sec:isingrel}, in which all but one hidden unit, say, $\mu=1$, are constrained to satisfy Eq.~(\ref{eq:Worth}). Based on our previous finding that $w_{i,1}\simeq w^\ast$, we obtain
\begin{multline}\label{eq:PrbmIsingConstr}
\log P_{\mathrm{rel.}}(\mathbf v) \simeq \sum_{\mu=2}^M \log\cosh\left(\sum_i w_{i\mu}v_i\right) + w^\ast\left| \sum_i v_i \right|,
\end{multline}
where we have approximated $\log\cosh x\simeq|x|$ for large arguments $x$ and have again neglected additive constants. Based on Eq.~(\ref{eq:PrbmIsingConstr}) we may proceed in two steps. First, as we empirically find that the released RBM is a good approximation of the ground-truth Ising distribution, we approximate $\log P_{\mathrm{rel.}}$ with $\log P_\mathrm{Ising}$. Second, the first term on the right-hand side of Eq.~(\ref{eq:PrbmIsingConstr}) expresses the log-probability of ${\bf v}$ computed by a RBM with weight vectors constrained to be orthogonal to $\mathbf{q}^{(1)}$, and can thus be identified with $\log P_{\mathrm{constr.}}$. We conclude, using Eq.~(\ref{eq:ising}),  that the effective energy function on the spin configuration encoded by the fully constrained RBM is approximately equal to
\begin{equation}\label{eq:IsingAbsM}
E_{\mathrm{constr.}} (\mathbf{v}) \simeq -\sum_{(ij)}v_i v_j + \frac{w^\ast}\beta \, \left| \sum_i v_i \right|\ .
\end{equation}
The effects of the constraints on the weights is to introduce a $L_1$-like penalty against magnetized configurations opposing the Ising energy, which tends to align spins. This explains both the disappearance of magnetization and the remanent correlations observed in Figure~\ref{fig:ising}C.

We can also estimate the value of $w^\ast$ selected through learning of the fully-constrained RBM, with an heuristic argument. Consider a typical configuration of the Ising model at low temperature, i.e. in the ferromagnetic regime corresponding to magnetization $m^* \ne 0$. The effective field acting on spin, say, $i$, reads, according to Eq.~\eqref{eq:IsingAbsM}, 
\begin{equation}
g_i^\textrm{eff} = \sum _{j\in {\cal N}_i} v_j - \frac{w^\ast}\beta \textrm{sign} \left( m^*\right) \ ,
\end{equation}
where ${\cal N}_i$ refers to the neighbourhood of spin $i$ on the squared grid. Taking the average over the spin $i$ we obtain the mean value of the effective field
\begin{equation}
\langle g^\textrm{eff} \rangle  = z\, m^* - \frac{w^\ast}\beta \textrm{sign} \left( m^*\right) \ ,
\end{equation}
where $z=4$ is the coordination number on the grid. We conclude that the effective field vanishes  when
\begin{equation}\label{eq:wstarpred}
w^\ast = \beta\, z\, | m^*|\ .
\end{equation}
The above expression gives the  minimal strength of the $L_1$ penalty capable of counterbalancing the local interactions tending to magnetize spins. It is expected to vanish in the paramagnetic regime. Higher values are disfavored during the RBM training phase as they would assign higher energies $E_\mathrm{constr.}$ in Eq.~\eqref{eq:IsingAbsM} to typical magnetized Ising configurations, and thus lower likelihoods.

We compare the heuristic estimate for $w^*$ provided by Eq.~\eqref{eq:wstarpred} to the numerical results for $w^\ast$ obtained from training partially-constrained RBM on 2D-Ising data in Figure \ref{fig:ising}I. Despite the presence of finite-size effects, we observe a good agreement between Eq.~\eqref{eq:wstarpred} and the simulation results.

\section{Application to MNIST handwritten digit images}

We next considered the MNIST handwritten digit dataset \cite{deng2012mnist}. Pixel intensities are binarized by thresholding at $0.5$. For simplicity, we start by considering the subset of images containing only digits 0 and 1 (MNIST0/1), for which the class label $u$ is binary.

\subsection{Learning with standard RBM}

\begin{figure*}
\includegraphics[width=0.75\textwidth]{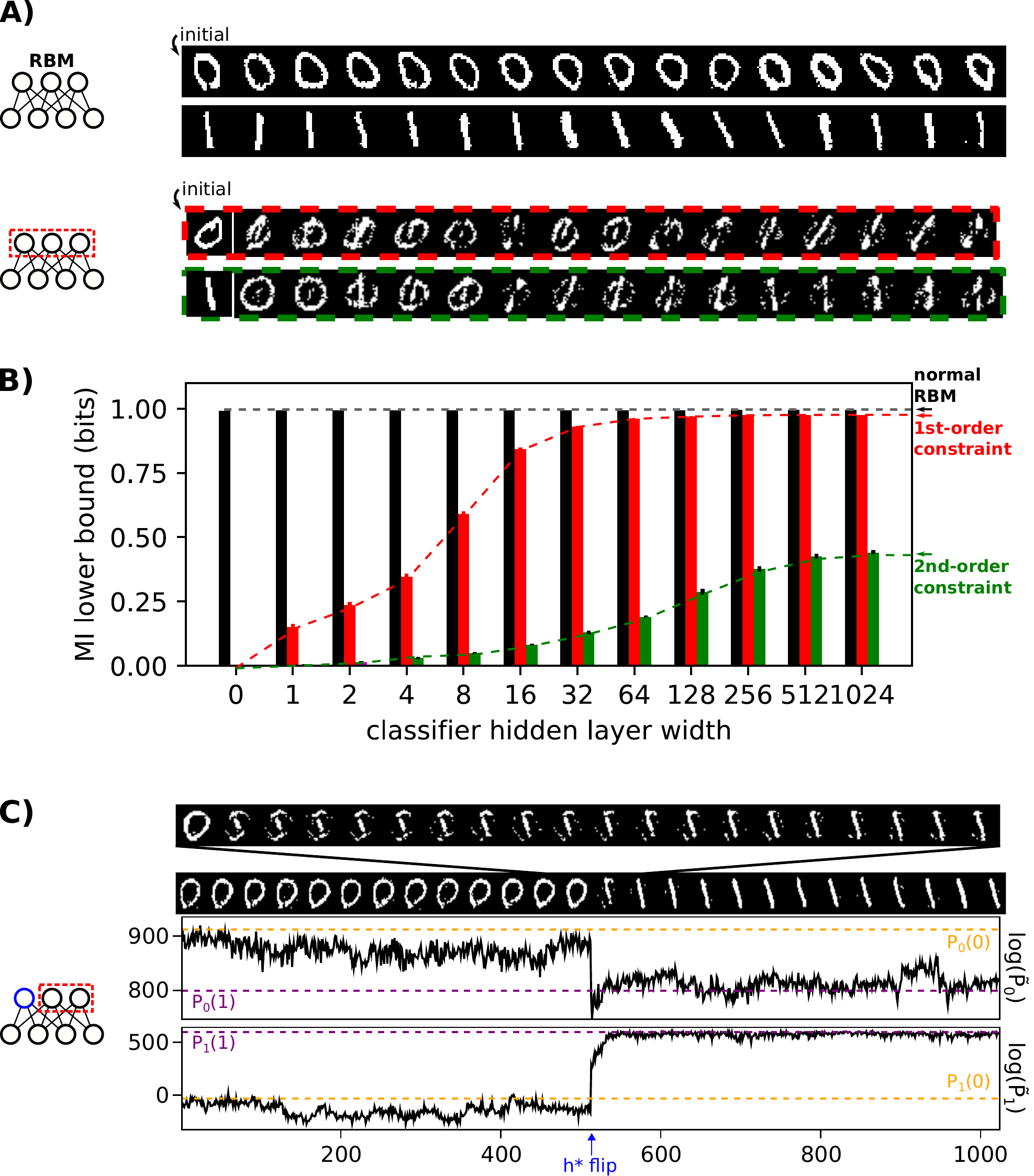}
\caption{\label{fig:mnist}\textbf{Manipulating representations of RBM trained on MNIST0/1.} \textbf{A)} Samples generated by RBM  initialized with a data image (0 or 1). Top two rows:  standard (unconstrained) RBM; Bottom two rows show samples from RBM trained with linear (red dashed) and quadratic (green dashed) constraints. In both cases, a Markov chain was generated by Gibbs sampling (starting from a 0 or a 1 data digit), and images were saved every 64 steps, until reaching a total of 16 samples as shown. \textbf{B)} Lower bound  $\mathcal S_\mathrm{label} + \mathcal L_\mathrm{class}$ to the mutual information between inputs and labels, see Eq.~\eqref{eq:Gibbs}, vs. classifier width. The bounds to MI is measured in bits and shown in discontinuous lines. Colors correspond to the different RBM models. Black: standard/unconstrained; Red: fully constrained with linear constraints, see Eq.~\eqref{eq:Worth}); Green:  fully constrained with quadratic constraints, see Eq.~\eqref{eq:Worth2}. \textbf{C)} Samples from RBM trained with 1st-order constraint acting on all but one hidden unit, which is flipped at the middle of the MC chain (blue arrow). Starting from a 0 data digit, samples were saved every 64 Gibbs steps. Top panel shows a zoomed view of the transition, with images every 3 steps instead. The lower panels show the logarithm of the unnormalized probability, $\ln\tilde{P}(\mathbf v) = \ln\left(\sum_\mathbf{h}\mathrm{e}^{-E(\mathbf v, \mathbf h)}\right)$ of generated digits by constrained RBMs, evaluated on RBMs trained only on 0's (RBM0) or 1's (RBM1). Purple and orange dashed lines correspond to the average $\ln\tilde{P}(\mathbf v)$ of data digits 0 and 1.}
\end{figure*}

We trained a standard RBM on MNIST0/1, with $M=400$ binary hidden units and $N=28\times28$ visible units, through maximization of the log-likelihood \eqref{eq:Lrbm} (see SM Appendix~\ref{SI:sec:RBM-architecture} for further details). Figure \ref{fig:mnist}A shows Markov chains of samples derived from Gibbs sampling of the resulting models. The machine generates strings of 0's or 1's, depending on the initial condition, with very rare transitions between these classes. Note that the absence of transitions from 0 to 1 (or vice versa) is likely due to the strong dissimilarities between these two digits in configuration space and the lack of low energy configurations connecting them; training the RBM on all digits tends to connect these two modes and to increase the frequency of observed transitions.

To quantify the information content in the inputs about the labels (digit value) we estimated the mutual information $\mathrm{MI}(u,\mathbf I(\mathbf v))$. While computing $\mathrm{MI}$ is very hard, a tractable lower bound can be obtained through the Gibbs variational inequality \cite{cover1999elements},
\begin{align}\label{eq:Gibbs}
\mathrm{MI}(u,\mathbf I(\mathbf v)) &\ge \sum_{u, \mathbf v} P_\mathcal{D}(u, \mathbf v) \ln\left(\frac{P_\mathrm{class}(u| \mathbf I(\mathbf v))}{P_\mathcal{D}(u)}\right) \nonumber \\
&= \mathcal{S}_\mathrm{label} + \mathcal{L}_\mathrm{class}
\end{align}
where $P_\mathcal{D}(u,\mathbf v)$ is the empirical distribution of labeled data, and $P_\mathrm{class}(u| \mathbf I(\mathbf v))$ is any conditional distribution, implemented here by a classifier attempting to predict the label. By rearranging terms, this equals the entropy of labels in the data ($\mathcal S_\mathrm{label}$) plus the log-likelihood of the classifier averaged over held out data ($\mathcal{L}_\mathrm{class}$).

This lower bound to MI is shown in Figure~\ref{fig:mnist}B (black bars) for classifiers of increasing complexity, corresponding to two-layer networks with a hidden layer of increasing width (horizontal axis in the figure), see SM Appendix \ref{SI:sec:classifiers} for details about the architecture and training of these classifiers. The simplest network is a linear classifier (perceptron, width $=0$), and already achieves nearly perfect prediction accuracy. In addition the weights of this optimal linear classifier are distributed over all hidden units, showing that information about the label is distributed across the latent representation. As the width of the classifier increases the lower bound to $\mathrm{MI}$ saturates at a value close to 1 bit, the maximum possible for two label classes, indicating that the RBM inputs capture maximum label information. We emphasize that the RBM has no direct access to the label values during training. 

\subsection{Partial erasure of information with fully constrained RBM}

We next train RBM with constrained applying on the weigth vectors attached to all hidden units.

\subsubsection{Linear constraints}

Figure \ref{fig:mnist}A (bottom, red) shows typical configurations  generated by RBM trained with constraints (\ref{eq:Worth}). As expected these configurations tend to be blurred mixtures of 0's and 1's.

A simple linear discriminator looking at the inputs to the hidden units is unable to predict the labels of these digits, in agreement with the adversarial interpretation of Eq.~\eqref{eq:Worth}. However, information about the digit class is still present in the RBM representations through higher-order correlations. Sufficiently complex classifiers are able to recover the label of data digits with maximum accuracy (Figure \ref{fig:mnist}B), and give lower bounds to $\mathrm{MI}$ close to unity. This result shows that, while condition \eqref{eq:Worth} is not sufficient to erase the label information from the representation extracted by the RBM, it does make retrieval of this information more difficult.

\subsubsection{Quadratic constraints}

Imposing the stronger, quadratic constraints in Eq.~\eqref{eq:Worth2} results in sample of worse quality, see green row in Figure \ref{fig:mnist}A, bottom. Figure \ref{fig:mnist}B shows that simple classifiers trained are unable to predict the labels from the inputs. Interestingly, more complex classifiers achieve a moderate non-zero prediction accuracy, but provide substantially lower estimates of the mutual information than when trained on linearly-constrained RBMs (compare green and red bars). The lower bounds to $\mathrm{MI}$ seems to saturate to a value well below 1 as the classifier widths increase. These results indicate that quadratic constraints erase a sizable part of the information about the labels.

\subsubsection{On the generative power of the fully constrained RBM} \label{sec:mnist_bw_generative}

Configurations sampled from the fully constrained RBMs in Figure \ref{fig:mnist}A (bottom) tend to be blurred mixtures of digits (0 and 1). In this case, the data are in fact a mixture of two widely separated distributions, associated to 0's and 1's. This is reminiscent of configurations of opposite magnetization in the Ising model at low temperature in Section \ref{sec:isingrel}, and the sampled blurred digits are in analogy to the `intermediate' configurations of zero magnetization that the fully constrained RBM samples in that case (Figure \ref{fig:ising}C top). We however saw that in the Ising model configurations sampled from the fully constrained RBM still carry relevant information in higher-order statistics, \emph{e.g.} as shown by the behavior of the heat capacity, Figure~\ref{fig:ising}C bottom.

To illustrate how fully constrained RBM can generate samples with meaningful information present in higher order statistics in the setting of handwritten digit images, we consider the following simple numerical experiment. For each 0 digit from MNIST, we produce an additional image where pixel colors were flipped (producing black zeros in white background), and define a binary label encoding the background color. We then train a fully constrained RBM on this data. Generated samples are shown in SM Fig.~\ref{SI:fig:mnist_black_white}. The fully constrained RBM generates recognizable 0 digits embedded in noisy backgrounds, where local patches in the digit strokes clearly tend to share the same color, indicating that the overall structure of the digit is preserved through correlations.

\subsection{Manipulating representations and digits with partially constrained {RBM}}

We now impose linear constraints \eqref{eq:Worth} to all but one (blue) hidden units. As stated in objective B, our intention is to promote concentration of label information on this released unit, see Figure \ref{fig:architectures}B. After learning the released weight vector is similar (up to a global scale factor) to  vector $\mathbf q^{(1)}$ (SM Fig. \ref{SI:fig:q01}), a direction forbidden to the other hidden units. Hence the average value of the unit conditioned to a visible configuration (digit) is an excellent predictor of the corresponding label.

Samples generated by the RBM are nice-looking 0's or 1's, in a manner consistent with the state of the released hidden unit. Furthermore, manipulating the state of this hidden unit i.e. freezing it to 0 or 1 helps generating samples with desired labels. We show in Figure~\ref{fig:mnist}D numerical experiments illustrating the effects of such manipulations. We initialize the RBM with a digit (0 in Figure~\ref{fig:mnist}D) extracted from the MNIST0/1 data set, and samples new configurations through alternate Gibbs samplings. As with standard RBM the samples vary over time, but the digit class remain unchanged. We then  flip the state of the hidden unit (middle of Figure~\ref{fig:mnist}D). As a consequence, the resulting visible configuration converges to the other digit class, after some short transient (see top part of panel).

To evaluate the quality of the generated digits, we train two RBMs only on 0's or 1's, respectively, and evaluate the log-likelihoods of the generated digits on two standard RBMs, one trained with 0 digits only, and another trained on 1's only. These two machines provide expected reference scores for 0's and 1's. Figure~\ref{fig:mnist}E shows  that the generated digits are of good quality, with log-likelihood values comparable to the ones of the data.

\begin{figure}
\includegraphics[width=0.9\linewidth]{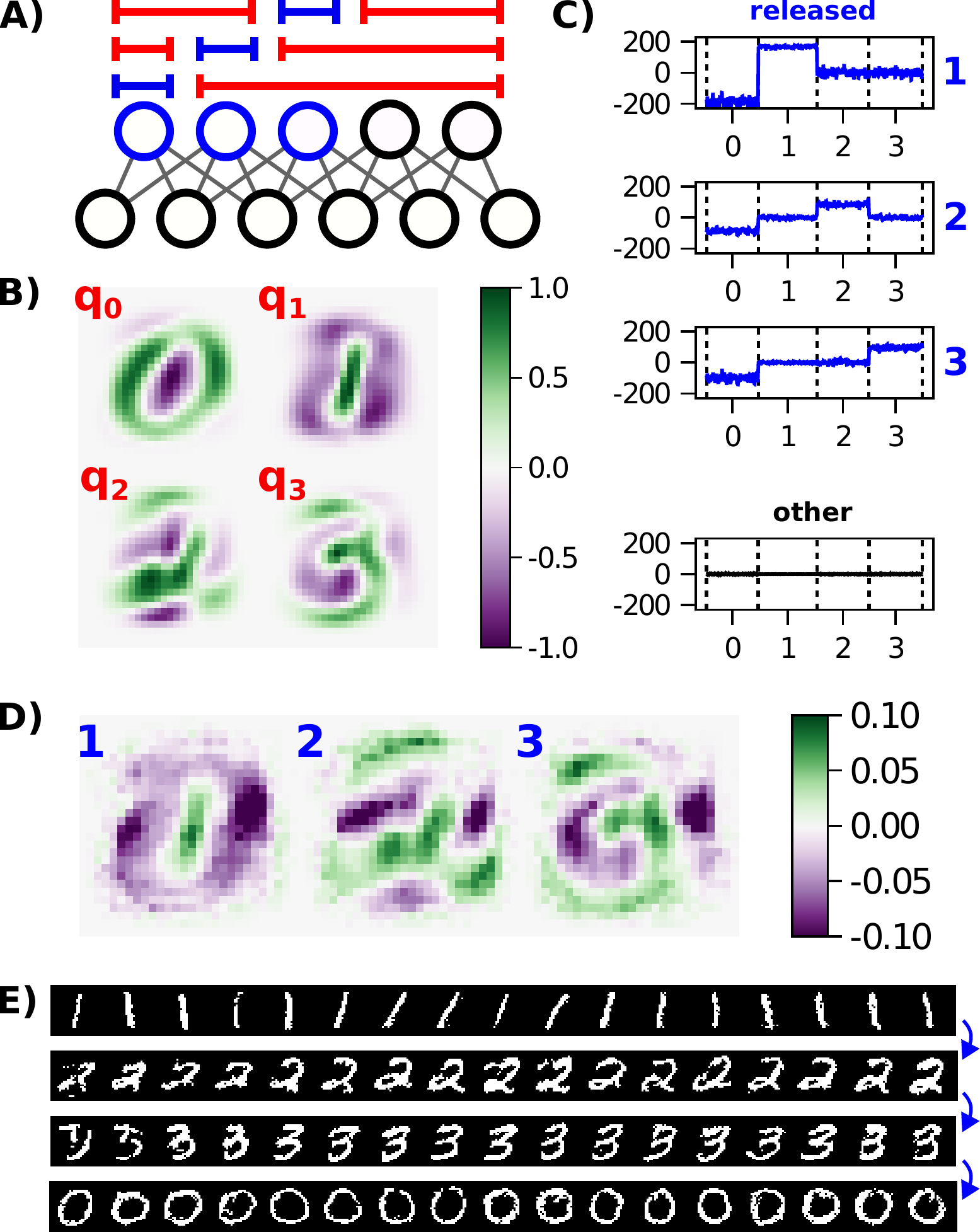}
\caption{\label{fig:mnist-multi}\textbf{Manipulating representations of RBM trained on MNIST0/1/2/3.} \textbf{A)} Sketch of the contraints applied to hidden unit weights in the case of multiple classes, here, $D=4$. \textbf{B)} Vectors $\mathbf{q}_d^{(1)}$ for digit classes 0, 1, 2 and 3, see Eq.~(\ref{eq:WorthMulti}). \textbf{C)} Inputs received by the three released  hidden units (in blue on panel A), when the 6,000 digit images in classes 0, 1, 2 and 3 are presented ($x$-axis). In the fourth, bottom panel, inputs received by a random hidden unit from the constrained group (black) are shown. \textbf{D)} Weights $w_{i\mu}$ learned weights by the released hidden units $\mu=1,2,3$. \textbf{E)} Samples generated from this machine by Gibbs sampling (images shown are taken every 64 steps). First (top row), released unit 1 is active, while the other two are inactive. Then, we activate unit 2 (second row) while inactivating unit 1 (blue arrow), and similarly for 3 (third row). In the last row, all three units are inactive.}
\end{figure}

\subsection{Case of more than two digits}

While we have focused  on the case of binary labels so far, our approach can be easily adapted to  more than two classes.  We consider the case of $D$ classes, and use one-hot encoding for the labels, i.e. introduce $D$ labels $u_d$, one for each class $d=0,1,...,D-1$. Due to one-hot encoding prescription each data configuration $\mathbf{v}$  is such that $D-1$ labels $u_d(\mathbf{v})$ vanish and one is equal to 1.

Analogously to \eqref{eq:Worth}, we define $D$ vectors (in the $N-$dimensional space of data)
\begin{equation}\label{eq:WorthMulti}
\mathbf{q}_d^{(1)}=\langle u_d (\mathbf{v})\, \mathbf{v} \rangle_{\cal D} - \langle u_d(\mathbf{v})\rangle_{\cal D} \langle \mathbf{v}\rangle _{\cal D}\ .
\end{equation}
We then generalize Eq.~\eqref{eq:Worth} to multiple classes by imposing that weight vectors be orthogonal to all $\mathbf{q}_d^{(1)}$, with $d=1,...D$. It is easy to check that the $D$ vectors in Eq.~\eqref{eq:WorthMulti} sum up to zero, a consequence of the one-hot encoding scheme. We therefore consider only the last $D - 1$ vectors, with indices $d=1,2,...D-1$ to obtain linearly independent constraints acting on the weights.

In practice the  constraints $\mathbf{w}^\mu\perp \mathbf{q}_d^{(1)}$ are enforced through the architecture shown in Figure \ref{fig:mnist-multi}A, in which a set of $D-1$ hidden units $h_d$ are released, each with respect to a {\em single} $\mathbf{q}_d^{(1)}$ and constrained to be orthogonal to all the other $D-1$ vectors. In this way, when activating one of these hidden units, say, $\mu$, the corresponding digit $d=\mu$, is expected to be sampled on the visible layer. When all first $D-1$ hidden units are silent, digit $d=0$ is expected to be sampled.

We illustrate this approach in the case of $D=4$ digits, with RBMs trained from MNIST0/1/2/3. The vectors $\mathbf{q}_d^{(1)}$ in Eq.~\eqref{eq:WorthMulti} are shown in Figure \ref{fig:mnist-multi}B. After training the RBM under the orthogonality constraints, the released hidden units $\mu=1,2,3$  are strongly activated by, respectively, digits $d=1,2,3$. In Figure \ref{fig:mnist-multi}C we show the average inputs to these hidden units when data digits are presented on the visible layer of the RBM; the corresponding  weight vectors are depicted in \ref{fig:mnist-multi}D. When digit 0 is present on the visible layer, the three hidden units are silent. Other hidden units are weakly activated by the different digits and capture information (small stretches, local constrast) crucial for generating high-quality digits but not directly related to their identity, see panel ``other'' in \ref{fig:mnist-multi}C.

We next manipulate these units to generate digits out of one of the four classes. The outcome is shown in Figure~\ref{fig:mnist-multi}E, where the  Markov chain is initialized with a 1 digit from the MNIST data, and the first released hidden unit ($\mu=1$) is on, while the other two ($\mu=2,3$) are off. Sampling the RBM in this condition generates a string of 1's as illustrated in the figure. Turning this unit off and turning the second $\mu=2$ on now produces a transition in the visible layer, and generates digits 2. Iterating this procedure, we generate 3's, and finally 0's by turning off all released hidden units (last row in Figure~\ref{fig:mnist-multi}E).

\section{Application to protein sequences with taxonomy annotations} \label{sec:pfam}

A protein family is a group of proteins that share a common evolutionary origin, reflected by their related functions and similarities in sequence or structure \cite{el2019pfam}. Protein families are often arranged into hierarchies, with proteins that share a common ancestor subdivided into smaller, more closely related groups. In recent years, RBMs have been successfully applied to extract structural, functional, and evolutionary information from the sequences attached to a protein family \cite{tubiana2019learning,shimagaki2019selection,bravi2021rbm}. Our aim here is to use partially constrained RBM to  disentangle  the  label defining the taxonomic domain (eukaryota or bacteria) a protein sequence belong to, and manipulate the domain-determining hidden unit to drive a continuous transition, or morphing, between one taxonomic domain to  the other during  sampling of artificial sequences.

\begin{figure*}
\includegraphics[width=\textwidth]{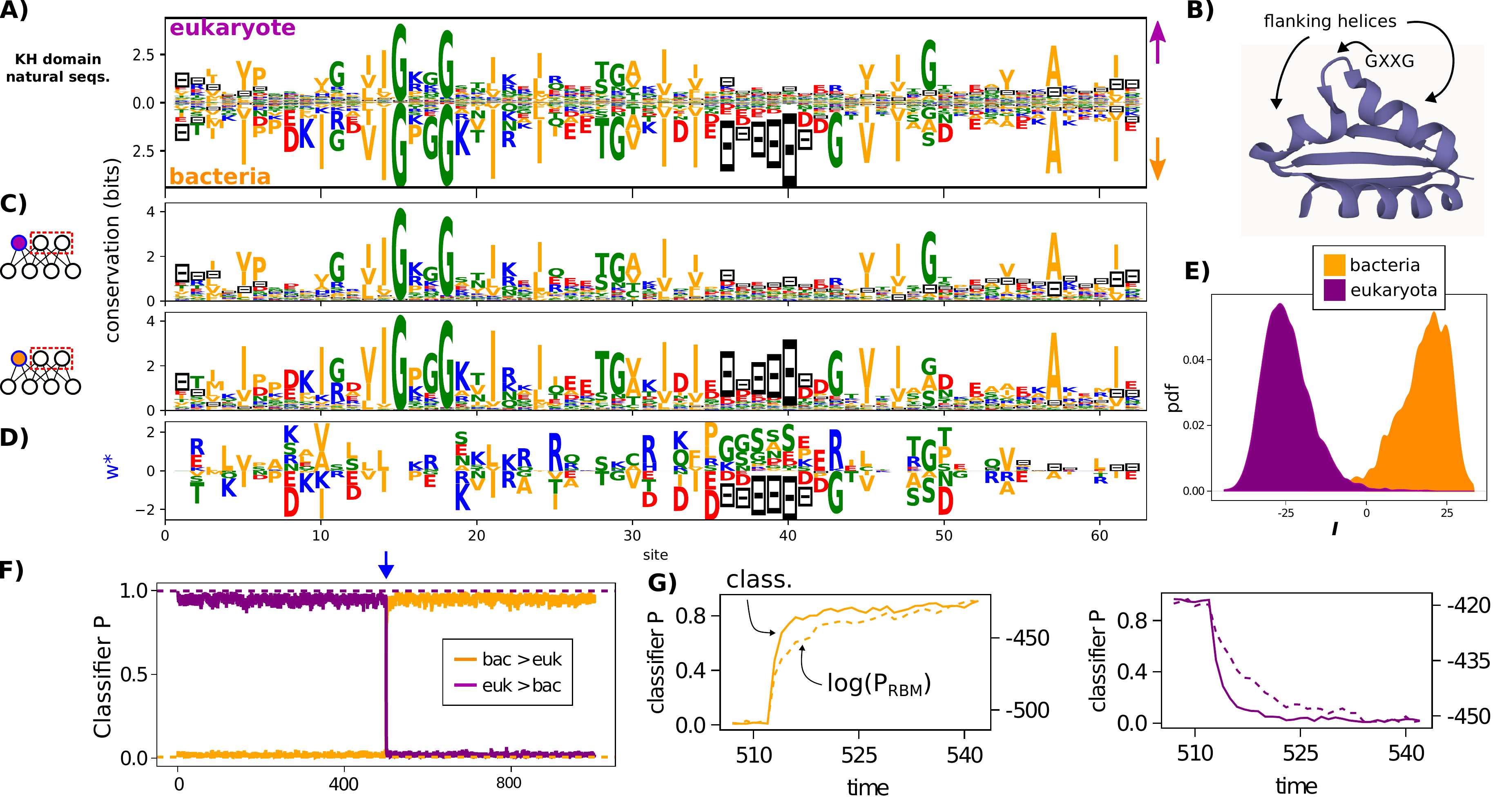}
\caption{\label{fig:pfam}\textbf{Taxonomy of protein families} \textbf{A)} Sequence logos of eukaryotic (purple, above) and bacterial (orange, below) sequences from the PF00013 protein family. We use the following color code: green for polar residues, blue for basic, red for acidic, and orange for hydrophobic. Gaps are shown in black. \textbf{B)} Ribbon structure of KH domain, showing locations of Gly--Gly loop and flanking helices. Image prepared with {Mol* Viewer} \cite{sehnal2021mol}. \textbf{C)} Sequence logos of 10,0000 generated sequences, when the released hidden unit is set to 1 (top) or 0 (below). To ensure that sampling is equilibrated, we track the average and standard deviation of the energy of the samples in time, and saw that these statistics were essentially constant after $\sim 200$ steps, suggesting that samples can be collected every 5000 steps. \textbf{D)} Weights of the released hidden unit. \textbf{E)} Inputs received by the released hidden unit when presented with sequences from the two classes. \textbf{F)} Markov chain, started from bacterial (orange) or eukaryotic (purple) sequences from the data. The panel shows the probability of being eukaryotic vs. bacterial sequence in a perceptron classifier. Discontinuous lines are the average value for data sequences of each class. A total of $1024$ Gibbs sampling steps were taken, and the flip of $h^\ast$ occurs at step $512$ (blue arrow). \textbf{G)} Zoomed view near the transition, showing also the log of the unnormalized marginal ($\log\tilde{P}_\mathrm{RBM}(\mathbf v)$) of sampled sequences (right axis), evaluated on an RBM trained on the full family.}
\end{figure*}

\subsection{The K Homology domain}

To illustrate the application of our model, we selected the K Homology (KH) module, a common nucleic acid binding motif in proteins found in multiple species, both eukaryotic and prokaryotic. Structurally, KH domains adopt a globular fold, constituted by three alpha-helices and three beta sheets  \cite{grishin2001kh,lunde2007rna,valverde2008structure},as shown in Fig.\ref{fig:pfam}A. A central feature of the KH domain is the presence of a signature IsoGlyXXGly motif (see Figure \ref{fig:pfam}A \& B), conserved across the entire family, which in cooperation with flanking helices, forms a cleft where recognition of four nucleotides in single-stranded DNA or RNA chains occurs \cite{valverde2008structure}. Mutations in these highly conserved residues result in loss of function \cite{musco1996three}. In particular, substitution of the moderately conserved isoleucine following the Gly$-$Gly loop (two sites after) by Asn, in a KH  domain locus of the fragile X mental retardation gene in humans, causes fragile X syndrome, a leading heritable cause of mental retardation \cite{kh2002decade}.

We have selected this family in our work as it has  a sufficient number of eukaryotic and bacterial sequences available in the PFAM database \cite{el2019pfam}. The PF00013 family of homologous sequences, includes $\sim 11,000$ bacterial sequences and $\sim 38,000$ eukaryotic sequences of the KH domain. After aligning, removing insertions and retaining only columns with less than 50\% gap (deletions) content, sequences end up having a common length of $L=62$ amino acids. As the taxonomic origin of every sequence can be simply queried through the Uniprot database \cite{uniprot2020}, we define label $u=0$ and $1$ for, respectively, bacterial and eukaryotic proteins. To reduce common ancestry bias, sequences are weighted according to their dissimilarity to other members of the same family \cite{cocco2018inverse,morcos2011direct}: the weight assigned to a sequence is proportional to the inverse of the number of sequences in the family with a Hamming distance smaller than 20\% of the sequence length. We also balance the total weights of eukaryotic and bacterial classes, so that both classes have equal weights.

Figure \ref{fig:pfam}A shows the sequence logos of the eukaryotic (top) and bacterial (bottom) sequences in the family after carrying out the above pre-processing steps. Some features are shared across KH domain sequences in both sub-families, such as the well-conserved Gly$-$Gly loop (Figure~\ref{fig:pfam}B). Bacterial sequences have an overall larger content of gaps (deletions) with respect to the consensus alignment, reflecting sequence length differences in the two sub-families.

\subsection{Learning a generative model with standard RBM}

Multiple Sequence alignments are represented using categorical or Potts variables, each site of the alignment having one of 21 possible values (20 amino-acids and one gap value). Gaps are necessary to model sequences of varying lengths \cite{cocco2018inverse}. Using  the one-hot encoding  a configuration $\mathbf{v}$ of the visible layer encodes a sequence over $21\times L$ units, where $L$ is the sequence length.

We first train a RBM on the full alignment, containing both eukaryotic and bacterial sequences, following \cite{tubiana2019learning}. The RBM captures statistics of the sequence alignment, such as conservation profiles at each site. In addition, simple linear classifiers trained on top of the hidden layer of the RBM achieve AUCs of 0.9 in distinguishing between these two classes.

\subsection{Fully constrained RBM are still able to generate foldable sequences}

We then train RBM with constraint \eqref{eq:Worth} acting on all hidden units. The resulting model continues to match the conservation profile of the MSA, and generates diverse sequences. We furthermore validate the foldability of sampled sequences using AlphaFold \cite{mirdita2022colabfold}. As explained in SM Appendix \ref{SI:TM}, we compute the Template Matching score (TM score) of predicted structures of sampled sequences in comparison to the natural sequences, obtaining values $>0.7$ for both the standard RBM and the fully constrained RBM, suggesting that these sequences are able to adopt the expected three-dimensional fold of the family. This result is in agreement with objective A: the model distribution should preserve all the data features unrelated to the label.

\subsection{Changing taxonomic domain with protein design}

We then apply the linear  orthogonality constraint in Eq.~\eqref{eq:Worth} to all but one weight vectors. The weights of the released hidden unit after training are shown in Figure \ref{fig:pfam}D, and capture features that differentiate the two classes. For example, bacterial sequences tend to have deletions (gaps) around positions 35 - 40 of the alignment, indicating that this segment is often absent in bacterial sequences. The learned $w_i^\ast$ reflect this by assigning negative weights to the gap symbol in this region. As a consequence, the distribution of inputs subtended by eukaryotic and bacterial sequences is well separated on this unit (Figure \ref{fig:pfam}E). Conversely, features shared by eukaryotes and bacteria, such as the Gly$-$Gly loop, or the conserved I22, are ignored by $\mathbf{w}^\ast$.

We generate many samples from the RBM distribution, each  conditioned to a fixed state of $h^\ast$, corresponding either to bacterial ($h^\ast=0$) or eukaryotic ($h^\ast=1$) classes. The sequence logos of the two sets of generated sequences are shown in Figure~\ref{fig:pfam}C; they closely match the ones of training data. The list of differences between the logos associated to the two sequence domains include:
\begin{enumerate}
\item The Gly$-$Gly loop is followed by a conserved Lys19 predominantly in bacteria, but not so in eukaryotic sequences.
\item Bacterial sequences conserve a Asp-Lys-Iso motif (positions 8-10) which the RBM with $h^\ast=0$ correctly emits, but not so in the $h^\ast=1$ case. 
\item Besides the two Gly conserved in the entire family, eukaryotic sequences also conserve Gly49, a site which appears less conserved in bacteria which admit also Ala or Ser at this position. The RBM correctly observes these variations.
\item Iso10 is highly conserved in bacteria, while in eukaryotes this site is not conserved, admitting in particular Val, Ala.
\end{enumerate}
These examples suggest that the RBM can sample each sub-family, conditioned on the value of $h^\ast$.

Next, we sample the RBM starting from one bacterial or one eukaryotic sequence in the dataset as initial condition, and with $h^\ast$ set to the value matching the initial condition. After some steps, the value of $h^\ast$ is flipped, and we monitor the dynamical evolution of the generated samples. Figure \ref{fig:pfam}F shows the probability that generated sequences are eukaryotic or bacterial, according to a linear classifier achieving AUC $>0.9$ on held-out test data (see SM Fig. \ref{SI:fig:inputs_classifier}).

Figure \ref{fig:pfam}G shows a magnified view of the classifier probabilities and of the log-likelihood in the vicinity of the hidden-unit switch. We evaluate the log-likelihood of the samples with a RBM trained on the full family (denoted $\log\tilde{P}_\mathrm{RBM}$ in the figure). The class switch, as measured by the classifier score, occurs faster than the relaxation dynamics following the $h^\ast$ flip, as measured by the likelihood. This suggests that the sampled sequences retain other features unrelated to the labeled class, that relax at a slower rate.

\section{Robustness  against the scarcity of labeled data}

\begin{figure*}
\includegraphics[width=\textwidth]{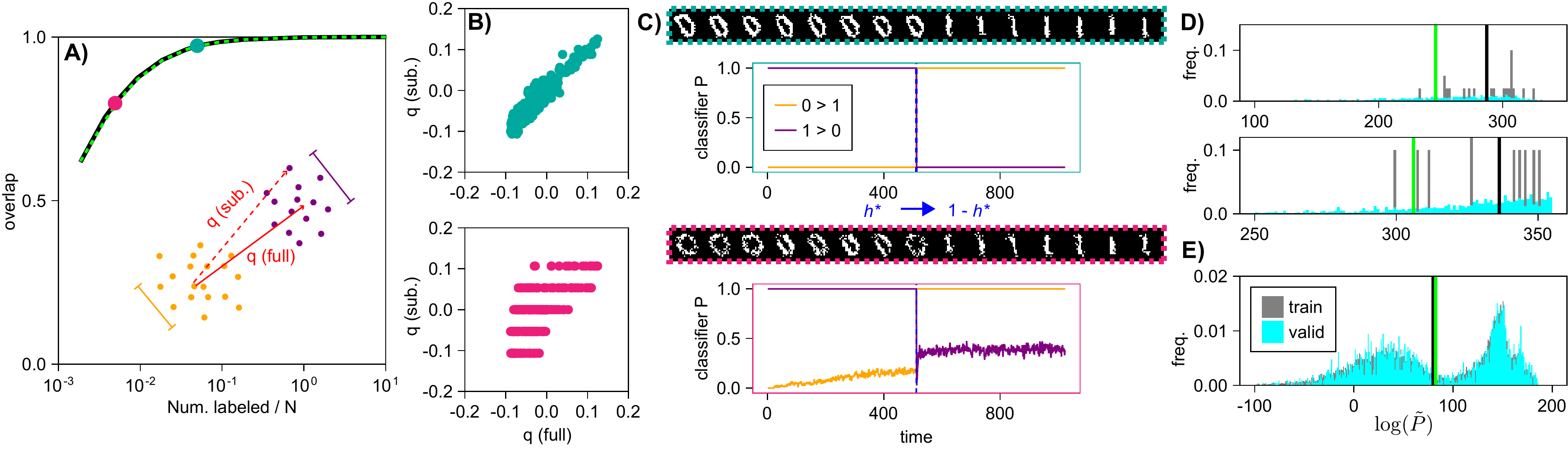}
\caption{\textbf{Semi-supervised training with a sub-sampled labeled dataset.}
\textbf{A)} Overlap \eqref{eq:overlap} between $\mathbf{q}^{(1)}_{sub}$ (computed on a sub-sampled labeled dataset) and $\mathbf{q}^{(1)}_{full}$ (computed on the full dataset), plotted as a function of the number of labeled examples in the sub-sampled dataset divided by the dimension ($28\times28=784$ for MNIST). An average over 100 random realizations of the sub-sampled dataset is taken. The black solid curve shows the empirical result, while the dashed green curve is the theoretical estimate \eqref{eq:overlap-bound}. Inset shows a cartoon diagram of how class separation relates to the overlap, in connection to \eqref{eq:overlap-bound}. \textbf{B)} For the pink and cyan dots of A), we plot an example of the obtained vectors $\mathbf{q}^{(1)}_{sub}$ in comparison to $\mathbf{q}^{(1)}_{full}$. \textbf{C)} Label manipulation, using the sub-sampled $\mathbf{q}^{(1)}_{sub}$ in the two cases. \textbf{D)} Histogram of log-likelihoods of (sub-sampled) training and withheld dataset, for an RBM trained on a subset of 0 (top) or 1 (bottom) digits, corresponding to the labeled datasets used in the cyan dot in the previous panels. The black and green vertical lines indicate the average values. \textbf{E)} Histogram of log-likelihoods of training and withheld dataset of the partially constrained RBM in the cyan setting of the previous panels.}
\label{fig:subsample}
\end{figure*}

One important advantage of our approach is that labeled data is only necessary to estimate the vector $\mathbf{q}^{(1)}$ \eqref{eq:q1} used in the first-order constraint \eqref{eq:Worth}, or the matrix $\mathbf{q}^{(2)}$ \eqref{eq:Q2} in the case of the second-order constraint \eqref{eq:Worth2}. Having determined $\mathbf{q}^{(1)}$ or $\mathbf{q}^{(2)}$, the training of the RBM benefits from additional unlabeled data, and in this regard our model is \emph{semi-supervised}. This property is useful in many real applications, where labels are assigned by humans, are costly to obtain, and thus available for only a small fraction of the data. An example is the KH domain protein sequence dataset considered in Section~\ref{sec:pfam}, where we were able to collect reliable taxonomic labels for only $10\%$ of the sequences. 

To better understand the amount of labelled data needed for our approach to be effective, we conduct further numerical experiments in which the fraction of labeled data is progressively decreased.  We  consider below linear constraint and the MNIST0/1 data for the sake of simplicity. Similar results for the KH domain are reported in SM Fig. \ref{SI:fig:subsample_pfam}.

Since $\mathbf{q}^{(1)}$ becomes trivially zero when there is no data in one of the label classes, we consider balanced sub-sampled labeled datasets with equal numbers of labeled examples in each class. Figure \ref{fig:subsample}A shows the average overlap between vector $\mathbf{q}^{(1)}$ computed on such a sub-sampled labeled dataset (referred to as $\mathbf{q}^{(1)}_{sub}$), and the vector $\mathbf{q}^{(1)}$ computed on the full labeled dataset (denoted by $\mathbf{q}^{(1)}_{full}$), as a function of the number $B$ of labeled examples available, divided by the dimension of the data $N$. Here the overlap is defined by:
\begin{equation} \label{eq:overlap}
 \phi = \frac{\mathbf{q}^{(1)}_{full}\cdot \mathbf{q}^{(1)}_{sub}}{|\mathbf{q}^{(1)}_{full}| \; |\mathbf{q}^{(1)}_{sub}|}
\end{equation}
For each given number of labeled examples, we have considered 100 random realizations of the sub-sampled labeled dataset, and estimate the average of $\phi$ over these realizations. It can be seen from Figure \ref{fig:subsample}A that the overlap never drops below $\approx 0.6$. This result can be understood by considering the separation between the two classes of data (see inset in the Figure). Writing the covariance matrix conditioned on the class label:
\begin{equation}
 C^{(u)}_{ij} = \langle v_i v_j|u \rangle - \langle v_i|u\rangle\langle v_j|u \rangle
\end{equation}
as well as the mean data vector associated to each class,
\begin{equation}
 v^{(u)}_{i} = \langle v_i|u \rangle
\end{equation}
we can derive a simple estimate connected to the average separation between the classes, $\mathbf v^{(0)} - \mathbf v^{(1)}$, and the variances inside each class $\Tr C^{(0)}$, $\Tr C^{(1)}$ (see Appendix SM \ref{SI:subsample-bound} for a derivation), that writes:
\begin{equation} \label{eq:overlap-bound}
 \langle\phi\rangle \approx 
 \left(1 + \frac{1}{B}\frac{\Tr(C^{(0)} + C^{(1)})}{\|\mathbf v^{(0)} - \mathbf v^{(1)}\|^2}\right)^{-1/2}
\end{equation}
where $B$ is the total number of labeled examples, and the average is take over all labeled datasets with $B/2$ examples in each class. Thus, the overlap increases with the separation between the classes ($\mathbf v^{(0)} - \mathbf v^{(1)}$), and decreases if the classes have large variances ($\Tr C^{(0)}$, $\Tr C^{(1)}$), as depicted in the inset cartoon of Figure \ref{fig:subsample}A. The estimate \eqref{eq:overlap-bound} is plotted in Figure \ref{fig:subsample}A and is in excellent agreement with the empirical average overlap.

Figure \ref{fig:subsample}B shows the scatter plots of the components of two example vectors $\mathbf q^{(1)}_{sub}$ computed from sub-sampled labeled data at the pink and cyan points highlighted in Figure \ref{fig:subsample}A, vs. the components of the vector $\mathbf q^{(1)}_{full}$ computed from all labeled data.
Using these vectors $\mathbf q^{(1)}_{sub}$, we then train two RBMs subject to \eqref{eq:Worth} acting on all but one hidden units. Then we attempt to manipulate the sampled data by controlling this released hidden unit. Results are shown in Figure \ref{fig:subsample}C. In both cases the RBMs generate acceptable data and the state of the released hidden unit $h^\ast$ correlates with the sampled digit, even though for the extremely sub-sampled case (pink) digits tend to be more noisy.

To further underline the advantage of our method with respect to supervised learning in a situation with few labeled data, we have trained normal RBMs on the sub-sampled labeled data, specializing on 0 or 1 digits only. As expected for the small amount of training data, these models tend to overfit. This is shown in the histograms of log-likelihood assigned to training and a withheld validation dataset in Figure \ref{fig:subsample}D (top for 0 digits and bottom for 1's). The gap in average log-likelihood of training and validation data (black and green vertical lines, respectively) is quite large, in both cases, indicating overfitting. In contrast, the partially constrained RBM (the same from the cyan dot in the previous panels of the figure) uses both the few labeled data and the large quantity of unlabeled data to avoid overfitting, and we show the log-likelihood histograms for training and validation data in Figure \ref{fig:subsample}E. The agreement between both subsets is excellent, indicating that this model is not overfitting.

In summary, these results provide evidence for the fact that our method is also applicable with limited labeled data.



\section{Estimating the costs of partial erasure and disentanglement}

\begin{figure*}
\includegraphics[width=0.9\textwidth]{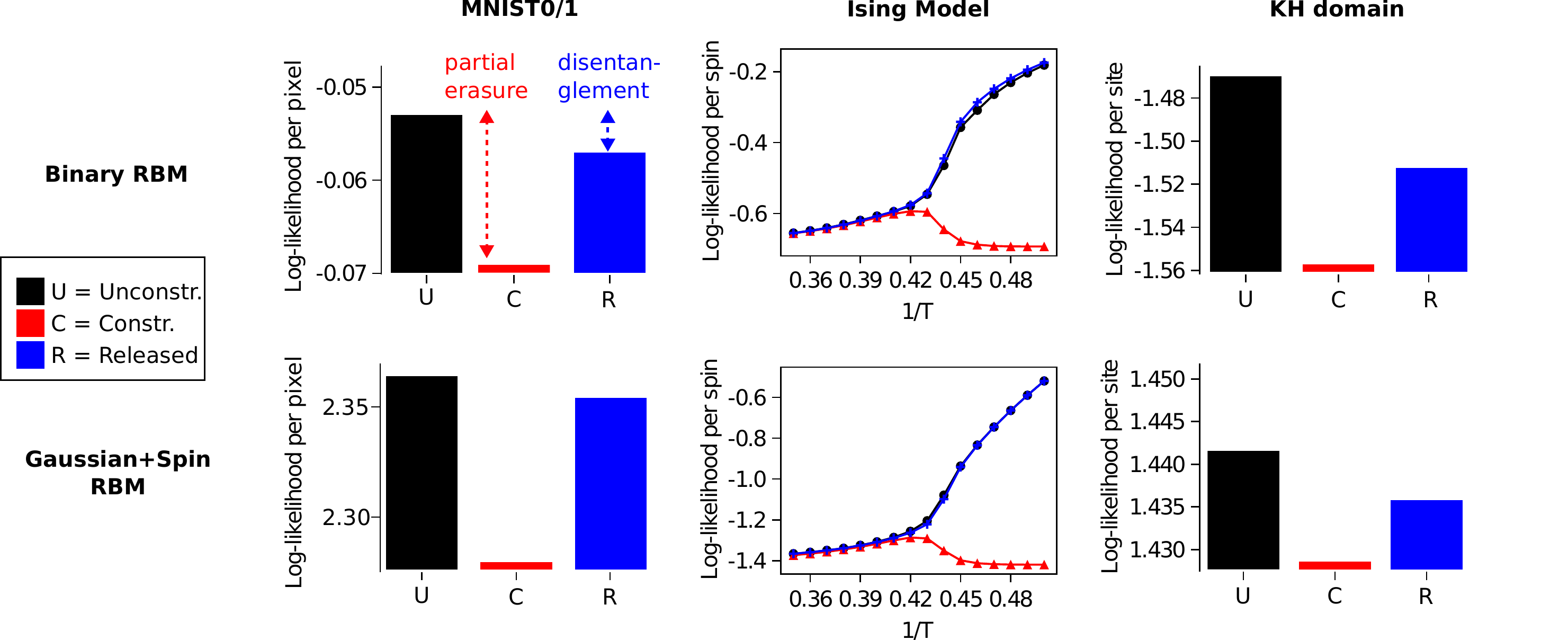}
\caption{\label{fig:poincare}\textbf{Likelihood calculations.} First row shows numerical estimates of the log-likelihood using RBMs with binary hidden units, along with the costs of applying \eqref{eq:Worth} partially or on the full hidden layer. Bottom row shows analytical results obtained in an RBM with one hidden spin unit and the remaining Gaussian hidden units (Figure~\ref{fig:poincare2}). First column shows the legend: Black for the unconstrained model, red for models with all hidden units constrained, and blue for models with the constraint acting on all but one hidden unit. Subsequent columns show the results for the three datasets considered: MNIST0/1, two-dimensional Ising model ($L=64$), and the KH protein domain. The discontinuous arrows in the first panel highlight the likelihood costs of partial label erasure (red) and disentanglement (blue).}
\end{figure*}

In this section we estimate the cost associated to disentanglement, see Section \ref{sec:introcost}, focusing on the impact of linear constraints on the weights. We resort to both numerical and analytical methods to estimates these costs.

\subsection{Numerical estimates}

Computing the likelihood requires estimating the normalization constant $Z$ in Eq.~\eqref{eq:rbm-energy}. Since the exact calculation of $Z$ is intractable  we use the annealed importance sampling (AIS) algorithm \cite{neal1998annealed}. AIS estimates $Z$ through a number of intermediate `annealed' distributions interpolating between the original RBM distribution and a simpler independent model that can be exactly sampled. This procedure provides a stochastic upper bound on the likelihood, which converges to the true value as the number of interpolating distribution increases. A stochastic lower bound can be obtained by a reverse interpolation procedure \cite{burda2015accurate}, which gradually `melts' the RBM back into the independent model; see SM Appendix \ref{SI:sec:impl} for details. Combining the two bounds sandwiches the true likelihood value and ensures that sampling has converged.

Results are shown for the Ising model, MNIST0/1, and PF00013 datasets considered in this work in the top row of Figure \ref{fig:poincare}. We have not considered CelebA for computational convenience. We first measure the likelihood costs $\Delta {\cal L}_{part.erasure}$, see Eq.~\eqref{eq:cost1a}, for making labels inaccessible to linear discriminators with the fully-constrained architexture (red bars or dots). In all datasets the labels considered are relevant to the nature of the data, and the costs (per data configuration) induced by the constraints on the weights are significant, see Table~\ref{table}.

\begin{table*}[t]
    \centering
    \begin{tabular}{|c|c||c|c||c|c|}
    \hline
    \text{model} & \text{label}  &$\Delta {\cal L}_{part.erasure}$  & \text{\% of unconstrained} & $\Delta {\cal L}_{disent.}$  & \text{\% of unconstrained}\\
      & & & \text{log-likelihood}& & \text{log-likelihood}\\ \hline
      MNIST0/1 & 0 \text{or} 1& 0.016  & 30\%  & 0.005 & 10\%\\\hline
      2D-Ising & \text{sign of} & 0.18& 40\% & $\simeq 0$ & $\simeq 0$\% \\
     ($L=64,\beta=0.44$)& \text{magnetization} & & & & \\ \hline
      KH domain & \text{bacteria or} & 0.09 & 6\%  &0.04  & 3\% \\
      & eukaryotic& & & & \\\hline
    \end{tabular}
    \caption{Decrease of log-likelihoods corresponding to partial erasure of the label with fully constrained RBM, $\Delta {\cal L}_\text{part. erasure}$, and to disentanglement with partially constrained RBM, $\Delta {\cal L}_\text{disent.}$. The changes on log-likelihoods are expressed per data configuration and per pixel for MNIST0/1, per spin for 2-Ising, and per protein site for the KH domain.}
    \label{table}
\end{table*}


The relation between label relevance and the likelihood cost is nicely portrayed in the two-dimensional Ising model dataset. At low $\beta$, the data is essentially random and the magnetization is mostly irrelevant to determining the probability of a configuration. In this regime, erasing label information has little likelihood cost. As the inverse temperature increases, the magnetization becomes more relevant, and it becomes necessary for the model to account for it to achieve good likelihood. In consequence, partially erasing the magnetization in this regime results in a large likelihood loss.

The top row of Figure \ref{fig:poincare} furthermore shows the values of the log-likelihoods after releasing one hidden unit (blue bars and dots). The log-likelihood loss with respect to the unconstrained RBM, $\Delta {\cal L}_\textrm{disent.}$ in Eq.~\eqref{eq:cost1b} is guaranteed to be non-negative. In practice, for the MNIST0/1 and Ising model datasets, and to a lesser extent for the KH domain, we estimate this cost to be small, see Table \ref{table}.
These results are consistent with the ability of the released RBM to fit and generate high-quality data in the three cases, as shown in previous sections.

\subsection{Analytical estimates}

\begin{figure}
\includegraphics[width=\linewidth]{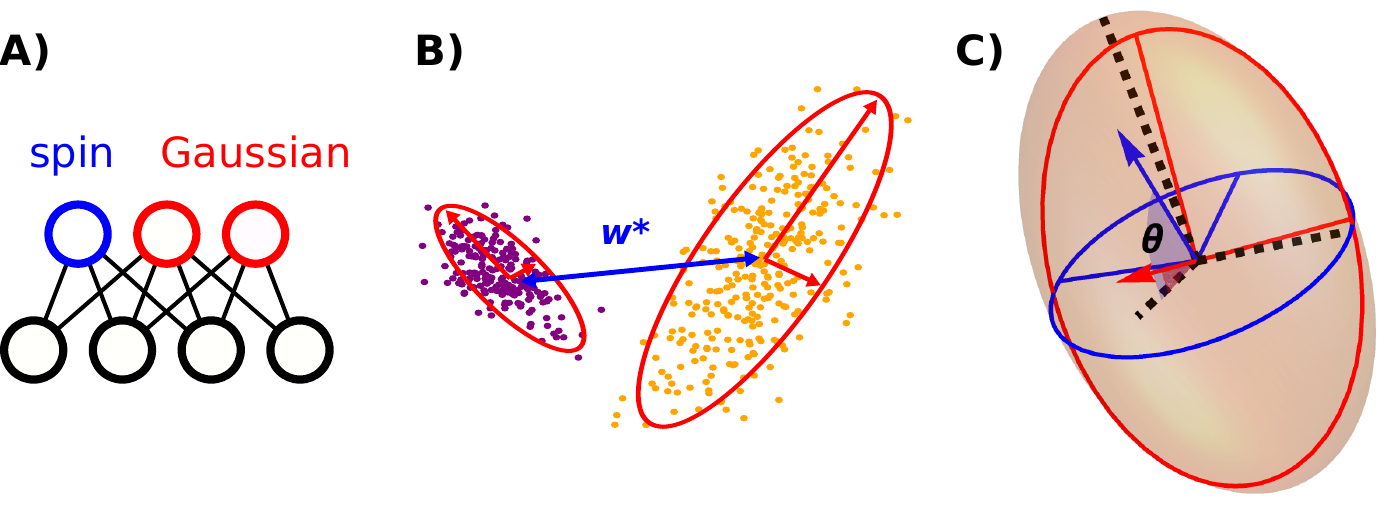}
\caption{\label{fig:poincare2} \textbf{Gaussian-Spin RBM.} \textbf{A)} The Gaussian-Spin RBM has one spin-like hidden unit, $h^\ast=h_1=\pm 1$, whereas all other hidden units are Gaussian. \textbf{B)} The spin hidden unit (blue) separates the two labelled classes. Gaussian hidden units (red) model intra-class variability. \textbf{C)} Illustration of Poincaré theorem.}
\end{figure}

We can gain some analytical insights about the origin of the costs of partial erasure and of disentanglement as follows. To make our RBM models mathematically tractable we now assume that the visible and hidden units of the RBM are all real valued and Gaussianly distributed, with the exception of a single spin-like hidden unit, $h^\ast=h_1=\pm 1$ (intended to be eventually released to help concentrating label-related information). This RBM model defines a bimodal Gaussian mixture distribution, with two modes associated to the label classes $u=\pm 1$, see Figure \ref{fig:poincare2}A \& B.

The energy function under this Gaussian-Spin RBM model (GS) writes,
\begin{multline}
E_\mathrm{GS}(\mathbf v, \mathbf h) = \sum_i \frac{v_i^2}{2\sigma_i^2} - \sum_i g_i v_i + \sum_{\mu \ge 2}\frac{h_\mu^2}{2} \\
- \sum_{i}\sum_{\mu \ge 2} w_{i\mu}v_i h_\mu - \sum_i w_i^\ast v_i h_1
\end{multline}
where the $\sigma_i$'s parametrize the standard deviations of the visible units, and the visible units are connected to the Gaussian hidden units through the weights $w_{i\mu}$, and to the spin hidden unit through $w_i^\ast$.

We first train the RBM in the absence of any constraint on the weights. The data are characterized by their empirical correlation matrix, $\mathbf{C}$, and the vector $\mathbf{q}^{(1)}$ separating the center of masses between the classes, see Figure~\ref{fig:datasets}C. Maximizing the likelihood of the data gives several conditions over the weight vectors that we list below:

\begin{enumerate}
\item The scaled weights $w_{i\mu}\sigma_i$ for $\mu\ge2$ are eigenvectors of the matrix $\tilde{\mathbf C} = \mathbf{D}(\mathbf{C}-\mathbf{q}^{(1)}(\mathbf{q}^{(1)})^\top)\mathbf{D}$, with corresponding eigenvalues $\lambda_\mu = 1/(1-\sum_i w_{i\mu}^2\sigma_i^2)$; here $\mathbf{D}$ is the diagonal matrix with entries $1/\sigma_i^2$. In practice, the top $M-1$  eigenvalues of $\tilde{\mathbf C}$ (larger than unity) have to be selected to maximize the likelihood.
\item The weights $\mathbf{w}^\ast$ onto hidden unit $\mu=1$ are given by $\boldsymbol\Sigma^{-1}\mathbf{q}^{(1)}$, where $\boldsymbol\Sigma=(\mathbf{D}-\mathbf{W}\mathbf{W}^\top)^{-1}$ denotes the conditional covariance matrix predicted by the model within each class, and $\mathbf{W}$ is the matrix of weight vectors $w_{i\mu}$ with $\mu\ge 2$.
\item The biases on the visible units are such that the model fits the independent site frequencies: $\mathbf{g}=\boldsymbol\Sigma^{-1}(\langle \mathbf{v} \rangle_{\mathcal{D}} - \mathbf{q}^{(1)})$.
\end{enumerate}
Details about the derivation can be found in SM Appendix \ref{SI:sec:GaussianSpin}. The log-likelihood reads
\begin{equation}\label{eq:likelihoodGS}
\mathcal L_\mathrm{GS}=\frac{1}{2} \sum_{\mu}(\lambda_{\mu} - 1 - \log \lambda_{\mu}) - \log\cosh\left(\mathbf g\cdot\mathbf{q}^{(1)}\right)
\end{equation}
where the $\lambda_\mu$'s are the selected eigenvalues of $\tilde{\mathbf{C}}$, and we have ignored irrelevant additive terms.

We next consider maximum likelihood training of a RBM in the presence of orthogonality constraints acting on the Gaussian weights, while $w_i^\ast$ is unconstrained, see Eq.~\eqref{eq:Worth}. Let us define the projection operator onto the subspace orthogonal to $\mathbf{q}^{(1)}$,
\begin{equation}\label{eq:projectionq}
\mathbf{P}=\mathbb{I} - \frac{\mathbf{q}^{(1)}(\mathbf{q}^{(1)})^\top}{|\mathbf{q}^{(1)}|^2}\ .
\end{equation}
It is easy to realize that conditions \eqref{eq:Worth} are equivalent to $\mathbf P \mathbf W = \mathbf W$. Consequently the discussion of the unconstrained learning case above applies to the constrained case provided the correlation matrix $\tilde{ \mathbf{C}}$ is replaced with the projected matrix $\tilde{\mathbf {C}}^\perp=\mathbf{P}\tilde{\mathbf {C}}\mathbf{P}$.

The eigenvalues of the projected matrix $\tilde{\mathbf C}^\perp$ have a precise ordering relationship to the eigenvalues of the original matrix $\tilde{\mathbf C}$, known as Poincar\'e separation theorem (see Theorem 11.11 of \cite{abadir2005matrix}). Denoting by $\lambda_1, \dots, \lambda_N$ the eigenvalues of the original matrix, and by $\lambda^\perp_1, \dots, \lambda^\perp_N$ the eigenvalues of the projected matrix, both ranked in decreasing order, we have
\begin{equation}\label{eq:Poincare}
\lambda_1 \ge \lambda^\perp_1 \ge \lambda_2 \ge \lambda^\perp_2 \ge \dots \ge \lambda_N \ge \lambda^\perp_N = 0 \ ,
\end{equation}
where $\lambda^\perp_N=0$ is due to the forbidden direction $\mathbf{q}^{(1)}$, which results in a drop of the rank of the matrix. Moreover, the gaps $\lambda_i - \lambda^\perp_i$, are connected to the angle between the forbidden direction $\mathbf{q}^{(1)}$ and the eigenvectors of the original correlation matrix. Figure \ref{fig:poincare2}C shows a low-dimensional example, in which a 3-dimensional ellipsoid symbolizing $\tilde{\mathbf{C}}$ is projected to the space orthogonal to one of the vectors shown. We consider two vectors with different angles to the ellipsoid principal axis, which define the projected ellipse $\tilde{\mathbf{C}}^\perp$.

The likelihood of the released Gaussian-Spin RBM is given by the same formula as for the unconstrained model, see Eq.~\eqref{eq:likelihoodGS}, upon replacement $\lambda_\mu\to \lambda_\mu^\perp$. As the function is monotonous in the eigenvalues (when they are larger than unity) Poincaré separation theorem in Eq.~\eqref{eq:Poincare} guarantees that the likelihood decreases when imposing the constraints on the weights.

Lastly, when the orthogonality constraint \eqref{eq:Worth} acts on all weights, the model is blind to the separation of the classes. We obtain the likelihood of the constrained RBM by simply replacing $\mathbf q^{(1)}$ in the above calculation with the zero vector, and consequently $w_i^\ast=0$ also.

The bottom row of Figure \ref{fig:poincare} shows the log-likelihoods estimates produced by this approximate calculation in the unconstrained, constrained and released cases. While the absolute values of the log-likelihoods cannot be directly compared to the binary RBM settings, we see that the relative changes from unconstrained to constrained, associated to the partial erasure cost, and  from constrained to released, defining the disentanglement cost fairly match their counterparts  computed by annealed importance sampling on Binary RBMs.

\section{Discussion}


In this work, we have proposed computationally efficient methods to train RBMs with disentangled representations. In turn, these representations can be used to generate samples with desired properties, {\em e.g.} with one attribute changed while the other features remain unaffected. This goal has been pursued in the literature \cite{hu2018disentangling,kim2018disentangling,lample2017fader,he2019attgan} with deep neural networks, predominantly with Variational Auto-Encoders (VAE) \cite{higgins2016beta,kingma2013auto} and adversarial networks \cite{goodfellow2014generative,lample2017fader,he2019attgan}. Despite the  broad success of adversarial learning and its importance in practical applications \cite{lample2017fader}, the aforementioned methods suffer from several drawbacks. Deep neural networks are difficult to interpret and require large amounts of data to train. Variational auto-encoders \cite{kingma2013auto} enforce a continuous mapping of the data to a Gaussian distribution, which is not always suitable, for instance if the data consist of separated peaks \cite{goldt2022gaussian}. Last of all, adversarial training suffers from instabilities that are not fully understood yet, making training difficult to implement in practice.

Our approach exploits the simplicity of the RBM architecture. Despite the limited number of layers the flexibility in the potentials on hidden units allows RBM to express complex representation distributions, contrary to VAE that require deeper architectures to map the data distribution onto Gaussian latent variables. We derive explicit constraints to be applied to the RBM weights during learning to favor disentangled representations. These constraints enforce that the data representations corresponding to different label classes are approximately indistinguishable. More precisely, we impose linear and quadratic constraints on the RBM weights that (partially) decorrelate the class label from the hidden-unit activities. As in an adversarial framework, imposing these constraints on a subset of hidden units only allows us to manipulate the samples generated from the model by controlling the state of the remaining hidden units. 

The resulting training algorithm is easily implementable and fast, being based on two steps. First, we estimate the required constraints from labeled data. Crucially, this is the only step that requires labels. Second, we train the RBM with standard learning procedures \cite{tieleman2008training}, making sure that, after each gradient update, the weights are projected into the subspace satisfying the constraints. The resulting procedure has similar computational cost as standard RBM training. It is therefore robust, not suffering from instability due to the maximization-minimization of the cost function appearing in adversarial learning schemes. We again stress that our approach combines the unsupervised nature of the RBM with constraints that are derived from labeled data. Therefore our model can be said to be \emph{semi-supervised}. We have shown how this synergy results in a model able to work in a regime with limited amount of labeled data. This result is important as, in many cases, labeled data are much more expensive to obtain than unlabeled data: data have to be annotated by humans (for instance, in the PF00013 dataset of the KH domain sequences, taxonomy labels are available for less than 10\% of the sequences), or costly experiments have to be done to get the label (this is the case for most biological data, which often require complex biophysical/ biochemical characterizations). 


We have demonstrated the effectiveness of this approach on four datasets from diverse domains: the CelebA dataset of face images \cite{liu2015faceattributes}, the Ising model from statistical physics, the MNIST collection of handwritten digit images \cite{deng2012mnist}, and protein sequences of the KH domain family \cite{el2019pfam}. 

CelebA \cite{liu2015faceattributes} and MNIST \cite{deng2012mnist} are popular benchmark datasets in machine-learning. In MNIST, the labels are straightforwardly associated to the digit identities. On this dataset, we have shown that RBM can be trained to associate one or few controlling hidden units to each digit class, which can be manipulated to sample and transition between classes. In CelebA, the labels correspond to subtle attributes of face images, like facial expressions (smiling / not smiling), or adornments (presence of eye glasses). Even for this complex dataset, RBM can sample good-looking images and are able to concentrate these attributes over few hidden units.

The two-dimensional Ising model is a very well studied system in statistical physics, with a precisely characterized phase transition controlled by the temperature. Standard RBM is able to reproduce the behaviors of observables, such as the magnetization, heat capacity, susceptibility, and correlation length. We then imposed a linear constraint on the weights (see \eqref{eq:Worth}), decorrelating the latent representation from the magnetization sign, and forcing the RBM to hallucinate a new system with interesting physical properties. Remarkably, the constrained RBM generates configurations with zero net magnetization, it preserves the structure of correlations between spins, as evident from second-order observables, such as the heat capacity and correlation length. Through an heuristic argument we proposed a Hamiltonian to describe the physical properties of this system, containing a non-analytic penalty term for the global magnetization, reminiscent of  non-analytic Landau potentials recently proposed to describe non-equilibrium steady states of the Ising magnet \cite{belitz2005generic, aron2020nonanalytic,aron2020landau}. Releasing a single hidden unit then restores the ability of the model to generate magnetized configurations, reproducing all statistics of the original Ising model.

Our last application was in protein design, based on model learning from sequence data, a field which has grown in importance in bio-engineering since the recent impressive developments of sequencing technologies \cite{rube2022prediction}. RBM trained on the K-Homology domain family under linear constraints decorrelating a subset of hidden inputs from the taxonomy of sequences, efficiently concentrate taxonomic information in a control hidden unit. Conditional sampling reproduces the fine statistical differences of the eukaryotic and bacterial sub-families. The transition between the two classes, takes place on a shorter time than the overall decorrelation time, suggesting that sequences might be able to change class while maintaining a memory of other, class-independent attributes. 


Concentrating information about important features of the data into one or few hidden units of the RBM could {\em a priori} be detrimental to the ability of the model to fit the data for two reasons. First, introducing constraints on the weights is expected to impact (decrease) the log-likelihood of the data generated by the RBM. We estimated the losses in log-likelihood due to partial erasure and to disentanglement for several datasets. The cost of partial erasure is related to the relevance of the label, as clearly illustrated in the dependence on temperature in the Ising model data. Remarkably, we find that disentanglement is achieved with a small relative likelihood loss, evidencing the robustness of the approach. Furthermore, when the data can be approximated as a mixture of two Gaussian distributions, we have shown how the log-likelihood losses could be analytically calculated, and have established a connection between the likelihood costs for erasure or disentanglement and the Poincar\'e separation theorem. 

Second, the few (often, single) released hidden units encode label-associated features in a prototype-like way. In the case of linear constraints released weights are aligned the $\mathbf{q}^{(1)}$ vector, equal to the relative difference between the centers of mass of the two label classes, see Figure~\ref{fig:celeba_sampling}A for an illustration on CelebA.  It is however widely believed that prototype-like representations are poorer than compositional ones, in which multiple features associated to many hidden units can be combinatorially combined to create high-quality and diverse data \cite{tubiana2017}. From this point of view, forcing some hidden units to generate prototypes could appear counterproductive. It is nevertheless a very effective way to drive class switching, see for instance Figure~\ref{fig:celeba_transition}. In addition all the important features defining the data distribution are learned by the vast number of other (constrained) hidden units, which, in turn, can be combined together to collectively participate in the data generation process. We also emphasize that, while a few hidden units capture enough label-associated features to manipulate and drive the label values, this does not mean that they concentrate all the information about the label. As clearly shown in Figure~\ref{fig:ising}F for Ising and Figure \ref{fig:mnist}B for MNIST0/1 there remains substantial information about the label in the constrained hidden units, accessible to deep decoders. Hence label-associated features are residually encoded in a combinatorial way by the RBM.


While disentangling and manipulating representations through our `partially constrained' RBM approach offers clear advantages in terms of usability and interpretability the other architecture we considered in this work, the so-called `fully constrained' RBM may also be of interest in practical applications. Informally speaking, fully constrained RBMs are appropriate to model the features in the data orthogonal to the ones associated to the label under consideration.  We have shown that fully constrained RBMs remain generative in two examples (CelebA and PF00013), where samples resemble data configurations with ambiguous class identity. In the MNIST0/1 and Ising model examples, however, the fully constrained RBM generates samples markedly different from the data (zero magnetization in the Ising case, and blurry mixtures of 0's and 1's for MNIST). We attribute this to the fact that in these later cases, the datasets corresponding to the two values of the label are widely separated. However, as we show in the Ising case, information is preserved in higher-order moments of the samples (\emph{e.g.} heat-capacity). Another example is shown in SM Fig.~\ref{SI:fig:mnist_black_white}, where a fully constrained RBM trained on zero MNIST digits in black or white backgrounds generates zeros encoded in the correlations between neighboring pixels. As a potential future direction for fully-constrained RBM our results on the KH domain open the way to the reconstruction of ancestral (backwards in evolutionary time) proteins, which were possibly more functionally promiscuous than their current counterparts. It would be very interesting to apply our approach to reconstruct putative ancient proteins, {\em e.g.} where details about binding specificity are erased while the other functionalities (stability, activity, ...) are maintained.


In summary, our work proposes a flexible semi-supervised framework for learning disentangled representations, easily implementable and amenable to approximate analytical calculations.  We hope our approach will make controlled generation of data and feature discovery  easier in future applications. Last of all, besides the applications to RBM we present here, it would be interesting to transfer our constraint-based framework to other architectures, as the principle of imposing constraints on the weights in the course of learning is quite general.

The codes needed to reproduce the results reported in this work are available on 
\href{https://github.com/cossio/AdvRBMs_App.jl}{Github}.

\begin{acknowledgments}
J.FdCD, S.C., R.M. are supported by the ANR-17 RBMPro CE30-0021-01 and ANR-19 Decrypted CE30-0021-01 grants.
\end{acknowledgments}

\bibliography{adv}

\providecommand{\noopsort}[1]{}\providecommand{\singleletter}[1]{#1}%
\begin{thebibliography}{67}%
\makeatletter
\providecommand \@ifxundefined [1]{%
 \@ifx{#1\undefined}
}%
\providecommand \@ifnum [1]{%
 \ifnum #1\expandafter \@firstoftwo
 \else \expandafter \@secondoftwo
 \fi
}%
\providecommand \@ifx [1]{%
 \ifx #1\expandafter \@firstoftwo
 \else \expandafter \@secondoftwo
 \fi
}%
\providecommand \natexlab [1]{#1}%
\providecommand \enquote  [1]{``#1''}%
\providecommand \bibnamefont  [1]{#1}%
\providecommand \bibfnamefont [1]{#1}%
\providecommand \citenamefont [1]{#1}%
\providecommand \href@noop [0]{\@secondoftwo}%
\providecommand \href [0]{\begingroup \@sanitize@url \@href}%
\providecommand \@href[1]{\@@startlink{#1}\@@href}%
\providecommand \@@href[1]{\endgroup#1\@@endlink}%
\providecommand \@sanitize@url [0]{\catcode `\\12\catcode `\$12\catcode
  `\&12\catcode `\#12\catcode `\^12\catcode `\_12\catcode `\%12\relax}%
\providecommand \@@startlink[1]{}%
\providecommand \@@endlink[0]{}%
\providecommand \url  [0]{\begingroup\@sanitize@url \@url }%
\providecommand \@url [1]{\endgroup\@href {#1}{\urlprefix }}%
\providecommand \urlprefix  [0]{URL }%
\providecommand \Eprint [0]{\href }%
\providecommand \doibase [0]{https://doi.org/}%
\providecommand \selectlanguage [0]{\@gobble}%
\providecommand \bibinfo  [0]{\@secondoftwo}%
\providecommand \bibfield  [0]{\@secondoftwo}%
\providecommand \translation [1]{[#1]}%
\providecommand \BibitemOpen [0]{}%
\providecommand \bibitemStop [0]{}%
\providecommand \bibitemNoStop [0]{.\EOS\space}%
\providecommand \EOS [0]{\spacefactor3000\relax}%
\providecommand \BibitemShut  [1]{\csname bibitem#1\endcsname}%
\let\auto@bib@innerbib\@empty
\bibitem [{\citenamefont {Bengio}(2012)}]{bengio2012deep}%
  \BibitemOpen
  \bibfield  {author} {\bibinfo {author} {\bibfnamefont {Y.}~\bibnamefont
  {Bengio}},\ }\bibfield  {title} {\bibinfo {title} {Deep learning of
  representations for unsupervised and transfer learning},\ }in\ \href@noop {}
  {\emph {\bibinfo {booktitle} {Proceedings of ICML workshop on unsupervised
  and transfer learning}}}\ (\bibinfo {organization} {JMLR Workshop and
  Conference Proceedings},\ \bibinfo {year} {2012})\ pp.\ \bibinfo {pages}
  {17--36}\BibitemShut {NoStop}%
\bibitem [{\citenamefont {Salakhutdinov}\ and\ \citenamefont
  {Hinton}(2009)}]{salakhutdinov2009deep}%
  \BibitemOpen
  \bibfield  {author} {\bibinfo {author} {\bibfnamefont {R.}~\bibnamefont
  {Salakhutdinov}}\ and\ \bibinfo {author} {\bibfnamefont {G.}~\bibnamefont
  {Hinton}},\ }\bibfield  {title} {\bibinfo {title} {Deep boltzmann machines},\
  }in\ \href@noop {} {\emph {\bibinfo {booktitle} {Artificial intelligence and
  statistics}}}\ (\bibinfo {organization} {PMLR},\ \bibinfo {year} {2009})\
  pp.\ \bibinfo {pages} {448--455}\BibitemShut {NoStop}%
\bibitem [{\citenamefont {Kingma}\ and\ \citenamefont
  {Welling}(2013)}]{kingma2013auto}%
  \BibitemOpen
  \bibfield  {author} {\bibinfo {author} {\bibfnamefont {D.~P.}\ \bibnamefont
  {Kingma}}\ and\ \bibinfo {author} {\bibfnamefont {M.}~\bibnamefont
  {Welling}},\ }\bibfield  {title} {\bibinfo {title} {Auto-encoding variational
  bayes},\ }\href@noop {} {\bibfield  {journal} {\bibinfo  {journal} {arXiv
  preprint arXiv:1312.6114}\ } (\bibinfo {year} {2013})}\BibitemShut {NoStop}%
\bibitem [{\citenamefont {Goodfellow}\ \emph {et~al.}(2014)\citenamefont
  {Goodfellow}, \citenamefont {Pouget-Abadie}, \citenamefont {Mirza},
  \citenamefont {Xu}, \citenamefont {Warde-Farley}, \citenamefont {Ozair},
  \citenamefont {Courville},\ and\ \citenamefont
  {Bengio}}]{goodfellow2014generative}%
  \BibitemOpen
  \bibfield  {author} {\bibinfo {author} {\bibfnamefont {I.~J.}\ \bibnamefont
  {Goodfellow}}, \bibinfo {author} {\bibfnamefont {J.}~\bibnamefont
  {Pouget-Abadie}}, \bibinfo {author} {\bibfnamefont {M.}~\bibnamefont
  {Mirza}}, \bibinfo {author} {\bibfnamefont {B.}~\bibnamefont {Xu}}, \bibinfo
  {author} {\bibfnamefont {D.}~\bibnamefont {Warde-Farley}}, \bibinfo {author}
  {\bibfnamefont {S.}~\bibnamefont {Ozair}}, \bibinfo {author} {\bibfnamefont
  {A.}~\bibnamefont {Courville}},\ and\ \bibinfo {author} {\bibfnamefont
  {Y.}~\bibnamefont {Bengio}},\ }\bibfield  {title} {\bibinfo {title}
  {Generative adversarial networks},\ }\href@noop {} {\bibfield  {journal}
  {\bibinfo  {journal} {arXiv preprint arXiv:1406.2661}\ } (\bibinfo {year}
  {2014})}\BibitemShut {NoStop}%
\bibitem [{\citenamefont {Johnston}\ \emph {et~al.}(2020)\citenamefont
  {Johnston}, \citenamefont {Palmer},\ and\ \citenamefont
  {Freedman}}]{johnston2020nonlinear}%
  \BibitemOpen
  \bibfield  {author} {\bibinfo {author} {\bibfnamefont {W.~J.}\ \bibnamefont
  {Johnston}}, \bibinfo {author} {\bibfnamefont {S.~E.}\ \bibnamefont
  {Palmer}},\ and\ \bibinfo {author} {\bibfnamefont {D.~J.}\ \bibnamefont
  {Freedman}},\ }\bibfield  {title} {\bibinfo {title} {Nonlinear mixed
  selectivity supports reliable neural computation},\ }\href@noop {} {\bibfield
   {journal} {\bibinfo  {journal} {PLOS computational biology}\ }\textbf
  {\bibinfo {volume} {16}},\ \bibinfo {pages} {e1007544} (\bibinfo {year}
  {2020})}\BibitemShut {NoStop}%
\bibitem [{\citenamefont {Locatello}\ \emph {et~al.}(2019)\citenamefont
  {Locatello}, \citenamefont {Bauer}, \citenamefont {Lucic}, \citenamefont
  {Raetsch}, \citenamefont {Gelly}, \citenamefont {Sch{\"o}lkopf},\ and\
  \citenamefont {Bachem}}]{locatello2019challenging}%
  \BibitemOpen
  \bibfield  {author} {\bibinfo {author} {\bibfnamefont {F.}~\bibnamefont
  {Locatello}}, \bibinfo {author} {\bibfnamefont {S.}~\bibnamefont {Bauer}},
  \bibinfo {author} {\bibfnamefont {M.}~\bibnamefont {Lucic}}, \bibinfo
  {author} {\bibfnamefont {G.}~\bibnamefont {Raetsch}}, \bibinfo {author}
  {\bibfnamefont {S.}~\bibnamefont {Gelly}}, \bibinfo {author} {\bibfnamefont
  {B.}~\bibnamefont {Sch{\"o}lkopf}},\ and\ \bibinfo {author} {\bibfnamefont
  {O.}~\bibnamefont {Bachem}},\ }\bibfield  {title} {\bibinfo {title}
  {Challenging common assumptions in the unsupervised learning of disentangled
  representations},\ }in\ \href@noop {} {\emph {\bibinfo {booktitle}
  {international conference on machine learning}}}\ (\bibinfo {organization}
  {PMLR},\ \bibinfo {year} {2019})\ pp.\ \bibinfo {pages}
  {4114--4124}\BibitemShut {NoStop}%
\bibitem [{\citenamefont {Lample}\ \emph {et~al.}(2017)\citenamefont {Lample},
  \citenamefont {Zeghidour}, \citenamefont {Usunier}, \citenamefont {Bordes},
  \citenamefont {Denoyer},\ and\ \citenamefont {Ranzato}}]{lample2017fader}%
  \BibitemOpen
  \bibfield  {author} {\bibinfo {author} {\bibfnamefont {G.}~\bibnamefont
  {Lample}}, \bibinfo {author} {\bibfnamefont {N.}~\bibnamefont {Zeghidour}},
  \bibinfo {author} {\bibfnamefont {N.}~\bibnamefont {Usunier}}, \bibinfo
  {author} {\bibfnamefont {A.}~\bibnamefont {Bordes}}, \bibinfo {author}
  {\bibfnamefont {L.}~\bibnamefont {Denoyer}},\ and\ \bibinfo {author}
  {\bibfnamefont {M.}~\bibnamefont {Ranzato}},\ }\bibfield  {title} {\bibinfo
  {title} {Fader networks: Manipulating images by sliding attributes},\
  }\href@noop {} {\bibfield  {journal} {\bibinfo  {journal} {arXiv preprint
  arXiv:1706.00409}\ } (\bibinfo {year} {2017})}\BibitemShut {NoStop}%
\bibitem [{\citenamefont {Kim}\ and\ \citenamefont
  {Mnih}(2018)}]{kim2018disentangling}%
  \BibitemOpen
  \bibfield  {author} {\bibinfo {author} {\bibfnamefont {H.}~\bibnamefont
  {Kim}}\ and\ \bibinfo {author} {\bibfnamefont {A.}~\bibnamefont {Mnih}},\
  }\bibfield  {title} {\bibinfo {title} {Disentangling by factorising},\ }in\
  \href@noop {} {\emph {\bibinfo {booktitle} {International Conference on
  Machine Learning}}}\ (\bibinfo {organization} {PMLR},\ \bibinfo {year}
  {2018})\ pp.\ \bibinfo {pages} {2649--2658}\BibitemShut {NoStop}%
\bibitem [{\citenamefont {Hu}\ \emph {et~al.}(2018)\citenamefont {Hu},
  \citenamefont {Szab{\'o}}, \citenamefont {Portenier}, \citenamefont
  {Favaro},\ and\ \citenamefont {Zwicker}}]{hu2018disentangling}%
  \BibitemOpen
  \bibfield  {author} {\bibinfo {author} {\bibfnamefont {Q.}~\bibnamefont
  {Hu}}, \bibinfo {author} {\bibfnamefont {A.}~\bibnamefont {Szab{\'o}}},
  \bibinfo {author} {\bibfnamefont {T.}~\bibnamefont {Portenier}}, \bibinfo
  {author} {\bibfnamefont {P.}~\bibnamefont {Favaro}},\ and\ \bibinfo {author}
  {\bibfnamefont {M.}~\bibnamefont {Zwicker}},\ }\bibfield  {title} {\bibinfo
  {title} {Disentangling factors of variation by mixing them},\ }in\ \href@noop
  {} {\emph {\bibinfo {booktitle} {Proceedings of the IEEE Conference on
  Computer Vision and Pattern Recognition}}}\ (\bibinfo {year} {2018})\ pp.\
  \bibinfo {pages} {3399--3407}\BibitemShut {NoStop}%
\bibitem [{\citenamefont {Esmaeili}\ \emph {et~al.}(2019)\citenamefont
  {Esmaeili}, \citenamefont {Wu}, \citenamefont {Jain}, \citenamefont
  {Bozkurt}, \citenamefont {Siddharth}, \citenamefont {Paige}, \citenamefont
  {Brooks}, \citenamefont {Dy},\ and\ \citenamefont
  {Meent}}]{esmaeili2019structured}%
  \BibitemOpen
  \bibfield  {author} {\bibinfo {author} {\bibfnamefont {B.}~\bibnamefont
  {Esmaeili}}, \bibinfo {author} {\bibfnamefont {H.}~\bibnamefont {Wu}},
  \bibinfo {author} {\bibfnamefont {S.}~\bibnamefont {Jain}}, \bibinfo {author}
  {\bibfnamefont {A.}~\bibnamefont {Bozkurt}}, \bibinfo {author} {\bibfnamefont
  {N.}~\bibnamefont {Siddharth}}, \bibinfo {author} {\bibfnamefont
  {B.}~\bibnamefont {Paige}}, \bibinfo {author} {\bibfnamefont {D.~H.}\
  \bibnamefont {Brooks}}, \bibinfo {author} {\bibfnamefont {J.}~\bibnamefont
  {Dy}},\ and\ \bibinfo {author} {\bibfnamefont {J.-W.}\ \bibnamefont
  {Meent}},\ }\bibfield  {title} {\bibinfo {title} {Structured disentangled
  representations},\ }in\ \href@noop {} {\emph {\bibinfo {booktitle} {The 22nd
  International Conference on Artificial Intelligence and Statistics}}}\
  (\bibinfo {organization} {PMLR},\ \bibinfo {year} {2019})\ pp.\ \bibinfo
  {pages} {2525--2534}\BibitemShut {NoStop}%
\bibitem [{\citenamefont {He}\ \emph {et~al.}(2019)\citenamefont {He},
  \citenamefont {Zuo}, \citenamefont {Kan}, \citenamefont {Shan},\ and\
  \citenamefont {Chen}}]{he2019attgan}%
  \BibitemOpen
  \bibfield  {author} {\bibinfo {author} {\bibfnamefont {Z.}~\bibnamefont
  {He}}, \bibinfo {author} {\bibfnamefont {W.}~\bibnamefont {Zuo}}, \bibinfo
  {author} {\bibfnamefont {M.}~\bibnamefont {Kan}}, \bibinfo {author}
  {\bibfnamefont {S.}~\bibnamefont {Shan}},\ and\ \bibinfo {author}
  {\bibfnamefont {X.}~\bibnamefont {Chen}},\ }\bibfield  {title} {\bibinfo
  {title} {Attgan: Facial attribute editing by only changing what you want},\
  }\href@noop {} {\bibfield  {journal} {\bibinfo  {journal} {IEEE transactions
  on image processing}\ }\textbf {\bibinfo {volume} {28}},\ \bibinfo {pages}
  {5464} (\bibinfo {year} {2019})}\BibitemShut {NoStop}%
\bibitem [{\citenamefont {Shen}\ \emph {et~al.}(2020)\citenamefont {Shen},
  \citenamefont {Gu}, \citenamefont {Tang},\ and\ \citenamefont
  {Zhou}}]{shen2020interpreting}%
  \BibitemOpen
  \bibfield  {author} {\bibinfo {author} {\bibfnamefont {Y.}~\bibnamefont
  {Shen}}, \bibinfo {author} {\bibfnamefont {J.}~\bibnamefont {Gu}}, \bibinfo
  {author} {\bibfnamefont {X.}~\bibnamefont {Tang}},\ and\ \bibinfo {author}
  {\bibfnamefont {B.}~\bibnamefont {Zhou}},\ }\bibfield  {title} {\bibinfo
  {title} {Interpreting the latent space of gans for semantic face editing},\
  }in\ \href@noop {} {\emph {\bibinfo {booktitle} {Proceedings of the IEEE/CVF
  Conference on Computer Vision and Pattern Recognition}}}\ (\bibinfo {year}
  {2020})\ pp.\ \bibinfo {pages} {9243--9252}\BibitemShut {NoStop}%
\bibitem [{\citenamefont {Zaidi}\ \emph {et~al.}(2020)\citenamefont {Zaidi},
  \citenamefont {Boilard}, \citenamefont {Gagnon},\ and\ \citenamefont
  {Carbonneau}}]{zaidi2020measuring}%
  \BibitemOpen
  \bibfield  {author} {\bibinfo {author} {\bibfnamefont {J.}~\bibnamefont
  {Zaidi}}, \bibinfo {author} {\bibfnamefont {J.}~\bibnamefont {Boilard}},
  \bibinfo {author} {\bibfnamefont {G.}~\bibnamefont {Gagnon}},\ and\ \bibinfo
  {author} {\bibfnamefont {M.-A.}\ \bibnamefont {Carbonneau}},\ }\bibfield
  {title} {\bibinfo {title} {Measuring disentanglement: A review of metrics},\
  }\href@noop {} {\bibfield  {journal} {\bibinfo  {journal} {arXiv preprint
  arXiv:2012.09276}\ } (\bibinfo {year} {2020})}\BibitemShut {NoStop}%
\bibitem [{\citenamefont {Feutry}\ \emph {et~al.}(2018)\citenamefont {Feutry},
  \citenamefont {Piantanida}, \citenamefont {Bengio},\ and\ \citenamefont
  {Duhamel}}]{feutry2018learning}%
  \BibitemOpen
  \bibfield  {author} {\bibinfo {author} {\bibfnamefont {C.}~\bibnamefont
  {Feutry}}, \bibinfo {author} {\bibfnamefont {P.}~\bibnamefont {Piantanida}},
  \bibinfo {author} {\bibfnamefont {Y.}~\bibnamefont {Bengio}},\ and\ \bibinfo
  {author} {\bibfnamefont {P.}~\bibnamefont {Duhamel}},\ }\bibfield  {title}
  {\bibinfo {title} {Learning anonymized representations with adversarial
  neural networks},\ }\href@noop {} {\bibfield  {journal} {\bibinfo  {journal}
  {arXiv preprint arXiv:1802.09386}\ } (\bibinfo {year} {2018})}\BibitemShut
  {NoStop}%
\bibitem [{\citenamefont {Zemel}\ \emph {et~al.}(2013)\citenamefont {Zemel},
  \citenamefont {Wu}, \citenamefont {Swersky}, \citenamefont {Pitassi},\ and\
  \citenamefont {Dwork}}]{zemel2013learning}%
  \BibitemOpen
  \bibfield  {author} {\bibinfo {author} {\bibfnamefont {R.}~\bibnamefont
  {Zemel}}, \bibinfo {author} {\bibfnamefont {Y.}~\bibnamefont {Wu}}, \bibinfo
  {author} {\bibfnamefont {K.}~\bibnamefont {Swersky}}, \bibinfo {author}
  {\bibfnamefont {T.}~\bibnamefont {Pitassi}},\ and\ \bibinfo {author}
  {\bibfnamefont {C.}~\bibnamefont {Dwork}},\ }\bibfield  {title} {\bibinfo
  {title} {Learning fair representations},\ }in\ \href@noop {} {\emph {\bibinfo
  {booktitle} {International conference on machine learning}}}\ (\bibinfo
  {organization} {PMLR},\ \bibinfo {year} {2013})\ pp.\ \bibinfo {pages}
  {325--333}\BibitemShut {NoStop}%
\bibitem [{\citenamefont {Arjovsky}\ and\ \citenamefont
  {Bottou}(2017)}]{arjovsky2017towards}%
  \BibitemOpen
  \bibfield  {author} {\bibinfo {author} {\bibfnamefont {M.}~\bibnamefont
  {Arjovsky}}\ and\ \bibinfo {author} {\bibfnamefont {L.}~\bibnamefont
  {Bottou}},\ }\bibfield  {title} {\bibinfo {title} {Towards principled methods
  for training generative adversarial networks},\ }\href@noop {} {\bibfield
  {journal} {\bibinfo  {journal} {arXiv preprint arXiv:1701.04862}\ } (\bibinfo
  {year} {2017})}\BibitemShut {NoStop}%
\bibitem [{\citenamefont {Mikolov}\ \emph {et~al.}(2013)\citenamefont
  {Mikolov}, \citenamefont {Chen}, \citenamefont {Corrado},\ and\ \citenamefont
  {Dean}}]{word2vec}%
  \BibitemOpen
  \bibfield  {author} {\bibinfo {author} {\bibfnamefont {T.}~\bibnamefont
  {Mikolov}}, \bibinfo {author} {\bibfnamefont {K.}~\bibnamefont {Chen}},
  \bibinfo {author} {\bibfnamefont {G.}~\bibnamefont {Corrado}},\ and\ \bibinfo
  {author} {\bibfnamefont {J.}~\bibnamefont {Dean}},\ }\href
  {https://doi.org/10.48550/ARXIV.1301.3781} {\bibinfo {title} {Efficient
  estimation of word representations in vector space}} (\bibinfo {year}
  {2013})\BibitemShut {NoStop}%
\bibitem [{\citenamefont {Hinton}(2012)}]{hinton2012practical}%
  \BibitemOpen
  \bibfield  {author} {\bibinfo {author} {\bibfnamefont {G.~E.}\ \bibnamefont
  {Hinton}},\ }\bibfield  {title} {\bibinfo {title} {A practical guide to
  training restricted boltzmann machines},\ }in\ \href@noop {} {\emph {\bibinfo
  {booktitle} {Neural networks: Tricks of the trade}}}\ (\bibinfo  {publisher}
  {Springer},\ \bibinfo {year} {2012})\ pp.\ \bibinfo {pages}
  {599--619}\BibitemShut {NoStop}%
\bibitem [{\citenamefont {Tubiana}\ \emph {et~al.}(2019)\citenamefont
  {Tubiana}, \citenamefont {Cocco},\ and\ \citenamefont
  {Monasson}}]{tubiana2019learning}%
  \BibitemOpen
  \bibfield  {author} {\bibinfo {author} {\bibfnamefont {J.}~\bibnamefont
  {Tubiana}}, \bibinfo {author} {\bibfnamefont {S.}~\bibnamefont {Cocco}},\
  and\ \bibinfo {author} {\bibfnamefont {R.}~\bibnamefont {Monasson}},\
  }\bibfield  {title} {\bibinfo {title} {Learning protein constitutive motifs
  from sequence data},\ }\href@noop {} {\bibfield  {journal} {\bibinfo
  {journal} {Elife}\ }\textbf {\bibinfo {volume} {8}},\ \bibinfo {pages}
  {e39397} (\bibinfo {year} {2019})}\BibitemShut {NoStop}%
\bibitem [{\citenamefont {Bravi}\ \emph {et~al.}(2021)\citenamefont {Bravi},
  \citenamefont {Tubiana}, \citenamefont {Cocco}, \citenamefont {Monasson},
  \citenamefont {Mora},\ and\ \citenamefont {Walczak}}]{bravi2021rbm}%
  \BibitemOpen
  \bibfield  {author} {\bibinfo {author} {\bibfnamefont {B.}~\bibnamefont
  {Bravi}}, \bibinfo {author} {\bibfnamefont {J.}~\bibnamefont {Tubiana}},
  \bibinfo {author} {\bibfnamefont {S.}~\bibnamefont {Cocco}}, \bibinfo
  {author} {\bibfnamefont {R.}~\bibnamefont {Monasson}}, \bibinfo {author}
  {\bibfnamefont {T.}~\bibnamefont {Mora}},\ and\ \bibinfo {author}
  {\bibfnamefont {A.~M.}\ \bibnamefont {Walczak}},\ }\bibfield  {title}
  {\bibinfo {title} {Rbm-mhc: A semi-supervised machine-learning method for
  sample-specific prediction of antigen presentation by hla-i alleles},\
  }\href@noop {} {\bibfield  {journal} {\bibinfo  {journal} {Cell systems}\
  }\textbf {\bibinfo {volume} {12}},\ \bibinfo {pages} {195} (\bibinfo {year}
  {2021})}\BibitemShut {NoStop}%
\bibitem [{\citenamefont {Salakhutdinov}\ \emph {et~al.}(2007)\citenamefont
  {Salakhutdinov}, \citenamefont {Mnih},\ and\ \citenamefont
  {Hinton}}]{salakhutdinov2007restricted}%
  \BibitemOpen
  \bibfield  {author} {\bibinfo {author} {\bibfnamefont {R.}~\bibnamefont
  {Salakhutdinov}}, \bibinfo {author} {\bibfnamefont {A.}~\bibnamefont
  {Mnih}},\ and\ \bibinfo {author} {\bibfnamefont {G.}~\bibnamefont {Hinton}},\
  }\bibfield  {title} {\bibinfo {title} {Restricted boltzmann machines for
  collaborative filtering},\ }in\ \href@noop {} {\emph {\bibinfo {booktitle}
  {Proceedings of the 24th international conference on Machine learning}}}\
  (\bibinfo {year} {2007})\ pp.\ \bibinfo {pages} {791--798}\BibitemShut
  {NoStop}%
\bibitem [{\citenamefont {Abadir}\ and\ \citenamefont
  {Magnus}(2005)}]{abadir2005matrix}%
  \BibitemOpen
  \bibfield  {author} {\bibinfo {author} {\bibfnamefont {K.~M.}\ \bibnamefont
  {Abadir}}\ and\ \bibinfo {author} {\bibfnamefont {J.~R.}\ \bibnamefont
  {Magnus}},\ }\href@noop {} {\emph {\bibinfo {title} {Matrix algebra}}},\
  Vol.~\bibinfo {volume} {1}\ (\bibinfo  {publisher} {Cambridge University
  Press},\ \bibinfo {year} {2005})\BibitemShut {NoStop}%
\bibitem [{\citenamefont {Liu}\ \emph {et~al.}(2015)\citenamefont {Liu},
  \citenamefont {Luo}, \citenamefont {Wang},\ and\ \citenamefont
  {Tang}}]{liu2015faceattributes}%
  \BibitemOpen
  \bibfield  {author} {\bibinfo {author} {\bibfnamefont {Z.}~\bibnamefont
  {Liu}}, \bibinfo {author} {\bibfnamefont {P.}~\bibnamefont {Luo}}, \bibinfo
  {author} {\bibfnamefont {X.}~\bibnamefont {Wang}},\ and\ \bibinfo {author}
  {\bibfnamefont {X.}~\bibnamefont {Tang}},\ }\bibfield  {title} {\bibinfo
  {title} {Deep learning face attributes in the wild},\ }in\ \href@noop {}
  {\emph {\bibinfo {booktitle} {Proceedings of International Conference on
  Computer Vision (ICCV)}}}\ (\bibinfo {year} {2015})\BibitemShut {NoStop}%
\bibitem [{\citenamefont {Deng}(2012)}]{deng2012mnist}%
  \BibitemOpen
  \bibfield  {author} {\bibinfo {author} {\bibfnamefont {L.}~\bibnamefont
  {Deng}},\ }\bibfield  {title} {\bibinfo {title} {The {MNIST} database of
  handwritten digit images for machine learning research},\ }\href@noop {}
  {\bibfield  {journal} {\bibinfo  {journal} {IEEE Signal Processing Magazine}\
  }\textbf {\bibinfo {volume} {29}},\ \bibinfo {pages} {141} (\bibinfo {year}
  {2012})}\BibitemShut {NoStop}%
\bibitem [{\citenamefont {El-Gebali}\ \emph {et~al.}(2019)\citenamefont
  {El-Gebali}, \citenamefont {Mistry}, \citenamefont {Bateman}, \citenamefont
  {Eddy}, \citenamefont {Luciani}, \citenamefont {Potter}, \citenamefont
  {Qureshi}, \citenamefont {Richardson}, \citenamefont {Salazar}, \citenamefont
  {Smart} \emph {et~al.}}]{el2019pfam}%
  \BibitemOpen
  \bibfield  {author} {\bibinfo {author} {\bibfnamefont {S.}~\bibnamefont
  {El-Gebali}}, \bibinfo {author} {\bibfnamefont {J.}~\bibnamefont {Mistry}},
  \bibinfo {author} {\bibfnamefont {A.}~\bibnamefont {Bateman}}, \bibinfo
  {author} {\bibfnamefont {S.~R.}\ \bibnamefont {Eddy}}, \bibinfo {author}
  {\bibfnamefont {A.}~\bibnamefont {Luciani}}, \bibinfo {author} {\bibfnamefont
  {S.~C.}\ \bibnamefont {Potter}}, \bibinfo {author} {\bibfnamefont
  {M.}~\bibnamefont {Qureshi}}, \bibinfo {author} {\bibfnamefont {L.~J.}\
  \bibnamefont {Richardson}}, \bibinfo {author} {\bibfnamefont {G.~A.}\
  \bibnamefont {Salazar}}, \bibinfo {author} {\bibfnamefont {A.}~\bibnamefont
  {Smart}}, \emph {et~al.},\ }\bibfield  {title} {\bibinfo {title} {The pfam
  protein families database in 2019},\ }\href@noop {} {\bibfield  {journal}
  {\bibinfo  {journal} {Nucleic acids research}\ }\textbf {\bibinfo {volume}
  {47}},\ \bibinfo {pages} {D427} (\bibinfo {year} {2019})}\BibitemShut
  {NoStop}%
\bibitem [{\citenamefont {Baxter}(2016)}]{baxter2016exactly}%
  \BibitemOpen
  \bibfield  {author} {\bibinfo {author} {\bibfnamefont {R.~J.}\ \bibnamefont
  {Baxter}},\ }\href@noop {} {\emph {\bibinfo {title} {Exactly solved models in
  statistical mechanics}}}\ (\bibinfo  {publisher} {Elsevier},\ \bibinfo {year}
  {2016})\BibitemShut {NoStop}%
\bibitem [{\citenamefont {Yevick}\ and\ \citenamefont
  {Melko}(2021)}]{yevick2021accuracy}%
  \BibitemOpen
  \bibfield  {author} {\bibinfo {author} {\bibfnamefont {D.}~\bibnamefont
  {Yevick}}\ and\ \bibinfo {author} {\bibfnamefont {R.}~\bibnamefont {Melko}},\
  }\bibfield  {title} {\bibinfo {title} {The accuracy of restricted boltzmann
  machine models of ising systems},\ }\href@noop {} {\bibfield  {journal}
  {\bibinfo  {journal} {Computer Physics Communications}\ }\textbf {\bibinfo
  {volume} {258}},\ \bibinfo {pages} {107518} (\bibinfo {year}
  {2021})}\BibitemShut {NoStop}%
\bibitem [{\citenamefont {Harsh}\ \emph {et~al.}(2020)\citenamefont {Harsh},
  \citenamefont {Tubiana}, \citenamefont {Cocco},\ and\ \citenamefont
  {Monasson}}]{harsh2020place}%
  \BibitemOpen
  \bibfield  {author} {\bibinfo {author} {\bibfnamefont {M.}~\bibnamefont
  {Harsh}}, \bibinfo {author} {\bibfnamefont {J.}~\bibnamefont {Tubiana}},
  \bibinfo {author} {\bibfnamefont {S.}~\bibnamefont {Cocco}},\ and\ \bibinfo
  {author} {\bibfnamefont {R.}~\bibnamefont {Monasson}},\ }\bibfield  {title}
  {\bibinfo {title} {‘place-cell’emergence and learning of invariant data
  with restricted boltzmann machines: breaking and dynamical restoration of
  continuous symmetries in the weight space},\ }\href@noop {} {\bibfield
  {journal} {\bibinfo  {journal} {Journal of Physics A: Mathematical and
  Theoretical}\ }\textbf {\bibinfo {volume} {53}},\ \bibinfo {pages} {174002}
  (\bibinfo {year} {2020})}\BibitemShut {NoStop}%
\bibitem [{\citenamefont {Cover}(1999)}]{cover1999elements}%
  \BibitemOpen
  \bibfield  {author} {\bibinfo {author} {\bibfnamefont {T.~M.}\ \bibnamefont
  {Cover}},\ }\href@noop {} {\emph {\bibinfo {title} {Elements of information
  theory}}}\ (\bibinfo  {publisher} {John Wiley \& Sons},\ \bibinfo {year}
  {1999})\BibitemShut {NoStop}%
\bibitem [{\citenamefont {Engel}\ and\ \citenamefont {Van~den
  Broeck}(2001)}]{engel2001statistical}%
  \BibitemOpen
  \bibfield  {author} {\bibinfo {author} {\bibfnamefont {A.}~\bibnamefont
  {Engel}}\ and\ \bibinfo {author} {\bibfnamefont {C.}~\bibnamefont {Van~den
  Broeck}},\ }\href@noop {} {\emph {\bibinfo {title} {Statistical mechanics of
  learning}}}\ (\bibinfo  {publisher} {Cambridge University Press},\ \bibinfo
  {year} {2001})\BibitemShut {NoStop}%
\bibitem [{\citenamefont {Brenner}\ \emph {et~al.}(2000)\citenamefont
  {Brenner}, \citenamefont {Bialek},\ and\ \citenamefont
  {Van~Steveninck}}]{brenner2000adaptive}%
  \BibitemOpen
  \bibfield  {author} {\bibinfo {author} {\bibfnamefont {N.}~\bibnamefont
  {Brenner}}, \bibinfo {author} {\bibfnamefont {W.}~\bibnamefont {Bialek}},\
  and\ \bibinfo {author} {\bibfnamefont {R.~d.~R.}\ \bibnamefont
  {Van~Steveninck}},\ }\bibfield  {title} {\bibinfo {title} {Adaptive rescaling
  maximizes information transmission},\ }\href@noop {} {\bibfield  {journal}
  {\bibinfo  {journal} {Neuron}\ }\textbf {\bibinfo {volume} {26}},\ \bibinfo
  {pages} {695} (\bibinfo {year} {2000})}\BibitemShut {NoStop}%
\bibitem [{\citenamefont {Scholkopf}\ and\ \citenamefont
  {Smola}(2018)}]{scholkopf2018learning}%
  \BibitemOpen
  \bibfield  {author} {\bibinfo {author} {\bibfnamefont {B.}~\bibnamefont
  {Scholkopf}}\ and\ \bibinfo {author} {\bibfnamefont {A.~J.}\ \bibnamefont
  {Smola}},\ }\href@noop {} {\emph {\bibinfo {title} {Learning with kernels:
  support vector machines, regularization, optimization, and beyond}}}\
  (\bibinfo  {publisher} {MIT press},\ \bibinfo {year} {2018})\BibitemShut
  {NoStop}%
\bibitem [{\citenamefont {Marchenko}\ and\ \citenamefont
  {Pastur}(1967)}]{marchenko1967distribution}%
  \BibitemOpen
  \bibfield  {author} {\bibinfo {author} {\bibfnamefont {V.~A.}\ \bibnamefont
  {Marchenko}}\ and\ \bibinfo {author} {\bibfnamefont {L.~A.}\ \bibnamefont
  {Pastur}},\ }\bibfield  {title} {\bibinfo {title} {Distribution of
  eigenvalues for some sets of random matrices},\ }\href@noop {} {\bibfield
  {journal} {\bibinfo  {journal} {Matematicheskii Sbornik}\ }\textbf {\bibinfo
  {volume} {114}},\ \bibinfo {pages} {507} (\bibinfo {year}
  {1967})}\BibitemShut {NoStop}%
\bibitem [{\citenamefont {Decelle}\ \emph {et~al.}(2021)\citenamefont
  {Decelle}, \citenamefont {Furtlehner},\ and\ \citenamefont
  {Seoane}}]{decelle2021equilibrium}%
  \BibitemOpen
  \bibfield  {author} {\bibinfo {author} {\bibfnamefont {A.}~\bibnamefont
  {Decelle}}, \bibinfo {author} {\bibfnamefont {C.}~\bibnamefont
  {Furtlehner}},\ and\ \bibinfo {author} {\bibfnamefont {B.}~\bibnamefont
  {Seoane}},\ }\bibfield  {title} {\bibinfo {title} {Equilibrium and
  non-equilibrium regimes in the learning of restricted boltzmann machines},\
  }\href@noop {} {\bibfield  {journal} {\bibinfo  {journal} {Advances in Neural
  Information Processing Systems}\ }\textbf {\bibinfo {volume} {34}},\ \bibinfo
  {pages} {5345} (\bibinfo {year} {2021})}\BibitemShut {NoStop}%
\bibitem [{\citenamefont {Onsager}(1944)}]{onsager1944crystal}%
  \BibitemOpen
  \bibfield  {author} {\bibinfo {author} {\bibfnamefont {L.}~\bibnamefont
  {Onsager}},\ }\bibfield  {title} {\bibinfo {title} {Crystal statistics. i. a
  two-dimensional model with an order-disorder transition},\ }\href@noop {}
  {\bibfield  {journal} {\bibinfo  {journal} {Physical Review}\ }\textbf
  {\bibinfo {volume} {65}},\ \bibinfo {pages} {117} (\bibinfo {year}
  {1944})}\BibitemShut {NoStop}%
\bibitem [{\citenamefont {Newman}\ and\ \citenamefont
  {Barkema}(1999)}]{newman1999monte}%
  \BibitemOpen
  \bibfield  {author} {\bibinfo {author} {\bibfnamefont {M.~E.}\ \bibnamefont
  {Newman}}\ and\ \bibinfo {author} {\bibfnamefont {G.~T.}\ \bibnamefont
  {Barkema}},\ }\href@noop {} {\emph {\bibinfo {title} {Monte Carlo methods in
  statistical physics}}}\ (\bibinfo  {publisher} {Clarendon Press},\ \bibinfo
  {year} {1999})\BibitemShut {NoStop}%
\bibitem [{\citenamefont {Yoshioka}\ \emph {et~al.}(2019)\citenamefont
  {Yoshioka}, \citenamefont {Akagi},\ and\ \citenamefont
  {Katsura}}]{yoshioka2019transforming}%
  \BibitemOpen
  \bibfield  {author} {\bibinfo {author} {\bibfnamefont {N.}~\bibnamefont
  {Yoshioka}}, \bibinfo {author} {\bibfnamefont {Y.}~\bibnamefont {Akagi}},\
  and\ \bibinfo {author} {\bibfnamefont {H.}~\bibnamefont {Katsura}},\
  }\bibfield  {title} {\bibinfo {title} {Transforming generalized ising models
  into boltzmann machines},\ }\href@noop {} {\bibfield  {journal} {\bibinfo
  {journal} {Physical Review E}\ }\textbf {\bibinfo {volume} {99}},\ \bibinfo
  {pages} {032113} (\bibinfo {year} {2019})}\BibitemShut {NoStop}%
\bibitem [{\citenamefont {Cossu}\ \emph {et~al.}(2019)\citenamefont {Cossu},
  \citenamefont {Del~Debbio}, \citenamefont {Giani}, \citenamefont {Khamseh},\
  and\ \citenamefont {Wilson}}]{cossu2019machine}%
  \BibitemOpen
  \bibfield  {author} {\bibinfo {author} {\bibfnamefont {G.}~\bibnamefont
  {Cossu}}, \bibinfo {author} {\bibfnamefont {L.}~\bibnamefont {Del~Debbio}},
  \bibinfo {author} {\bibfnamefont {T.}~\bibnamefont {Giani}}, \bibinfo
  {author} {\bibfnamefont {A.}~\bibnamefont {Khamseh}},\ and\ \bibinfo {author}
  {\bibfnamefont {M.}~\bibnamefont {Wilson}},\ }\bibfield  {title} {\bibinfo
  {title} {Machine learning determination of dynamical parameters: The ising
  model case},\ }\href@noop {} {\bibfield  {journal} {\bibinfo  {journal}
  {Physical Review B}\ }\textbf {\bibinfo {volume} {100}},\ \bibinfo {pages}
  {064304} (\bibinfo {year} {2019})}\BibitemShut {NoStop}%
\bibitem [{\citenamefont {Selke}(2006)}]{selke2006critical}%
  \BibitemOpen
  \bibfield  {author} {\bibinfo {author} {\bibfnamefont {W.}~\bibnamefont
  {Selke}},\ }\bibfield  {title} {\bibinfo {title} {Critical binder cumulant of
  two-dimensional ising models},\ }\href@noop {} {\bibfield  {journal}
  {\bibinfo  {journal} {The European Physical Journal B-Condensed Matter and
  Complex Systems}\ }\textbf {\bibinfo {volume} {51}},\ \bibinfo {pages} {223}
  (\bibinfo {year} {2006})}\BibitemShut {NoStop}%
\bibitem [{\citenamefont {Shimagaki}\ and\ \citenamefont
  {Weigt}(2019)}]{shimagaki2019selection}%
  \BibitemOpen
  \bibfield  {author} {\bibinfo {author} {\bibfnamefont {K.}~\bibnamefont
  {Shimagaki}}\ and\ \bibinfo {author} {\bibfnamefont {M.}~\bibnamefont
  {Weigt}},\ }\bibfield  {title} {\bibinfo {title} {Selection of sequence
  motifs and generative hopfield-potts models for protein families},\
  }\href@noop {} {\bibfield  {journal} {\bibinfo  {journal} {Physical Review
  E}\ }\textbf {\bibinfo {volume} {100}},\ \bibinfo {pages} {032128} (\bibinfo
  {year} {2019})}\BibitemShut {NoStop}%
\bibitem [{\citenamefont {Sehnal}\ \emph {et~al.}(2021)\citenamefont {Sehnal},
  \citenamefont {Bittrich}, \citenamefont {Deshpande}, \citenamefont
  {Svobodov{\'a}}, \citenamefont {Berka}, \citenamefont {Bazgier},
  \citenamefont {Velankar}, \citenamefont {Burley}, \citenamefont
  {Ko{\v{c}}a},\ and\ \citenamefont {Rose}}]{sehnal2021mol}%
  \BibitemOpen
  \bibfield  {author} {\bibinfo {author} {\bibfnamefont {D.}~\bibnamefont
  {Sehnal}}, \bibinfo {author} {\bibfnamefont {S.}~\bibnamefont {Bittrich}},
  \bibinfo {author} {\bibfnamefont {M.}~\bibnamefont {Deshpande}}, \bibinfo
  {author} {\bibfnamefont {R.}~\bibnamefont {Svobodov{\'a}}}, \bibinfo {author}
  {\bibfnamefont {K.}~\bibnamefont {Berka}}, \bibinfo {author} {\bibfnamefont
  {V.}~\bibnamefont {Bazgier}}, \bibinfo {author} {\bibfnamefont
  {S.}~\bibnamefont {Velankar}}, \bibinfo {author} {\bibfnamefont {S.~K.}\
  \bibnamefont {Burley}}, \bibinfo {author} {\bibfnamefont {J.}~\bibnamefont
  {Ko{\v{c}}a}},\ and\ \bibinfo {author} {\bibfnamefont {A.~S.}\ \bibnamefont
  {Rose}},\ }\bibfield  {title} {\bibinfo {title} {Mol* viewer: modern web app
  for 3d visualization and analysis of large biomolecular structures},\
  }\href@noop {} {\bibfield  {journal} {\bibinfo  {journal} {Nucleic Acids
  Research}\ }\textbf {\bibinfo {volume} {49}},\ \bibinfo {pages} {W431}
  (\bibinfo {year} {2021})}\BibitemShut {NoStop}%
\bibitem [{\citenamefont {Grishin}(2001)}]{grishin2001kh}%
  \BibitemOpen
  \bibfield  {author} {\bibinfo {author} {\bibfnamefont {N.~V.}\ \bibnamefont
  {Grishin}},\ }\bibfield  {title} {\bibinfo {title} {Kh domain: one motif, two
  folds},\ }\href@noop {} {\bibfield  {journal} {\bibinfo  {journal} {Nucleic
  acids research}\ }\textbf {\bibinfo {volume} {29}},\ \bibinfo {pages} {638}
  (\bibinfo {year} {2001})}\BibitemShut {NoStop}%
\bibitem [{\citenamefont {Lunde}\ \emph {et~al.}(2007)\citenamefont {Lunde},
  \citenamefont {Moore},\ and\ \citenamefont {Varani}}]{lunde2007rna}%
  \BibitemOpen
  \bibfield  {author} {\bibinfo {author} {\bibfnamefont {B.~M.}\ \bibnamefont
  {Lunde}}, \bibinfo {author} {\bibfnamefont {C.}~\bibnamefont {Moore}},\ and\
  \bibinfo {author} {\bibfnamefont {G.}~\bibnamefont {Varani}},\ }\bibfield
  {title} {\bibinfo {title} {Rna-binding proteins: modular design for efficient
  function},\ }\href@noop {} {\bibfield  {journal} {\bibinfo  {journal} {Nature
  reviews Molecular cell biology}\ }\textbf {\bibinfo {volume} {8}},\ \bibinfo
  {pages} {479} (\bibinfo {year} {2007})}\BibitemShut {NoStop}%
\bibitem [{\citenamefont {Valverde}\ \emph {et~al.}(2008)\citenamefont
  {Valverde}, \citenamefont {Edwards},\ and\ \citenamefont
  {Regan}}]{valverde2008structure}%
  \BibitemOpen
  \bibfield  {author} {\bibinfo {author} {\bibfnamefont {R.}~\bibnamefont
  {Valverde}}, \bibinfo {author} {\bibfnamefont {L.}~\bibnamefont {Edwards}},\
  and\ \bibinfo {author} {\bibfnamefont {L.}~\bibnamefont {Regan}},\ }\bibfield
   {title} {\bibinfo {title} {Structure and function of kh domains},\
  }\href@noop {} {\bibfield  {journal} {\bibinfo  {journal} {The FEBS journal}\
  }\textbf {\bibinfo {volume} {275}},\ \bibinfo {pages} {2712} (\bibinfo {year}
  {2008})}\BibitemShut {NoStop}%
\bibitem [{\citenamefont {Musco}\ \emph {et~al.}(1996)\citenamefont {Musco},
  \citenamefont {Stier}, \citenamefont {Joseph}, \citenamefont {Morelli},
  \citenamefont {Nilges}, \citenamefont {Gibson},\ and\ \citenamefont
  {Pastore}}]{musco1996three}%
  \BibitemOpen
  \bibfield  {author} {\bibinfo {author} {\bibfnamefont {G.}~\bibnamefont
  {Musco}}, \bibinfo {author} {\bibfnamefont {G.}~\bibnamefont {Stier}},
  \bibinfo {author} {\bibfnamefont {C.}~\bibnamefont {Joseph}}, \bibinfo
  {author} {\bibfnamefont {M.~A.~C.}\ \bibnamefont {Morelli}}, \bibinfo
  {author} {\bibfnamefont {M.}~\bibnamefont {Nilges}}, \bibinfo {author}
  {\bibfnamefont {T.~J.}\ \bibnamefont {Gibson}},\ and\ \bibinfo {author}
  {\bibfnamefont {A.}~\bibnamefont {Pastore}},\ }\bibfield  {title} {\bibinfo
  {title} {Three-dimensional structure and stability of the kh domain:
  molecular insights into the fragile x syndrome},\ }\href@noop {} {\bibfield
  {journal} {\bibinfo  {journal} {Cell}\ }\textbf {\bibinfo {volume} {85}},\
  \bibinfo {pages} {237} (\bibinfo {year} {1996})}\BibitemShut {NoStop}%
\bibitem [{\citenamefont {O'Donnell}\ and\ \citenamefont
  {Warren}(2002)}]{kh2002decade}%
  \BibitemOpen
  \bibfield  {author} {\bibinfo {author} {\bibfnamefont {W.~T.}\ \bibnamefont
  {O'Donnell}}\ and\ \bibinfo {author} {\bibfnamefont {S.~T.}\ \bibnamefont
  {Warren}},\ }\bibfield  {title} {\bibinfo {title} {A decade of molecular
  studies of fragile x syndrome},\ }\href@noop {} {\bibfield  {journal}
  {\bibinfo  {journal} {Annual review of neuroscience}\ }\textbf {\bibinfo
  {volume} {25}},\ \bibinfo {pages} {315} (\bibinfo {year} {2002})}\BibitemShut
  {NoStop}%
\bibitem [{\citenamefont {Consortium}(2020)}]{uniprot2020}%
  \BibitemOpen
  \bibfield  {author} {\bibinfo {author} {\bibfnamefont {T.~U.}\ \bibnamefont
  {Consortium}},\ }\bibfield  {title} {\bibinfo {title} {{UniProt: the
  universal protein knowledgebase in 2021}},\ }\href
  {https://doi.org/10.1093/nar/gkaa1100} {\bibfield  {journal} {\bibinfo
  {journal} {Nucleic Acids Research}\ }\textbf {\bibinfo {volume} {49}},\
  \bibinfo {pages} {D480} (\bibinfo {year} {2020})},\ \Eprint
  {https://arxiv.org/abs/https://academic.oup.com/nar/article-pdf/49/D1/D480/35364103/gkaa1100.pdf}
  {https://academic.oup.com/nar/article-pdf/49/D1/D480/35364103/gkaa1100.pdf}
  \BibitemShut {NoStop}%
\bibitem [{\citenamefont {Cocco}\ \emph {et~al.}(2018)\citenamefont {Cocco},
  \citenamefont {Feinauer}, \citenamefont {Figliuzzi}, \citenamefont
  {Monasson},\ and\ \citenamefont {Weigt}}]{cocco2018inverse}%
  \BibitemOpen
  \bibfield  {author} {\bibinfo {author} {\bibfnamefont {S.}~\bibnamefont
  {Cocco}}, \bibinfo {author} {\bibfnamefont {C.}~\bibnamefont {Feinauer}},
  \bibinfo {author} {\bibfnamefont {M.}~\bibnamefont {Figliuzzi}}, \bibinfo
  {author} {\bibfnamefont {R.}~\bibnamefont {Monasson}},\ and\ \bibinfo
  {author} {\bibfnamefont {M.}~\bibnamefont {Weigt}},\ }\bibfield  {title}
  {\bibinfo {title} {Inverse statistical physics of protein sequences: a key
  issues review},\ }\href@noop {} {\bibfield  {journal} {\bibinfo  {journal}
  {Reports on Progress in Physics}\ }\textbf {\bibinfo {volume} {81}},\
  \bibinfo {pages} {032601} (\bibinfo {year} {2018})}\BibitemShut {NoStop}%
\bibitem [{\citenamefont {Morcos}\ \emph {et~al.}(2011)\citenamefont {Morcos},
  \citenamefont {Pagnani}, \citenamefont {Lunt}, \citenamefont {Bertolino},
  \citenamefont {Marks}, \citenamefont {Sander}, \citenamefont {Zecchina},
  \citenamefont {Onuchic}, \citenamefont {Hwa},\ and\ \citenamefont
  {Weigt}}]{morcos2011direct}%
  \BibitemOpen
  \bibfield  {author} {\bibinfo {author} {\bibfnamefont {F.}~\bibnamefont
  {Morcos}}, \bibinfo {author} {\bibfnamefont {A.}~\bibnamefont {Pagnani}},
  \bibinfo {author} {\bibfnamefont {B.}~\bibnamefont {Lunt}}, \bibinfo {author}
  {\bibfnamefont {A.}~\bibnamefont {Bertolino}}, \bibinfo {author}
  {\bibfnamefont {D.~S.}\ \bibnamefont {Marks}}, \bibinfo {author}
  {\bibfnamefont {C.}~\bibnamefont {Sander}}, \bibinfo {author} {\bibfnamefont
  {R.}~\bibnamefont {Zecchina}}, \bibinfo {author} {\bibfnamefont {J.~N.}\
  \bibnamefont {Onuchic}}, \bibinfo {author} {\bibfnamefont {T.}~\bibnamefont
  {Hwa}},\ and\ \bibinfo {author} {\bibfnamefont {M.}~\bibnamefont {Weigt}},\
  }\bibfield  {title} {\bibinfo {title} {Direct-coupling analysis of residue
  coevolution captures native contacts across many protein families},\
  }\href@noop {} {\bibfield  {journal} {\bibinfo  {journal} {Proceedings of the
  National Academy of Sciences}\ }\textbf {\bibinfo {volume} {108}},\ \bibinfo
  {pages} {E1293} (\bibinfo {year} {2011})}\BibitemShut {NoStop}%
\bibitem [{\citenamefont {Mirdita}\ \emph {et~al.}(2022)\citenamefont
  {Mirdita}, \citenamefont {Sch{\"u}tze}, \citenamefont {Moriwaki},
  \citenamefont {Heo}, \citenamefont {Ovchinnikov},\ and\ \citenamefont
  {Steinegger}}]{mirdita2022colabfold}%
  \BibitemOpen
  \bibfield  {author} {\bibinfo {author} {\bibfnamefont {M.}~\bibnamefont
  {Mirdita}}, \bibinfo {author} {\bibfnamefont {K.}~\bibnamefont
  {Sch{\"u}tze}}, \bibinfo {author} {\bibfnamefont {Y.}~\bibnamefont
  {Moriwaki}}, \bibinfo {author} {\bibfnamefont {L.}~\bibnamefont {Heo}},
  \bibinfo {author} {\bibfnamefont {S.}~\bibnamefont {Ovchinnikov}},\ and\
  \bibinfo {author} {\bibfnamefont {M.}~\bibnamefont {Steinegger}},\ }\bibfield
   {title} {\bibinfo {title} {Colabfold: making protein folding accessible to
  all},\ }\href@noop {} {\bibfield  {journal} {\bibinfo  {journal} {Nature
  Methods}\ ,\ \bibinfo {pages} {1}} (\bibinfo {year} {2022})}\BibitemShut
  {NoStop}%
\bibitem [{\citenamefont {Neal}(1998)}]{neal1998annealed}%
  \BibitemOpen
  \bibfield  {author} {\bibinfo {author} {\bibfnamefont {R.}~\bibnamefont
  {Neal}},\ }\bibfield  {title} {\bibinfo {title} {Annealed importance sampling
  (technical report 9805 (revised))},\ }\href@noop {} {\bibfield  {journal}
  {\bibinfo  {journal} {Department of Statistics, University of Toronto}\ }
  (\bibinfo {year} {1998})}\BibitemShut {NoStop}%
\bibitem [{\citenamefont {Burda}\ \emph {et~al.}(2015)\citenamefont {Burda},
  \citenamefont {Grosse},\ and\ \citenamefont
  {Salakhutdinov}}]{burda2015accurate}%
  \BibitemOpen
  \bibfield  {author} {\bibinfo {author} {\bibfnamefont {Y.}~\bibnamefont
  {Burda}}, \bibinfo {author} {\bibfnamefont {R.}~\bibnamefont {Grosse}},\ and\
  \bibinfo {author} {\bibfnamefont {R.}~\bibnamefont {Salakhutdinov}},\
  }\bibfield  {title} {\bibinfo {title} {Accurate and conservative estimates of
  mrf log-likelihood using reverse annealing},\ }in\ \href@noop {} {\emph
  {\bibinfo {booktitle} {Artificial Intelligence and Statistics}}}\ (\bibinfo
  {organization} {PMLR},\ \bibinfo {year} {2015})\ pp.\ \bibinfo {pages}
  {102--110}\BibitemShut {NoStop}%
\bibitem [{\citenamefont {Higgins}\ \emph {et~al.}(2016)\citenamefont
  {Higgins}, \citenamefont {Matthey}, \citenamefont {Pal}, \citenamefont
  {Burgess}, \citenamefont {Glorot}, \citenamefont {Botvinick}, \citenamefont
  {Mohamed},\ and\ \citenamefont {Lerchner}}]{higgins2016beta}%
  \BibitemOpen
  \bibfield  {author} {\bibinfo {author} {\bibfnamefont {I.}~\bibnamefont
  {Higgins}}, \bibinfo {author} {\bibfnamefont {L.}~\bibnamefont {Matthey}},
  \bibinfo {author} {\bibfnamefont {A.}~\bibnamefont {Pal}}, \bibinfo {author}
  {\bibfnamefont {C.}~\bibnamefont {Burgess}}, \bibinfo {author} {\bibfnamefont
  {X.}~\bibnamefont {Glorot}}, \bibinfo {author} {\bibfnamefont
  {M.}~\bibnamefont {Botvinick}}, \bibinfo {author} {\bibfnamefont
  {S.}~\bibnamefont {Mohamed}},\ and\ \bibinfo {author} {\bibfnamefont
  {A.}~\bibnamefont {Lerchner}},\ }\bibfield  {title} {\bibinfo {title} {Beta
  {VAE}: Learning basic visual concepts with a constrained variational
  framework},\ }\href@noop {} {\bibfield  {journal} {\bibinfo  {journal}
  {ICLR}\ } (\bibinfo {year} {2016})}\BibitemShut {NoStop}%
\bibitem [{\citenamefont {Goldt}\ \emph {et~al.}(2022)\citenamefont {Goldt},
  \citenamefont {Loureiro}, \citenamefont {Reeves}, \citenamefont {Krzakala},
  \citenamefont {M{\'e}zard},\ and\ \citenamefont
  {Zdeborov{\'a}}}]{goldt2022gaussian}%
  \BibitemOpen
  \bibfield  {author} {\bibinfo {author} {\bibfnamefont {S.}~\bibnamefont
  {Goldt}}, \bibinfo {author} {\bibfnamefont {B.}~\bibnamefont {Loureiro}},
  \bibinfo {author} {\bibfnamefont {G.}~\bibnamefont {Reeves}}, \bibinfo
  {author} {\bibfnamefont {F.}~\bibnamefont {Krzakala}}, \bibinfo {author}
  {\bibfnamefont {M.}~\bibnamefont {M{\'e}zard}},\ and\ \bibinfo {author}
  {\bibfnamefont {L.}~\bibnamefont {Zdeborov{\'a}}},\ }\bibfield  {title}
  {\bibinfo {title} {The gaussian equivalence of generative models for learning
  with shallow neural networks},\ }in\ \href@noop {} {\emph {\bibinfo
  {booktitle} {Mathematical and Scientific Machine Learning}}}\ (\bibinfo
  {organization} {PMLR},\ \bibinfo {year} {2022})\ pp.\ \bibinfo {pages}
  {426--471}\BibitemShut {NoStop}%
\bibitem [{\citenamefont {Tieleman}(2008)}]{tieleman2008training}%
  \BibitemOpen
  \bibfield  {author} {\bibinfo {author} {\bibfnamefont {T.}~\bibnamefont
  {Tieleman}},\ }\bibfield  {title} {\bibinfo {title} {Training restricted
  boltzmann machines using approximations to the likelihood gradient},\ }in\
  \href@noop {} {\emph {\bibinfo {booktitle} {Proceedings of the 25th
  international conference on Machine learning}}}\ (\bibinfo {year} {2008})\
  pp.\ \bibinfo {pages} {1064--1071}\BibitemShut {NoStop}%
\bibitem [{\citenamefont {Belitz}\ \emph {et~al.}(2005)\citenamefont {Belitz},
  \citenamefont {Kirkpatrick},\ and\ \citenamefont
  {Vojta}}]{belitz2005generic}%
  \BibitemOpen
  \bibfield  {author} {\bibinfo {author} {\bibfnamefont {D.}~\bibnamefont
  {Belitz}}, \bibinfo {author} {\bibfnamefont {T.}~\bibnamefont
  {Kirkpatrick}},\ and\ \bibinfo {author} {\bibfnamefont {T.}~\bibnamefont
  {Vojta}},\ }\bibfield  {title} {\bibinfo {title} {How generic scale
  invariance influences quantum and classical phase transitions},\ }\href@noop
  {} {\bibfield  {journal} {\bibinfo  {journal} {Reviews of modern physics}\
  }\textbf {\bibinfo {volume} {77}},\ \bibinfo {pages} {579} (\bibinfo {year}
  {2005})}\BibitemShut {NoStop}%
\bibitem [{\citenamefont {Aron}\ and\ \citenamefont
  {Kulkarni}(2020)}]{aron2020nonanalytic}%
  \BibitemOpen
  \bibfield  {author} {\bibinfo {author} {\bibfnamefont {C.}~\bibnamefont
  {Aron}}\ and\ \bibinfo {author} {\bibfnamefont {M.}~\bibnamefont
  {Kulkarni}},\ }\bibfield  {title} {\bibinfo {title} {Nonanalytic
  nonequilibrium field theory: Stochastic reheating of the ising model},\
  }\href@noop {} {\bibfield  {journal} {\bibinfo  {journal} {Physical Review
  Research}\ }\textbf {\bibinfo {volume} {2}},\ \bibinfo {pages} {043390}
  (\bibinfo {year} {2020})}\BibitemShut {NoStop}%
\bibitem [{\citenamefont {Aron}\ and\ \citenamefont
  {Chamon}(2020)}]{aron2020landau}%
  \BibitemOpen
  \bibfield  {author} {\bibinfo {author} {\bibfnamefont {C.}~\bibnamefont
  {Aron}}\ and\ \bibinfo {author} {\bibfnamefont {C.}~\bibnamefont {Chamon}},\
  }\bibfield  {title} {\bibinfo {title} {Landau theory for non-equilibrium
  steady states},\ }\href@noop {} {\bibfield  {journal} {\bibinfo  {journal}
  {SciPost Physics}\ }\textbf {\bibinfo {volume} {8}},\ \bibinfo {pages} {074}
  (\bibinfo {year} {2020})}\BibitemShut {NoStop}%
\bibitem [{\citenamefont {Rube}\ \emph {et~al.}(2022)\citenamefont {Rube},
  \citenamefont {Rastogi}, \citenamefont {Feng}, \citenamefont {Kribelbauer},
  \citenamefont {Li}, \citenamefont {Becerra}, \citenamefont {Melo},
  \citenamefont {Do}, \citenamefont {Li}, \citenamefont {Adam} \emph
  {et~al.}}]{rube2022prediction}%
  \BibitemOpen
  \bibfield  {author} {\bibinfo {author} {\bibfnamefont {H.~T.}\ \bibnamefont
  {Rube}}, \bibinfo {author} {\bibfnamefont {C.}~\bibnamefont {Rastogi}},
  \bibinfo {author} {\bibfnamefont {S.}~\bibnamefont {Feng}}, \bibinfo {author}
  {\bibfnamefont {J.~F.}\ \bibnamefont {Kribelbauer}}, \bibinfo {author}
  {\bibfnamefont {A.}~\bibnamefont {Li}}, \bibinfo {author} {\bibfnamefont
  {B.}~\bibnamefont {Becerra}}, \bibinfo {author} {\bibfnamefont {L.~A.}\
  \bibnamefont {Melo}}, \bibinfo {author} {\bibfnamefont {B.~V.}\ \bibnamefont
  {Do}}, \bibinfo {author} {\bibfnamefont {X.}~\bibnamefont {Li}}, \bibinfo
  {author} {\bibfnamefont {H.~H.}\ \bibnamefont {Adam}}, \emph {et~al.},\
  }\bibfield  {title} {\bibinfo {title} {Prediction of protein--ligand binding
  affinity from sequencing data with interpretable machine learning},\
  }\href@noop {} {\bibfield  {journal} {\bibinfo  {journal} {Nature
  Biotechnology}\ ,\ \bibinfo {pages} {1}} (\bibinfo {year}
  {2022})}\BibitemShut {NoStop}%
\bibitem [{\citenamefont {Tubiana}\ and\ \citenamefont
  {Monasson}(2017)}]{tubiana2017}%
  \BibitemOpen
  \bibfield  {author} {\bibinfo {author} {\bibfnamefont {J.}~\bibnamefont
  {Tubiana}}\ and\ \bibinfo {author} {\bibfnamefont {R.}~\bibnamefont
  {Monasson}},\ }\bibfield  {title} {\bibinfo {title} {Emergence of
  compositional representations in restricted boltzmann machines},\ }\href
  {https://doi.org/10.1103/PhysRevLett.118.138301} {\bibfield  {journal}
  {\bibinfo  {journal} {Phys. Rev. Lett.}\ }\textbf {\bibinfo {volume} {118}},\
  \bibinfo {pages} {138301} (\bibinfo {year} {2017})}\BibitemShut {NoStop}%
\bibitem [{\citenamefont {Kingma}\ and\ \citenamefont
  {Ba}(2014)}]{kingma2014adam}%
  \BibitemOpen
  \bibfield  {author} {\bibinfo {author} {\bibfnamefont {D.~P.}\ \bibnamefont
  {Kingma}}\ and\ \bibinfo {author} {\bibfnamefont {J.}~\bibnamefont {Ba}},\
  }\bibfield  {title} {\bibinfo {title} {Adam: A method for stochastic
  optimization},\ }\href@noop {} {\bibfield  {journal} {\bibinfo  {journal}
  {arXiv preprint arXiv:1412.6980}\ } (\bibinfo {year} {2014})}\BibitemShut
  {NoStop}%
\bibitem [{\citenamefont {Melchior}\ \emph {et~al.}(2016)\citenamefont
  {Melchior}, \citenamefont {Fischer},\ and\ \citenamefont
  {Wiskott}}]{melchior2016center}%
  \BibitemOpen
  \bibfield  {author} {\bibinfo {author} {\bibfnamefont {J.}~\bibnamefont
  {Melchior}}, \bibinfo {author} {\bibfnamefont {A.}~\bibnamefont {Fischer}},\
  and\ \bibinfo {author} {\bibfnamefont {L.}~\bibnamefont {Wiskott}},\
  }\bibfield  {title} {\bibinfo {title} {How to center deep boltzmann
  machines},\ }\href@noop {} {\bibfield  {journal} {\bibinfo  {journal} {The
  Journal of Machine Learning Research}\ }\textbf {\bibinfo {volume} {17}},\
  \bibinfo {pages} {3387} (\bibinfo {year} {2016})}\BibitemShut {NoStop}%
\bibitem [{\citenamefont {Salakhutdinov}(2008)}]{salakhutdinov2008learning}%
  \BibitemOpen
  \bibfield  {author} {\bibinfo {author} {\bibfnamefont {R.}~\bibnamefont
  {Salakhutdinov}},\ }\bibfield  {title} {\bibinfo {title} {Learning and
  evaluating boltzmann machines},\ }\href@noop {} {\bibfield  {journal}
  {\bibinfo  {journal} {Utml Tr}\ }\textbf {\bibinfo {volume} {2}},\ \bibinfo
  {pages} {21} (\bibinfo {year} {2008})}\BibitemShut {NoStop}%
\bibitem [{\citenamefont {Sauvola}\ and\ \citenamefont
  {Pietik{\"a}inen}(2000)}]{sauvola2000adaptive}%
  \BibitemOpen
  \bibfield  {author} {\bibinfo {author} {\bibfnamefont {J.}~\bibnamefont
  {Sauvola}}\ and\ \bibinfo {author} {\bibfnamefont {M.}~\bibnamefont
  {Pietik{\"a}inen}},\ }\bibfield  {title} {\bibinfo {title} {Adaptive document
  image binarization},\ }\href@noop {} {\bibfield  {journal} {\bibinfo
  {journal} {Pattern recognition}\ }\textbf {\bibinfo {volume} {33}},\ \bibinfo
  {pages} {225} (\bibinfo {year} {2000})}\BibitemShut {NoStop}%
\bibitem [{Ima(2022)}]{ImageBinarization_jl}%
  \BibitemOpen
  \href@noop {} {\bibinfo {title} {{ImageBinarization.jl Julia package}}},\
  \bibinfo {howpublished}
  {\url{https://github.com/JuliaImages/ImageBinarization.jl}} (\bibinfo {year}
  {2022})\BibitemShut {NoStop}%
\bibitem [{\citenamefont {Klambauer}\ \emph {et~al.}(2017)\citenamefont
  {Klambauer}, \citenamefont {Unterthiner}, \citenamefont {Mayr},\ and\
  \citenamefont {Hochreiter}}]{klambauer2017self}%
  \BibitemOpen
  \bibfield  {author} {\bibinfo {author} {\bibfnamefont {G.}~\bibnamefont
  {Klambauer}}, \bibinfo {author} {\bibfnamefont {T.}~\bibnamefont
  {Unterthiner}}, \bibinfo {author} {\bibfnamefont {A.}~\bibnamefont {Mayr}},\
  and\ \bibinfo {author} {\bibfnamefont {S.}~\bibnamefont {Hochreiter}},\
  }\bibfield  {title} {\bibinfo {title} {Self-normalizing neural networks},\
  }\href@noop {} {\bibfield  {journal} {\bibinfo  {journal} {Advances in neural
  information processing systems}\ }\textbf {\bibinfo {volume} {30}} (\bibinfo
  {year} {2017})}\BibitemShut {NoStop}%
\bibitem [{\citenamefont {Zhang}\ and\ \citenamefont
  {Skolnick}(2004)}]{zhang2004scoring}%
  \BibitemOpen
  \bibfield  {author} {\bibinfo {author} {\bibfnamefont {Y.}~\bibnamefont
  {Zhang}}\ and\ \bibinfo {author} {\bibfnamefont {J.}~\bibnamefont
  {Skolnick}},\ }\bibfield  {title} {\bibinfo {title} {Scoring function for
  automated assessment of protein structure template quality},\ }\href@noop {}
  {\bibfield  {journal} {\bibinfo  {journal} {Proteins: Structure, Function,
  and Bioinformatics}\ }\textbf {\bibinfo {volume} {57}},\ \bibinfo {pages}
  {702} (\bibinfo {year} {2004})}\BibitemShut {NoStop}%
\end{thebibliography}%

\clearpage
\appendix
\onecolumngrid
\section*{Supplemental Material}

\begin{center}
\LARGE{
Disentangling representations in Restricted Boltzmann Machines \\ without adversaries}\\
\normalsize
\vspace{1cm}
Jorge Fernandez-de-Cossio-Diaz, Simona Cocco, R\'emi Monasson
\end{center}

\setcounter{page}{1}
\setcounter{table}{0}
\renewcommand{\thetable}{S\arabic{table}}%
\setcounter{figure}{0}
\renewcommand{\thefigure}{S\arabic{figure}}%

\section{Implementation details} \label{SI:sec:impl}

\subsection{Gibbs sampling}

One important property of RBMs is that the conditional distributions $P(\mathbf{h}|\mathbf{v})$, $P(\mathbf{v}|\mathbf{h})$ factorize,
\begin{align}
P(\mathbf{h}|\mathbf{v}) &\propto
\prod_\mu \mathrm{e}^{-\mathcal{U}_\mu(h_\mu) + \sum_i w_{i\mu} v_i h_\mu}
\label{eq:P(h|v)} \\
P(\mathbf{v}|\mathbf{h}) &\propto
\prod_i \mathrm{e}^{-\mathcal{V}_i(v_i) + \sum_\mu w_{i\mu} v_i h_\mu}
\label{eq:P(v|h)}
\end{align}
and therefore are easy to sample. They are important because $P(\mathbf{h}|\mathbf{v})$ allows us to map points $\mathbf{v}$ in data-space to their stochastic representations $\mathbf{h}$, while $P(\mathbf{v}|\mathbf{h})$ allows us to reconstruct a data point from its representation $\mathbf{h}$. The Gibbs algorithm for sampling from the RBM, consists of the following steps:
\begin{itemize}
\item Start from an initial configuration $\mathbf v_0$ in data-space. This can be random, or a data point.
\item For $t$ in $1, \dots, T$, where $T$ is the total number of steps, repeat the following steps:
\begin{itemize}
    \item Sample $\mathbf h_t$ using \eqref{eq:P(h|v)}, conditioned on $\mathbf v_{t-1}$.
    \item Sample $\mathbf v_t$ using \eqref{eq:P(v|h)}, conditioned on $\mathbf h_t$.
\end{itemize}
\item Return the last sample obtained, $\mathbf v_T$, where $T$ is the number of steps taken.
\end{itemize}
For large enough $T$, the resulting sample $\mathbf v_T$ approaches an equilibrium sample from the RBM \cite{newman1999monte}.

\subsection{Training algorithm for the standard RBM}

Taking the gradient of the likelihood (Equation \eqref{eq:Lrbm} in the main text) with respect to a generic parameter $\omega$ of the RBM, results in a moment-matching condition \cite{hinton2012practical}:
\begin{equation}\label{SI:eq:likelihood-gradient}
\frac{\partial\mathcal L}{\partial \omega} =
\left\langle\frac{\partial E}{\partial\omega}\right\rangle -
\left\langle\frac{\partial E}{\partial\omega}\right\rangle_\mathcal{D}
\end{equation}
where the right-hand side expectations $\langle\cdot\rangle$ are taken under the model distribution, and the left-hand side $\langle\cdot\rangle_\mathcal{D}$ under the empirical data distribution. The model can be trained by gradient ascent, where the parameters are updated according to
\begin{equation}
\theta \rightarrow \theta + \eta \frac{\partial\mathcal L}{\partial \omega}
\label{eq:gradient-ascent}
\end{equation}
with a suitable small learning rate $\eta$. This requires computing the averages $\langle\cdot\rangle$ over the model distribution, which can be computationally difficult. In practice, we use the persistent contrastive divergence algorithm \cite{tieleman2008training}, whereby a number $K=100$ of Markov chains are sampled from the model and updated by Gibbs sampling after each parameter update. These chains are then used to compute the averages $\langle\cdot\rangle$ over the model. The data average $\langle\cdot\rangle_\mathcal{D}$ is also estimated on mini-batches sampled from the data, also of size $K$. Finally, we combine \eqref{eq:gradient-ascent} with a number of tricks to speedup convergence:
\begin{itemize}
\item The RBM is initialized so that the visible units means match the averages computed on the data, while the weights are initialized to small random Gaussian values, with a standard deviation equal to $0.1 / \sqrt{N}$, where $N$ is then number of visible units.
\item We combine \eqref{eq:gradient-ascent} with a momentum term and an adaptive learning rate \cite{hinton2012practical}. This results in the ADAM optimization algorithm \cite{kingma2014adam}.
\item We use the so-called \emph{centering trick}, whereby gradients in the weights are estimated using centered moments. See \cite{melchior2016center} for details.
\item For the CelebA dataset, we found that during training some hidden units saturated in a always on or always off state. To resolve this, we checked after every 5 epochs for hidden units in this condition, and we dynamically disconnected these units (reset their weights to small random Gaussian distributed values with standard deviation $0.1/\sqrt{N}$), adding their contributions to the visible fields, and resumed training. This procedure allowed the previously stuck hidden units to learn again from a meaningful gradient.
\end{itemize}

\subsection{Training algorithm with linear constraints}

Under linear constraints (Equation \eqref{eq:Worth} in the main text), we modify the training algorithm as follows. After each parameter update \eqref{eq:gradient-ascent}, we project the weights $\mathbf W \rightarrow \mathbf P\mathbf W$ to ensure the constraint \eqref{eq:Worth} is still satisfied, where
\begin{equation}
\mathbf P = \mathbb I - \frac{\mathbf q^{(1)} {(\mathbf q^{(1)})}^\top}{{(\mathbf q^{(1))}}^\top \mathbf q^{(1)}}
\end{equation}
If the constraint applies to a subset of hidden units, we project only the constrained columns of $\mathbf W$.

\subsection{Training algorithm with quadratic constraints}

As explained in the main text, the quadratic constraints (Equation \eqref{eq:Worth2} from the main text) are implemented in practice by adding a penalty term to the log-likelihood,
\begin{equation}
\mathcal L - \chi^{(2)} \|\mathbf W^\top\mathbf q^{(2)}\mathbf W)\|^2
\end{equation}
where $\chi^{(2)}\ge0$ is a penalty weight, which we set to 100 in the experiments conducted in the paper (similar results were obtained for $\chi^{(2)}=10$ and $\chi^{(2)}=1000$). The additional gradient coming from this term evaluates:
\begin{equation}
\frac{1}{2}\frac{\partial}{\partial\mathbf W}\|\mathbf W^\top\mathbf q^{(2)}\mathbf W)\|^2 = 2\mathbf q^{(2)}\mathbf W\mathbf W^\top\mathbf q^{(2)}\mathbf W
\end{equation}
We then subtract this times $\chi^{(2)}$ from \eqref{SI:eq:likelihood-gradient} before each parameter update.

\subsection{Regularization}

A small $L2$ regularization is added to the RBM weights during training.
\begin{equation}
\frac{\gamma_{L2}}{2}\|\mathbf W\|^2 = \frac{\gamma_{L2}}{2}\sum_{i\mu} w_{i\mu}^2
\end{equation}
We use $\gamma_{L2}=0.007$ in our tests. The objective gradient in each weight is then modified by subtracting $\gamma_{L2} w_{i\mu}$.

\subsection{Further details about the different RBMs used} \label{SI:sec:RBM-architecture}

\begin{table}[h!]
\centering
\begin{tabular}{||c c c c c||} 
 \hline
 Dataset & Visible units & Hidden units & $L2$ reg. & Train iters. \\ [0.5ex] 
 \hline\hline
 CelebA & $64\times64$ (binary) & 5000 (binary) & 0.001 & 500000  \\
 Ising & $64\times64$ (spin) & 400 (spin) & 0.001 & 50000  \\
 MNIST & $28\times28$ (binary) & 400 (binary) & 0.001 & 50000  \\
 PF00013 & $21\times62$ (onehot) & 400 (binary) & 0.001 & 50000  \\
 [1ex]
 \hline
\end{tabular}
\caption{Details on the architecture of the RBMs used for the different datasets.}
\label{table:SI-RBM-arch}
\end{table}

Table \ref{table:SI-RBM-arch} summarizes the number and kind of visible and hidden units used for each of the datasets, as well as the value of the $L2$-regularization weight used for the weights. The visible units are arranged in a two-dimensional grid of the dimensionalities indicated.

The duration of training was set by counting the number of parameter updates (\emph{i.e.}, the number of times \eqref{eq:gradient-ascent} or its version with adaptive momentum was applied). More precisely, for a given number of epochs, the number of iterations is given by:
\begin{equation}
\textrm{number iters} = \textrm{number epochs} \times \textrm{number data points} / \textrm{mini-batch size}
\end{equation}
This number is reported in the column `Train iters.' of the table \ref{table:SI-RBM-arch}.

\subsection{Annealed importance sampling to estimate the likelihood}

Computing the likelihood of data in the RBM requires evaluating the partition function, but the exact computation of the partition function of the RBM is intractable in general. To get around this problem we can use annealed importance sampling (AIS), see \cite{salakhutdinov2008learning} for details.

\begin{figure}
\centering
\includegraphics[width=0.6\linewidth]{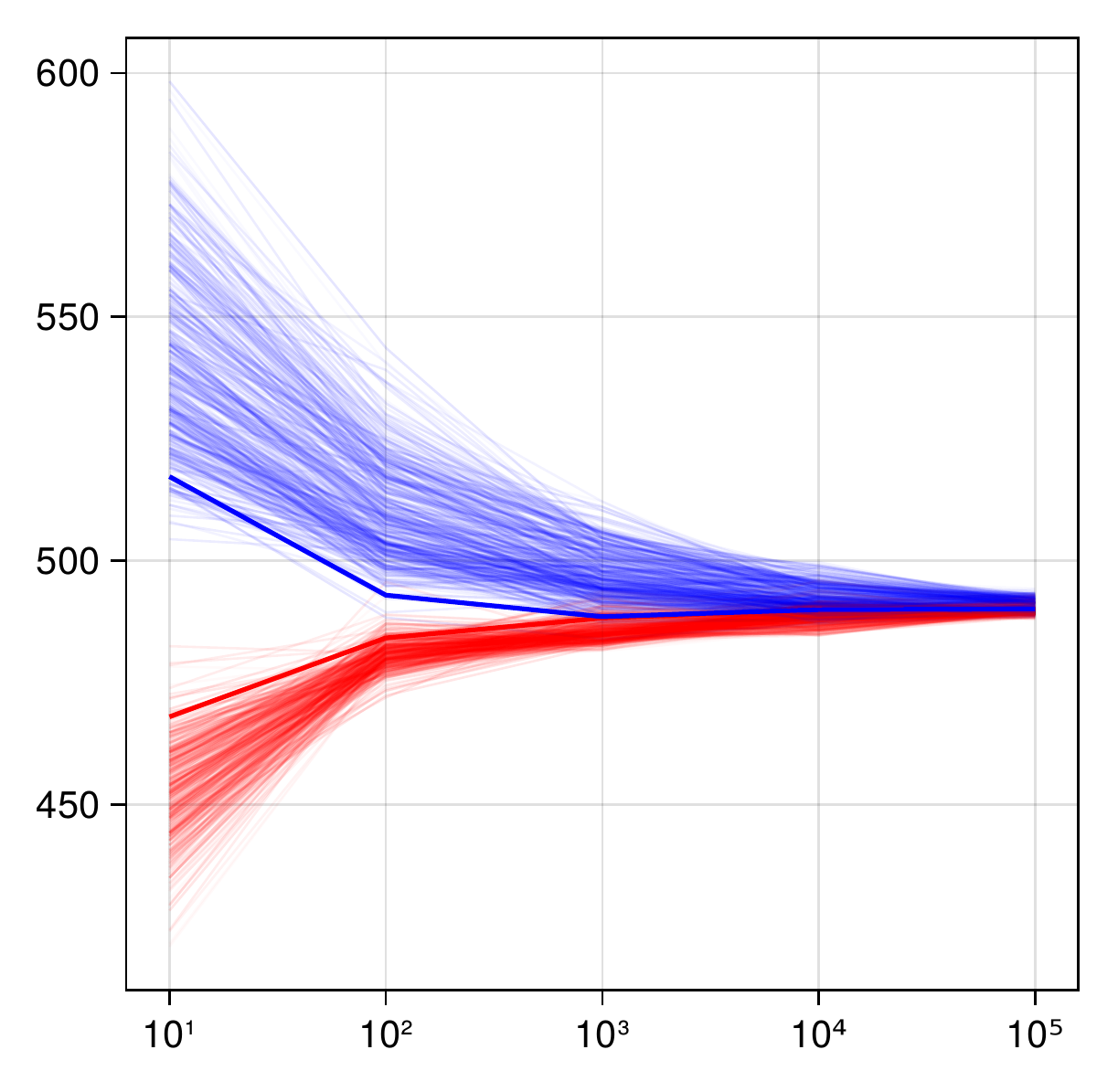}
\caption{\label{SI:fig:aisraise}\textbf{AIS/RAISE estimates sandwiching the partition function.} Estimation of the partition function of an RBM trained on MNIST0/1. The $x$-axis shows the number of interpolation distributions used. The $y$-axis is the estimated value of the log-partition function, for each sample. Here 100 samples were taken (thin lines), and their average (computed with the log-mean-exp trick) is shown (thick line). Red are the AIS estimates, which tend to be lower bounds, and blue are the RAISE estimates, which tend to be upper bounds.}
\end{figure}

AIS tends to produce stochastic lower bounds of the log-partition function. A related procedure, called reverse annealed importance sampling (RAISE) \cite{burda2015accurate}, can be used to obtain stochastic upper bounds. Combining the two then sandwiches the likelihhood, and allows us to assess the convergence of the procedure. An example is shown in Figure \ref{SI:fig:aisraise}.

\subsection{Sampling from the Ising model}

Ising model configurations were sampled using a mixture of Metropolis and Wolff algorithms \cite{newman1999monte}. Which kind of step to take at each iteration is determined dynamically, by tracking the number of spins moved in average by each type of sampler, and selecting the one more likely to move more spins. Note that detailed balance is secured, since both kind of moves satisfy this property.

\section{Details on the CelebA celebrity image dataset} \label{SI:sec:celeba-processing}

Color images from the CelebA dataset \cite{liu2015faceattributes} were first converted to grayscale by taking the average of the three color channels. Then, gray-scale intensities were binarized to black and white pixels following the Sauvola–Pietikäinen adaptive image binarization algorithm \cite{sauvola2000adaptive} (with the Julia implementation available in \cite{ImageBinarization_jl}).

The Sauvola–Pietikäinen algorithm defines an adaptive threshold $T(x,y)$ for each pixel $(x,y)$, and then sets the pixel to one if its intensity is $> T(x,y)$, and otherwise sets it to zero. The threshold is computed as follows:
\begin{equation}
T(x,y) = m(x,y)\left[1 + k\left(\frac{s(x,y)}{R} - 1\right)\right]
\end{equation}
where $m(x,y)$ and $s(x,y)$ are the mean and standard deviation of the intensity across a window surrounding pixel $(x,y)$. The constant $R$ is set to the maximum value of $s(x,y)$ across all pixels in the image, and serves to normalize the influence of the variability of the standard deviation in the image. The value of $k$ and the window size were set to $k=0.05$ and $8$, respectively.

\begin{figure}[h]
\centering
\includegraphics[width=0.7\linewidth]{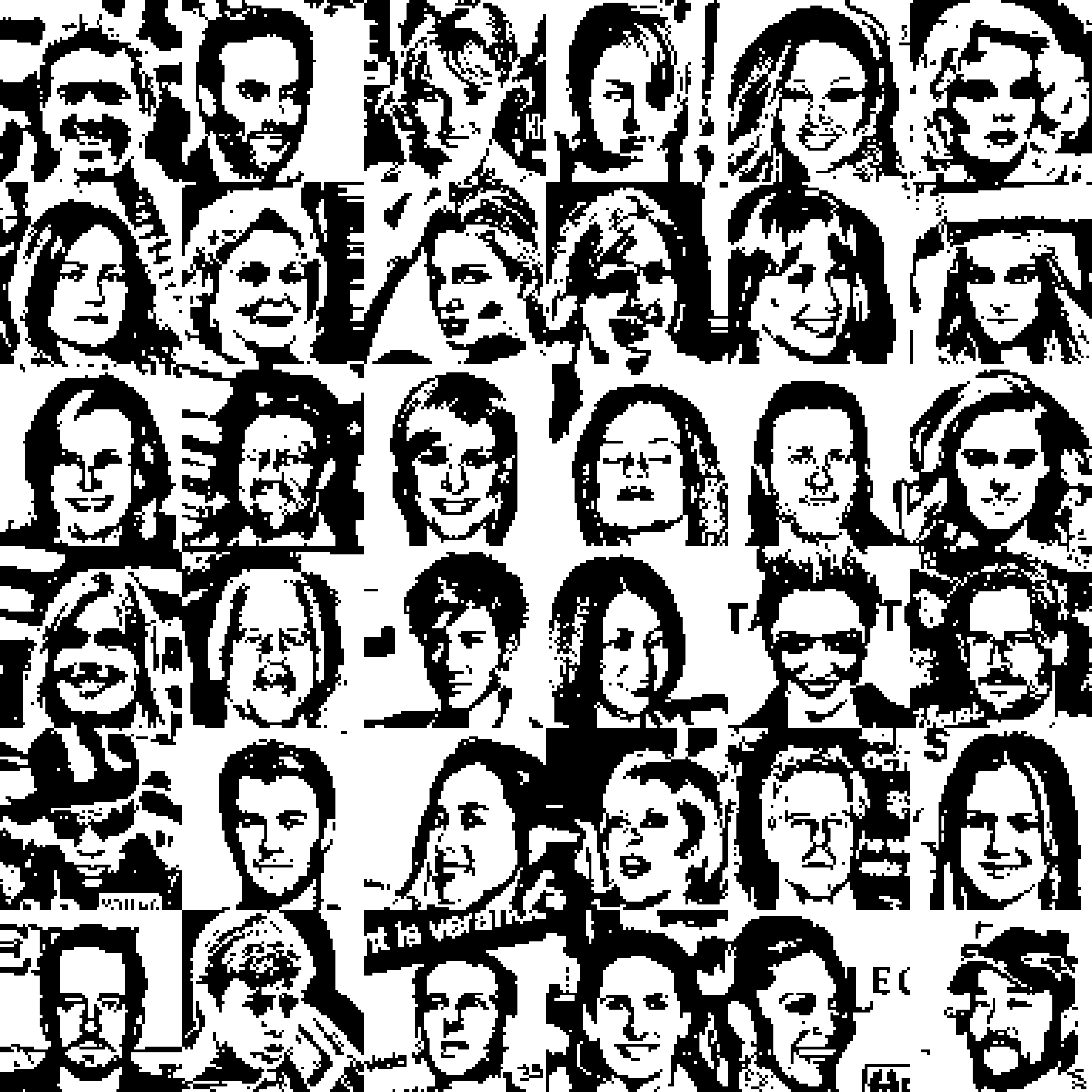}
\caption{\label{SI:fig:celeba_sauvola} Randomly selected images of the CelebA dataset \cite{liu2015faceattributes}, after processing as explained in Sec. \ref{SI:sec:celeba-processing}.}
\end{figure}

Finally, images were resized to a resolution of $64\times64$ to save memory and computation time.
Figure \ref{SI:fig:celeba_sauvola} shows some randomly selected images of the resulting dataset.

\section{Adversarial formulation} \label{SI:sec:adv}

This section discusses the adversarial inspiration for the linear and quadratic constraints (Equations \eqref{eq:Worth} and \eqref{eq:Worth2} in the main-text), and related analytical results.

\subsection{Auxiliary classifier to extract label information from the RBM hidden inputs}

Let $\mathcal{D}=\{(u^b,\mathbf v^b)\}_{b=1}^B$ be a labeled dataset, consisting of $B$ pairs $(u^b,\mathbf v^b)$, where $\mathbf v$ denotes a system configuration and $u$ a label. We denote by $P_\mathcal{D}(u,\mathbf v)$ the empirical distribution:
\begin{equation}
P_\mathcal{D}(u, \mathbf v) = \frac{1}{B} \sum_{b=1}^B \delta(u;u^b)\delta(\mathbf v;\mathbf v^b)
\end{equation}
where $\delta(x;y)=1$ if $x=y$ and $\delta(x;y)=0$ otherwise. Let $P(\mathbf v, \mathbf h)$ be the distribution defined by the RBM (see Equation \eqref{eq:Prbm} in the main text), with latent variables $\mathbf h$. We train the RBM to fit the observations of $\mathbf v$ by maximizing the log-likelihood:
\begin{equation}\label{SI:eq:likelihood}
\mathcal L = \sum_{\mathbf v} P_\mathcal{D}(\mathbf v) \ln P(\mathbf v)
\end{equation}
Let's introduce an auxiliary adversarial classifier model $P_\mathrm{class}(u|\mathbf I)$ that attempts to predict the label $u$ from the RBM inputs $\mathbf I=\mathbf W^\top \mathbf v$. The classifier $P_\mathrm{class}(u|\mathbf I)$ is trained by maximizing the log-likelihood that it makes correct label predictions:
\begin{equation}\label{SI:eq:Ladv}
\mathcal L_\mathrm{class} = \frac{1}{B}\sum_{b=1}^B
\ln P_\mathrm{class}(u^b|\mathbf W^\top\mathbf v^b)
\end{equation}
The performance of the classifier is a measure of the information content of the inputs about the label.

\subsection{Adversarial training}

To reduce the information content about the label in the RBM inputs, we can train the generator and the adversarial classifier together, by solving the following max-min optimization problem:
\begin{equation}\label{SI:eq:minmax}
\max_{\omega_\mathrm{RBM}} \min_{\omega_\mathrm{class}}\left\{
\mathcal L(\omega_\mathrm{RBM}) - \alpha
\mathcal L_\mathrm{class}(\omega_\mathrm{class}; \omega_\mathrm{RBM}) \right\}
=\max_{\omega_\mathrm{RBM}}\left\{
\mathcal L(\omega_\mathrm{RBM}) - \alpha \max_{\omega_\mathrm{class}}
\mathcal L_\mathrm{class}(\omega_\mathrm{class}; \omega_\mathrm{RBM}) \right\}
\end{equation}
where $\alpha\geqslant 0$ is a parameter weighting the relative importance of the two objectives, $\omega_\mathrm{RBM}$ denote the RBM parameters, and $\omega_\mathrm{class}$ denote the classifier parameters. Note that we recover the standard maximum likelihood training of the model, if we set $\alpha = 0$. For $\alpha > 0$, this objective favors RBM parameters for which the best classifier parameters give low performance.
This max-min objective is reminiscent of the training objective of generative adversarial networks \cite{goodfellow2014generative}.

\subsection{Optimal non-parametric classifier}

To gain insight into the meaning of the adversarial penalty term, we carry out the optimization of $\omega_\mathrm{class}$ in the non-parametric limit, and for fixed $\omega_\mathrm{RBM}$. We first define a empirical distribution of inputs
\begin{equation}
P_\mathcal{D}(u,\mathbf I)=\sum_\mathbf{v} P_\mathcal{D}(u,\mathbf v) \delta(\mathbf W^\top\mathbf v; \mathbf I)
\end{equation}
Then, by Gibbs inequality \cite{cover1999elements}:
\begin{equation}\label{SI:eq:gibbsineq}
\begin{aligned}
\mathcal L_\mathrm{class}
&= \sum_{u,\mathbf I} P_\mathcal{D}(u,\mathbf I)\ln P_\mathrm{class}(u|\mathbf I) \\
&\le \sum_{u,\mathbf I} P_\mathcal{D}(u,\mathbf I)\ln\left(\frac{P_\mathcal{D}(u,\mathbf I)}{P_\mathcal{D}(\mathbf{I})}\right) \\
&= \mathrm{MI}(u,\mathbf I) - \mathcal S_\mathrm{label}
\end{aligned}
\end{equation}
where $\mathcal S_\mathrm{label}$ is the entropy of labels in the data. The optimal classifier then satisfies
\begin{equation}\label{SI:eq:bayes_classifier}
P_\mathrm{class}(u|\mathbf I)=\frac{P_\mathcal{D}(u,\mathbf I)}{P_\mathcal{D}(\mathbf{I})}
\end{equation}

In this case, \eqref{SI:eq:minmax} is seen to be penalizing the mutual information between $u$ and $\mathbf I$. In practice, we implement a classifier neural network parameterized by some layer weights and biases $\omega_\mathrm{class}$. If the neural network is powerful enough, it might be able to approximate \eqref{SI:eq:bayes_classifier} closely, but in general this is not the case, and we only attain a lower bound in \eqref{SI:eq:gibbsineq}.

\subsection{Estimation of Mutual Information}

We can use \eqref{SI:eq:gibbsineq} to estimate the mutual information between labels and inputs, as follows. First rewrite \eqref{SI:eq:gibbsineq} as:
\begin{equation}\label{SI:eq:mibound}
\mathrm{MI}(u,\mathbf I) \ge \mathcal L_\mathrm{class} + \mathcal S_\mathrm{label}
\end{equation}
Then we train a set of classifiers, of diverse complexities (hidden layer widths, depth, etc.), and obtain a set of values for $\mathcal L_\mathrm{class}$ on a held-out validation dataset. The bound \eqref{SI:eq:mibound} is tighter for more complex classifiers, as long as they don't overfit. The maximum value obtained for the right-hand side of \eqref{SI:eq:gibbsineq} can be used as an estimate of the mutual information. This procedure is used in Figure \ref{fig:mnist} of the main text.

\subsection{Limits of information erasure from RBM inputs}

In the discussion so far, we have considered arbitrarily complex adversarial classifiers. This section shows a simple counter-example, where the data and the label are such, that the RBM is forced to capture some information about the label, or set its weights to zero ($\mathbf W=0$). This is an undesirable situation because an RBM without weights is a trivial independent-site model.

In this counter-example, the data $\mathbf v$ and the label $u$ are continuous variables. Suppose $\mathbf v$ is a standard multivariate Gaussian random variable in $N$ dimensions, and the label is some function of the magnitude of $\mathbf v$, for instance
\begin{equation}
u(\mathbf v) = \begin{cases}
0, & \text{if $\|\mathbf v\| \le 1$} \\
1, & \text{if $\|\mathbf v\| > 1$}.
\end{cases}
\end{equation}
It is easy to see that $P(v_i|u) \neq P(v_i)$ for any component $v_i$. It follows that $\mathrm{MI}(v_i,u)>0$. By rotation symmetry and scale invariance, $\mathrm{MI}(\mathbf w^\top\mathbf v, u) \neq 0$ for any non-zero vector $\mathbf w$. If the RBM weights are non-zero, it follows that $\mathrm{MI}(\mathbf W^\top\mathbf{v}, u) \geqslant \mathrm{MI}(\mathbf w^\top_\mu \mathbf v, u) > 0$ for any column $\mathbf w_\mu$ of $\mathbf W$. Since rotating the columns of $\mathbf W$ is not sufficient to set the mutual information to zero, it follows that $\mathrm{MI}(\mathbf W\mathbf v, u) = 0$ implies $\mathbf W=0$ in this example.

This example suggests that demanding zero mutual information between labels and inputs can a too strong condition for the RBM.

\subsection{Linear constraint derived from a linear perceptron adversary}

The constraint $\mathrm{MI}(u,\mathbf I)=0$ emerges from \eqref{SI:eq:minmax} by considering arbitrarily complex classifiers. As we have just seen, this constraint might be too strict for the RBM. Weaker conditions can be obtained, by considering \eqref{SI:eq:minmax} under a restricted class of classifiers. In this section, we focus on the linear perceptron classifier.

For a linear perceptron classifier, with binary labels, the likelihood \eqref{SI:eq:Ladv} reads
\begin{equation}
\mathcal L_\mathrm{class} = \left\langle\ln\left(\frac{\mathrm e^{u(\mathbf a^\top\mathbf I + b)}}{1 + \mathrm e^{\mathbf a^\top\mathbf I + b}}\right)\right\rangle_\mathcal{D}
\end{equation}
where $\mathbf a, b$ are the perceptron's weights and bias. The gradient evaluates:
\begin{align}
\frac{\partial \mathcal L_\mathrm{class}}{\partial\mathbf a} &=
\left\langle \left(u - \frac{1}{1 +
\mathrm e^{-\mathbf a^\top I - b}} \right)
\mathbf I \right\rangle_{\mathcal D} \\
\frac{\partial \mathcal L_\mathrm{class}}{\partial b} &= 
\left\langle u - \frac{1}{1 + \mathrm e^{-\mathbf a^\top \mathbf I - b}}
\right\rangle_{\mathcal D}
\end{align}
If the perceptron is unable to extract any information about the label from the inputs, we must have $\mathcal L_\mathcal{D} \le -\mathcal S_\mathrm{label}$, \emph{i.e.}, the classifier is not doing better than randomly guessing the labels. Since the same base performance is achievable with $\mathbf a = 0$, $b=\ln(\langle u\rangle_\mathcal{D}/(1 - \langle u\rangle_\mathcal{D}))$, it follows that these values must set the gradient to zero. We then obtain the condition
\begin{equation}
\langle u\mathbf I\rangle_\mathcal{D} -\langle u\rangle_\mathcal{D}\langle\mathbf I\rangle_\mathcal{D} = 0
\end{equation}
equivalent to \eqref{eq:Worth} in the main text. This argument shows sufficiency of this condition. This condition is also necessary, as follows from the concavity of $\mathcal L_\mathrm{class}$ in $\mathbf a,b$.

\subsection{Generalization to arbitrary kernel functions}

Now we consider a kernel perceptron,
\begin{equation}
\mathcal L_\mathrm{class} = \left\langle\ln\left(\frac{\mathrm e^{u(\mathbf a^\top \boldsymbol{\phi}(\mathbf I) + b)}}{1 + \mathrm e^{\mathbf a^T\boldsymbol{\phi}(\mathbf I) + b}}\right)\right\rangle_\mathcal{D}
\end{equation}
where $\boldsymbol{\phi}(\mathbf I)$ is a function (the `kernel') that extracts a vector of features from the inputs, $\mathbf{a}$ are weights assigned by the perceptron to these features, and $b$ a bias scalar. Taking the gradient in $\mathbf a, b$,
\begin{align}
\frac{\partial \mathcal L_\mathrm{class}}{\partial\mathbf a} 
&=
\left\langle \left(u - \frac{1}{1 +
\mathrm e^{-\mathbf a^\top \boldsymbol\phi(\mathbf I) - b}} \right)
\boldsymbol\phi(\mathbf I) \right\rangle_{\mathcal D}
\\
\frac{\partial\mathcal L_\mathrm{class}}{\partial b}
&= 
\left\langle u - \frac{1}{1 + \mathrm e^{-\mathbf a^\top \boldsymbol\phi(\mathbf I) - b}}
\right\rangle_{\mathcal D}
\end{align}
By an analogue argument, we find that if the perceptron is unable to extract information about the label from the inputs, then $\mathbf a=0$, $b=\ln(\langle u\rangle_\mathcal{D}/(1 - \langle u\rangle_\mathcal{D}))$, must set this gradient to zero. Therefore, we obtain the conditions:
\begin{equation}
\langle u \boldsymbol\phi(\mathbf I)\rangle_\mathcal{D} - \langle u \rangle_\mathcal{D}\langle \boldsymbol\phi(\mathbf I)\rangle_\mathcal{D} = 0
\end{equation}
which generalize the above linear conditions to arbitrary kernel functions.

\subsection{Quadratic constraint and the quadratic kernel perceptron adversary}

If we consider a set of quadratic features $\phi_{\mu\nu}(\mathbf I) = I_\mu I_\nu$, for $\mu<\nu$, we obtain
\begin{equation}
\langle u I_\mu I_\nu\rangle_\mathcal{D} - \langle u \rangle_\mathcal{D} \langle I_\mu I_\nu\rangle_\mathcal{D} = 0
\end{equation}
which is equivalent to \eqref{eq:constraint2nd_h} in the main text. If the RBM weights satisfy this constraint, it follows that a perceptron with a quadratic kernel cannot do better than random guessing of the labels, as stated in the main text.

\subsection{Generalization to multi-categorical labels}

So far we have considered binary labels. The generalization to multi-categorical labels is straightforward. We consider a one-hot encoded label with $D$ possible classes. The label $\mathbf u^b$ is now a $D$-dimensional vector, with components $u_d^b=1$ if the $b$'th data point belongs to class $d$, and $u_d^b=0$ otherwise. Therefore, $\mathbf u^b$ is a vector with binary components, where one component equals 1, and all the other components equal 0. We consider now the general kernel perceptron, trained to predict these labels from the RBM inputs. Its likelihood reads:
\begin{equation}
\mathcal L_\mathrm{class}=
\left\langle \ln\left(\frac{\mathrm e^{\mathbf u^\top(\mathbf A^\top \boldsymbol\phi(\mathbf I) + \mathbf b)}}{\sum_{d=1}^D\mathrm e^{(\mathbf A^\top \boldsymbol\phi(\mathbf I) + \mathbf b)_d}}\right)\right\rangle_\mathcal{D}
\end{equation}
Due to the multiple class values, now $\mathbf A$ is a matrix and $\mathbf b$ a vector. As before, we look for the condition that $\mathbf A=0$ gives a stationary point of the gradient of $\mathcal L_\mathrm{class}$. We then obtain the following condition:
\begin{equation}
\langle u_d \boldsymbol\phi(\mathbf I)\rangle_\mathcal{D} - \langle u_d \rangle_\mathcal{D}\langle \boldsymbol\phi(\mathbf I)\rangle_\mathcal{D} = 0
\end{equation}
In the linear case, $\mathbf\phi(\mathbf I)=\mathbf I$, this condition yields the orthogonality constraint of the RBM weights to the vectors $\mathbf q_d^{(1)}$, defined in \eqref{eq:WorthMulti} (main text).

\subsection{Case of Gaussian data with linear labels}

In the previous sections, we have seen how the RBM inputs might be unable to erase completely label information in some cases, and then considered weakened linear and quadratic constraints. In this section we prove that, at least in a simplified case, these constraints can be sufficient to completely erase the label information.

\paragraph*{Theorem.} For Gaussian distributed data, and binary labels assigned by a linear perceptron, constraint \eqref{eq:Worth} in the main text is sufficient to erase the label from the RBM inputs.

\paragraph*{Proof.}
To be precise, suppose the data follows a multivariate Gaussian distribution,
\begin{equation}
P_\mathcal{D}(\mathbf v) = \frac{(2\pi)^{- \frac{N}{2}}}{\sqrt{\det(\mathbf C)}} \exp\left(-\frac{1}{2}\mathbf v^\top \mathbf C^{-1}\mathbf v\right)
\end{equation}
where $\mathbf C$ is the covariance matrix. We consider zero means for simplicity, since non-zero means can be treated by simply translating the origin of coordinates, without losing generality. Suppose the label is assigned by a linear perceptron with weights $\mathbf r$ and bias $c$.
\begin{equation}
P_\mathrm{label}(u|\mathbf v) =
\frac{\mathrm e^{u(\mathbf r^\top\mathbf v + c)}}{1 + \mathrm e^{\mathbf r^\top\mathbf v + c}}
\end{equation}
Therefore the label $u$ depends on the data only through the dot product $\mathbf r^\top\mathbf v$.

We consider the joint multivariate distribution of the variables $\mathbf r^\top\mathbf v$, $\mathbf I=\mathbf W^\top\mathbf v$. Since the data is Gaussian, this joint distribution is also Gaussian, with covariance matrix 
\begin{equation}
\left\langle \left(\begin{array}{c}
  \mathbf{W}^{\top} \mathbf{v}\\
  \mathbf{r}^{\top} \mathbf{v}
\end{array}\right) \left(\begin{array}{cc}
  \mathbf{v}^{\top} \mathbf{W} & \mathbf{v}^{\top} \mathbf{r}
\end{array}\right) \right\rangle = \left(\begin{array}{cc}
  \mathbf{W}^{\top} \mathbf{C}\mathbf{W} & \mathbf{W}^{\top}
  \mathbf{C}\mathbf{r}\\
  \mathbf{r}^{\top} \mathbf{C}\mathbf{W} & \mathbf{r}^{\top}
  \mathbf{C}\mathbf{r}
\end{array}\right)
\end{equation}
In particular, we can compute the mutual information analytically:
\begin{equation}
\mathrm{MI}(\mathbf r^\top\mathbf v,\mathbf I) = -\frac{1}{2}\ln(1-\rho^2)
\end{equation}
where
\begin{equation}
\rho^2 = \frac{\mathbf r^\top \mathbf C\mathbf W (\mathbf W^\top
\mathbf C\mathbf W)^{-1}\mathbf W^\top
\mathbf C\mathbf r}{\mathbf r^\top\mathbf C\mathbf r} 
\end{equation}
We have that $\mathrm{MI}(\mathbf r^\top\mathbf v, \mathbf W^\top\mathbf v) = 0$ if and only if $\mathbf W^\top\mathbf C\mathbf r=0$; that is, the patterns have to be orthogonal to $\mathbf C\mathbf r$. Now, we show that the vector $\mathbf q^{(1)}$, that we defined in \eqref{eq:q1} in the main text, is proportional to $\mathbf C\mathbf r$. Indeed,
\begin{equation}
\mathbf q^{(1)}
=\langle u\mathbf v\rangle_\mathcal{D}
=\left\langle \frac{1}{1+\mathrm e^{-\mathbf r^\top\mathbf v - c}}\mathbf v\right\rangle_\mathcal{D}
\end{equation}
Now suppose we multiply $\mathbf q^{(1)}$ by an arbitrary vector $\mathbf n$,
\begin{equation}
\mathbf n^\top \mathbf q^{(1)}
=\left\langle \frac{\mathbf n^\top\mathbf v}{1+\mathrm e^{-\mathbf r^\top\mathbf v - c}}\right\rangle_\mathcal{D}
\end{equation}
The variables $\mathbf r^\top\mathbf v$ and $\mathbf n^\top\mathbf v$, are jointly Gaussian, with zero means, and covariance matrix:
\begin{equation}
\left\langle \left(\begin{array}{c}
  \mathbf{n}^{\top} \mathbf{v}\\
  \mathbf{r}^{\top} \mathbf{v}
\end{array}\right) \left(\begin{array}{cc}
  \mathbf{v}^{\top} \mathbf{n} & \mathbf{v}^{\top} \mathbf{r}
\end{array}\right) \right\rangle = \left(\begin{array}{cc}
  \mathbf{n}^{\top} \mathbf{C}\mathbf{n} & \mathbf{n}^{\top}
  \mathbf{C}\mathbf{r}\\
  \mathbf{r}^{\top} \mathbf{C}\mathbf{n} & \mathbf{r}^{\top}
  \mathbf{C}\mathbf{r}
\end{array}\right)
\end{equation}
If $\mathbf n^\top\mathbf C\mathbf r=0$, then $\mathbf r^\top\mathbf v$ and $\mathbf n^\top\mathbf v$ are independent. In this case, $\mathbf n^\top \mathbf q^{(1)}=0$. Therefore, any vector $\mathbf n$ that is orthogonal to $\mathbf C\mathbf r$, is also orthogonal to $\mathbf q^{(1)}$. We conclude that $\mathbf C\mathbf r$ and $\mathbf q^{(1)}$ have the same direction. We have thus shown that $\mathrm{MI}(\mathbf r^\top\mathbf v,\mathbf I)=0$ if $\mathbf W^\top\mathbf q^{(1)}=0$, proving the theorem.

\section{Gaussian-Spin model} \label{SI:sec:GaussianSpin}

We consider an RBM with $M$ hidden units, the first of which is a spin unit taking values $h_1=\pm1$, while the remaining $M$ are Gaussian, taking real values $h_\mu\in\mathbb{R}$, for $2\le\mu\le M$. All $N$ visible units are also Gaussian. The energy function writes:
\begin{equation}
E_\mathrm{GS}(\mathbf v,\mathbf h) = \sum_{i=1}^N \frac{v_i^2}{2\sigma_i^2} - \sum_{i=1}^N g_i v_i - \theta h_1 + \sum_{\mu=2}^M\frac{h_\mu^2}{2} - \sum_{i=1}^N w_i^\ast v_i h_1 - \sum_{i=1}^N\sum_{\mu=2}^M w_{i\mu}v_i h_\mu
\end{equation}
where $\sigma_i$ are the empirical standard deviations of the visible units (that we estimate directly from the data), $w_{i\mu}$, $w_i^\ast=w_{i,1}$ the weights, $g_i$ the visible fields, and $\theta$ the bias field for $h_1$. The partition function of the model can be evaluated,
\begin{align}
Z_\mathrm{GS} &= 
\sum_{h_1=\pm1}\int \mathrm{e}^{-E_\mathrm{GS} (\mathbf{v},\mathbf{h})}
\mathrm{d}v_1\ldots\mathrm{d}v_N \mathrm{d}h_2\ldots\mathrm{d}h_M \\
&= 2 (2\pi)^{\frac{N+M}{2}} \sqrt{\det(\Sigma)} \mathrm e^{\frac{1}{2}
(\mathbf g^\top \Sigma \mathbf g+\mathbf w_1^\top\Sigma
\mathbf w_1)} \cosh(\theta h_1 + \mathbf g^\top \Sigma\mathbf w_1)
\end{align}
where
\begin{equation}
\Sigma^{-1}=\mathbf{D}-\mathbf{W}\mathbf{W}^\top
\end{equation}
$\mathbf D$ is a diagonal matrix with entries $1/\sigma_i^2$, and $\mathbf W$ is the matrix with entries $w_{i\mu}$ for $\mu>1$.
The distribution defined by the model can be written $P_\mathrm{GS}(\mathbf v,h_1)=P_\mathrm{GS}(h_1)P_\mathrm{GS}(\mathbf v|h_1)$, where
\begin{equation}
P_\mathrm{GS}(h_1)=\frac{\mathrm{e}^{(\theta + \mathbf g^\top\Sigma\mathbf w_1) h_1}}{2 \cosh(\mathbf g^\top\Sigma \mathbf w_1)},
\quad
P_\mathrm{GS}(\mathbf v|h_1)=\frac{\mathrm e^{-\frac{1}{2}(\mathbf v - \Sigma\mathbf g- h_1\Sigma\mathbf w_1)^\top \Sigma^{-1} (\mathbf v- \Sigma \mathbf g - h_1\Sigma\mathbf w_1)}}{\sqrt{\det(2\pi\Sigma)}}
\end{equation}
The later is a multivariate normal, with mean $\langle\mathbf v|h_1\rangle = \Sigma\mathbf g+h_1\Sigma\mathbf w_1$, and covariance matrix $\Sigma$.

We consider data $\mathbf v^1, \dots, \mathbf v^B$. For simplicity, we assume that the classes are well separated and balanced. The machine encodes the class label in the spin variable $h_1$, and we assume its value is known. In this setting, the average likelihood reads:
\begin{equation}
\mathcal L_\mathrm{GS} = \frac{1}{B}\sum_{n=1}^B \ln P_\mathrm{GS}(\mathbf v^n, h_1^n)
\end{equation}
Ignoring constant terms,
\begin{equation}
\mathcal L_\mathrm{GS} = -\frac{1}{2} \Tr(\Sigma^{-1} \langle (\mathbf v - \Sigma \mathbf g - h_1 \Sigma \mathbf w_1) (\mathbf{v} - \Sigma \mathbf g - h_1 \Sigma \mathbf w_1)^\top \rangle_\mathcal{D}) - \ln\cosh (\theta + \mathbf g^\top \Sigma \mathbf w_1) - \frac{1}{2} \ln\det(\Sigma)
\end{equation}
where we used $\langle h_1\rangle_\mathcal{D}=0$, since the classes are balanced. Training the RBM amounts to maximizing $\mathcal L_\mathrm{GS}$ in the parameters. Taking the gradient, we obtain the moment-matching conditions:
\begin{align}
\frac{\partial\mathcal L_\mathrm{GS}}{\partial\theta} &= - \langle h_1 \rangle=0 \\
\frac{\partial\mathcal L_\mathrm{GS}}{\partial g_i} &= \langle v_i \rangle_\mathcal{D} - \langle v_i \rangle = 0 \\
\frac{\partial\mathcal L_\mathrm{GS}}{\partial w_i^\ast} &= \langle v_i h_1\rangle_\mathcal{D} - \langle v_i h_1 \rangle=0 \\
\frac{\partial\mathcal L_\mathrm{GS}}{\partial w_{i\mu}} &= \langle v_i h_\mu \rangle_\mathcal{D} - \langle v_i h_\mu \rangle=0\quad (\mu > 1)
\end{align}
After some algebra, the second equation rewrites:
\begin{equation}
(\mathbf D-\mathbf W\mathbf W^\top)^{-1}\mathbf W=(\mathbf C - \langle h_1\mathbf v\rangle_{\mathcal D}\langle h_1\mathbf v\rangle_{\mathcal D}^\top) \mathbf W
\end{equation}
where $\mathbf C$ is the empirical covariance matrix of $\mathbf v$:
\begin{equation}
\mathbf C = \langle\mathbf v\mathbf v^\top\rangle_{\mathcal D} - \langle\mathbf v\rangle_{\mathcal D}\langle\mathbf v\rangle_{\mathcal D}^\top
\end{equation}
Note that $\mathbf C - \langle h_1\mathbf v\rangle_{\mathcal D}\langle h_1\mathbf v\rangle_{\mathcal D}^\top$ amounts to the covariance matrix of the data, if the two classes are collapsed, by bringing their centers of mass together. To solve this equation, the scaled weights $w_{i\mu}\sigma_i$, $\mu>1$, must be eigenvectors of the scaled matrix 
\begin{equation}
\tilde{\mathbf{C}} = \mathbf D(\mathbf C - \langle h_1\mathbf v\rangle_{\mathcal D}\langle h_1\mathbf v\rangle_{\mathcal D}^\top)\mathbf D,
\end{equation}
with eigenvalues $\lambda_\mu = (1 - \sum_i w_{i\mu}^2/\sigma_i^2)^{-1}$. The likelihood after training then evaluates:
\begin{equation}\label{SI:eq:likelihoodGS}
\mathcal L_\mathrm{GS}=\frac{1}{2} \sum_{\mu}(\lambda_{\mu} - 1 - \log \lambda_{\mu}) - \log\cosh\left(\mathbf g^\top\mathbf{q}^{(1)}\right)
\end{equation}
where the $\lambda_\mu$'s are the selected eigenvalues of $\tilde{\mathbf{C}}$, and we have ignored irrelevant additive terms.

\section{Classifier architectures and training} \label{SI:sec:classifiers}

We considered a number of classifier architectures, that we fit to the inputs of the RBM. Let $M$ be the dimensionality of an input data point: $N=28^2=782$ for MNIST images, $N=32^2$ or $N=64^2$ for the Ising model (where we considered grids of length 32 and 64), and $N=21L$ for a one-hot encoded protein of length $L$. While the input size differs for each dataset, the output size is 2 for all, since we only considered binary labels. The classifiers considered are:
\begin{itemize}
\item A perceptron classifier, with no hidden layer.
\item 11 classifiers with one hidden layer of widths $2^n$ with $n=0, 1, \dots, 10$.
\item 7 classifiers, with a first hidden layer of width 128, and a second hidden layer with widths $2^n$ with $n=0, 1, \dots, 6$.
\item 8 classifiers, with a first hidden layer of width 256, and a second hidden layer with widths $2^n$ with $n=0, 1, \dots, 7$.
\item 9 classifiers, with a first hidden layer of width 512, and a second hidden layer with widths $2^n$ with $n=0, 1, \dots, 8$.
\end{itemize}
for a total of 36 classifiers.

All classifiers are trained for 50000 parameter update steps, with batchsize equal to 128, with the ADAM optimizer \cite{kingma2013auto} and with a learning rate of $10^{-3}$. In each case we verified convergence by ensuring that prediction accuracy and the cross-entropy evaluated on the training data reached saturating values. The hidden units have a SELU nonlinearity \cite{klambauer2017self}.

\section{Details on the TM-score} \label{SI:TM}

To compare the inferred structures of the sampled sequences from the different RBM models, to those of natural sequences, we used the standard Template Modelling (TM) score \cite{zhang2004scoring}. Compared to other similarity scores (like root-mean-square deviation (RMSD)) it gives a more accurate measure since it relies more on the global similarity of the full sequence rather than on local similarities. Practically, we consider a target sequence of length $L$ and a template one whose structure has to be compared with. First, we align the two sequences and we take the $L$ common pairs of residues that commonly appear aligned. Then the score
is computed as
\begin{equation}
    \text{TM-score} = \max_{\{d_i\}} \left[
        \frac{1}{L_\mathrm{target}} \sum_{i=1}^{L_\mathrm{common}} \frac{1}{1 + \frac{d_i^2}{d_0^2(L_\mathrm{target}}}
    \right]
\end{equation}
where $d_i$ is the distance between the $i$'th pair of residues between the template and the target structures (after alignment), and $d_0(L_\mathrm{target}) = 1.24(L_\mathrm{target} - 15)^{1/3} - 1.8$ is a normalized distance scale. This formula gives a score between 0 and 1. If $\text{TM-score}<0.2$, the two sequences are structurally uncorrelated, while they can be considered to have significantly similar structures if $\text{TM-score}>0.5$.

\section{Quadratic constraint for the Ising model} \label{SI:sec:Q2}

The general second-order constraint derived in the main text, utilizes matrix $\mathbf{q^{(2)}}$ related to second-order correlations between the label and the data,
\begin{equation}
q^{(2)}_{ij} = \langle u v_i v_j\rangle_\mathrm{d} - \langle u \rangle_\mathrm{d} \langle v_i v_j \rangle_\mathrm{d}
\end{equation}
For the Ising model however, where we use the label $u=0$ for configurations with negative magnetization, and $u=1$ for positive magnetization, we can verify that $q^{(2)}_{ij}=0$ identically, as a consequence of the invariance of the Ising model energy to flipping signs of all spins.
Therefore the constraint in its original form is trivial. However, a meaningful constraint can be obtained by considering the addition of a small external field to spins. Consider the energy:
\begin{equation}
E(\mathbf{v}) - h \sum_l v_l = \frac{1}{N} \sum_{(ij)} v_i v_j - h \sum_l v_l
\end{equation}
where $(ij)$ refers to pairs of connected sites on the rectangular grid, and $h$ is a small external field.
We want to compute
\begin{equation}
Q_{ij}(h) = \langle u s_i s_j \rangle_h - \langle u \rangle_h \langle s_i s_j \rangle_h
\end{equation}
for small $h$, where
\begin{equation}
u = u(\mathbf{s}) = \begin{cases}
 1 & \sum_i s_i > 0 \\
 0 & \sum_i s_i \le 0
\end{cases}
\end{equation}

Consider any function of the spins,
\begin{align}
\langle f(\mathbf{s})\rangle_h 
&= \frac{\sum_\mathbf{s} \exp\left(-E(\mathbf{s}) + h\sum_l s_l\right) f(\mathbf{s})}{\sum_\mathbf{s} \exp\left(-E(\mathbf{s}) + h\sum_l s_l\right)} \approx  \frac{\sum_\mathbf{s} \mathrm{e}^{-E(\mathbf{s})} \left(1 + N h\sum_l s_l\right) f(\mathbf{s})}{\sum_\mathbf{s} \mathrm{e}^{-E(\mathbf{s})} \left(1 + N h\sum_l s_l\right)} \\
&\approx \frac{\sum_\mathbf{s} \mathrm{e}^{-E(\mathbf{s})} f(\mathbf{s})}{\sum_{\mathbf{s}} \mathrm{e}^{- E (\mathbf{s})}} + h \frac{\sum_l
\sum_{\mathbf{s}} \mathrm{e}^{- E (\mathbf{s})} s_l f
(\mathbf{s})}{\sum_{\mathbf{s}} \mathrm{e}^{- E (\mathbf{s})}} - h \frac{\left(
\sum_{\mathbf{s}} \mathrm{e}^{- E (\mathbf{s})} f (\mathbf{s}) \right) \left(
\sum_l \sum_{\mathbf{s}} \mathrm{e}^{- E (\mathbf{s})} s_l \right)}{\left(
\sum_{\mathbf{s}} \mathrm{e}^{- E (\mathbf{s})} \right)^2} \\
&= \langle f (\mathbf{s}) \rangle_0 - h \langle f (\mathbf{s}) \rangle_0
\left\langle \sum_l s_l \right\rangle_0 + h \left\langle \sum_l s_l f
(\mathbf{s}) \right\rangle_0 = \langle f (\mathbf{s}) \rangle_0 + h \left\langle \sum_l s_l f (\mathbf{s}) \right\rangle_0
\end{align}
where in the last step we used $\langle s_l\rangle_0 = 0$ due to sign reversal symmetry.
Applying this repeatedly:
\begin{equation}
\langle u \rangle_h = \langle u \rangle_0 + h \left\langle u \sum_l s_l
\right\rangle_0 = \frac{1}{2} + \frac{h}{2} \left\langle \left| \sum_l s_l
\right| \right\rangle_0
\end{equation}

\begin{equation}
\langle s_i s_j \rangle_h = \langle s_i s_j \rangle_0 + h \left\langle \sum_l s_l s_i s_j \right\rangle_0 = \langle s_i s_j \rangle_0
\end{equation}

\begin{equation}
\langle u s_i s_j \rangle_h = \langle u s_i s_j \rangle_0 + h \left\langle u \sum_l s_l s_i s_j \right\rangle_0 = \frac{\langle s_i s_j \rangle_0}{2} + \frac{h}{2} \left\langle \left| \sum_l s_l \right| s_i s_j \right\rangle_0
\end{equation}
Finally, substituting in $Q$,
\begin{align}
Q_{ij}(h) &= \langle u s_i s_j \rangle_h - \langle u \rangle_h \langle s_i s_j \rangle_h \\
&\approx \frac{h}{2} \left( \left\langle \left| \sum_l s_l \right| s_i s_j \right\rangle_0 - \left\langle \left| \sum_l s_l \right| \right\rangle_0 \langle s_i s_j \rangle_0 \right)
\end{align}
to first-order in $h$. Therefore
\begin{align}
Q_{ij}^{(2)} &= \lim_{h\rightarrow 0^+} \frac{Q_{ij}(h)}{h} = \frac{1}{2} \left( \left\langle \left| \sum_l s_l \right| s_i s_j \right\rangle_0 - \left\langle \left| \sum_l s_l \right| \right\rangle_0 \langle s_i s_j \rangle_0 \right)\propto
\langle |m| s_i s_j \rangle - \langle|m|\rangle \langle s_i s_j \rangle
\end{align}
is the matrix we use for the second-order constraint in the Ising model. To speed up the calculation, we compute this quantity using two-dimensional fast Fourier transform across the lattice.

\section{Estimating the overlap for the semi-supervised learning with subsampled labeled dataset} \label{SI:subsample-bound}

To derive the overlap estimate in \eqref{eq:overlap-bound} of the main-text, we consider two classes following multivariate normal distributions, $\mathcal{N}(\mathbf v^{(u)}, C^{(u)})$, for $u=0,1$, where $\mathbf v^{(u)}, C^{(u)}$ are the mean and covariance matrix, respectively. We then draw $B$ samples from each of these distributions, $\mathbf v_1^{(u)}, \dots, \mathbf v_B^{(u)}$ and construct the estimate:
\begin{equation}
 \mathbf{q}_{sub} = \frac{1}{B} \sum_{b=1}^B (\mathbf v^{(0)}_b - \mathbf v^{(1)}_b)
\end{equation}
Note that $\mathbf q_{sub}$ is also Gaussianly distributed, with mean $\mathbf{q}=\mathbf v^{(0)} - \mathbf v^{(1)}$ and covariance $(C^{(0)} + C^{(1)}) / B$. Consequently,  the squared norm of $\mathbf q_{sub}$ is on average equal to
\begin{equation}
 |\mathbf q_{sub} |^2 = \frac{1}{B}\Tr(C^{(0)}+C^{(1)}) + |\mathbf {q}|^2 \ .
\end{equation}
We may now compute the average value of the overlap $\phi$ between $\mathbf{q}$ and $\mathbf{q}_{sub}$, see definition in Eq.~\eqref{eq:overlap}, 
\begin{align}
  \langle\phi\rangle 
  &\approx \frac{\mathbf q \cdot \langle\mathbf q_{sub}\rangle}{|\mathbf q|\; |\mathbf q_{sub}|} 
  = \frac{|\mathbf q|^2}{|\mathbf q| \sqrt{\frac{1}{B} \Tr(C^{(0)}+C^{(1)}) + |\mathbf q|^2}}
  = \frac{1}{\sqrt{1 + \frac{1}{B} \frac{\Tr(C^{(0)}+C^{(1)})}{|\mathbf q|^2}}}
\end{align}
This expression coincides with Eq.~\eqref{eq:overlap-bound}  in the main text.

\clearpage

\section{Supplementary figures} \label{sec:SI:suppl-fig}

The following figures contain additional results for the datasets (CelebA, MNIST, Ising model, and KH protein domain) that are referred to in the main text.

\vskip 2cm

\begin{figure}
\centering
\includegraphics[width=\linewidth]{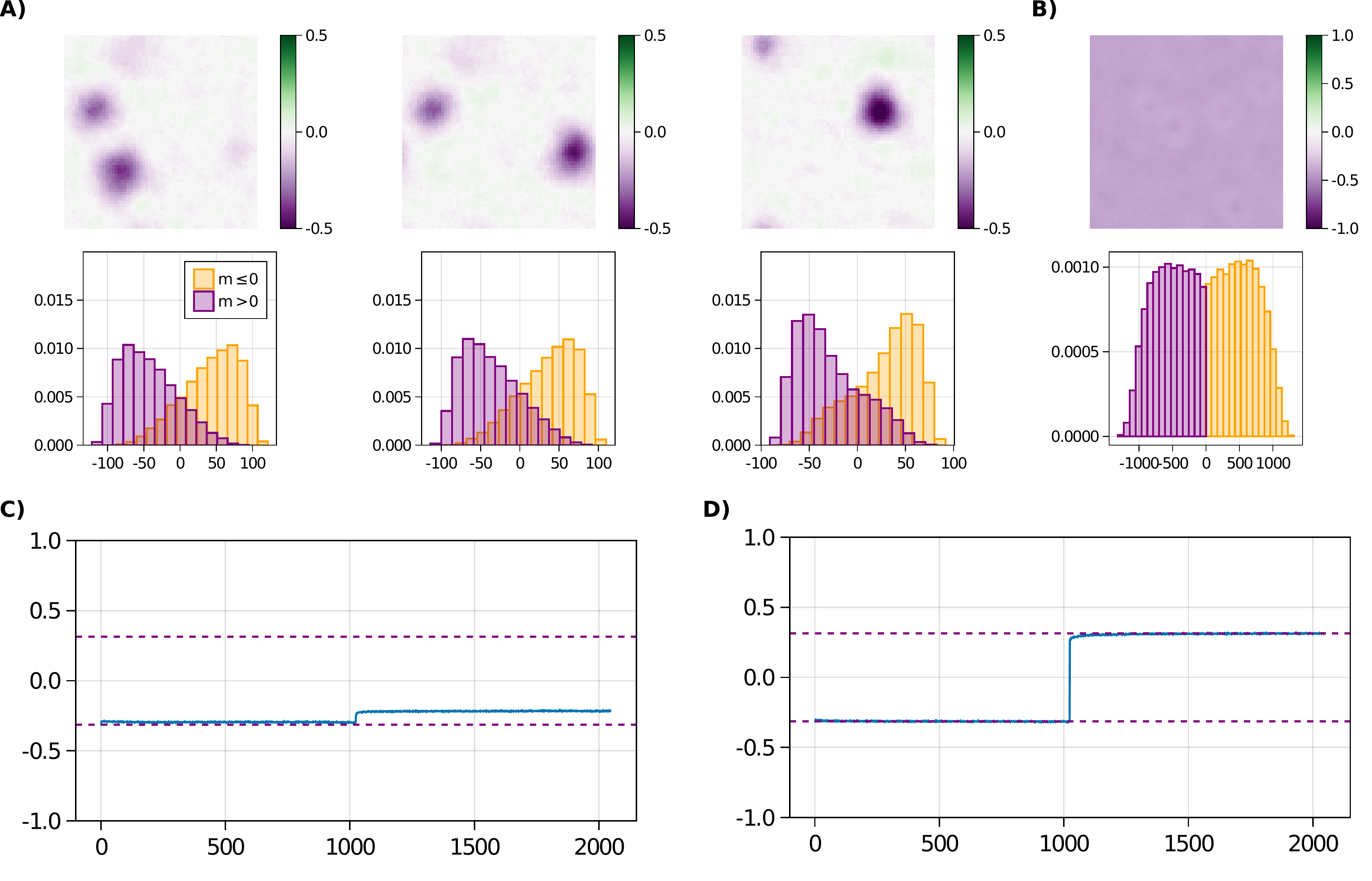}
\caption{\label{SI:fig:ising_entangled}\textbf{A)} Hidden unit weights learned by a normal RBM trained on Ising data ($\beta=0.443$, $L=64$). We select the 3 hidden units for which the correlation between inputs and magnetization is highest. Bottom panel shows the input histograms, colored according to the sign of the magnetization for each configuration. \textbf{B)} Same as A), but for the released hidden unit in an RBM trained with constraint \eqref{eq:Worth} (main text) imposed on all but one hidden unit. \textbf{C)} Manipulating the top correlated hidden unit in a normal RBM fails to flip the sign of the magnetization of the sampled configurations. \textbf{D)} Manipulating the released hidden unit (shown in B) succeeds in flipping the magnetization sign of sampled configurations.}
\end{figure}

\begin{figure}
\centering
\includegraphics[width=\linewidth]{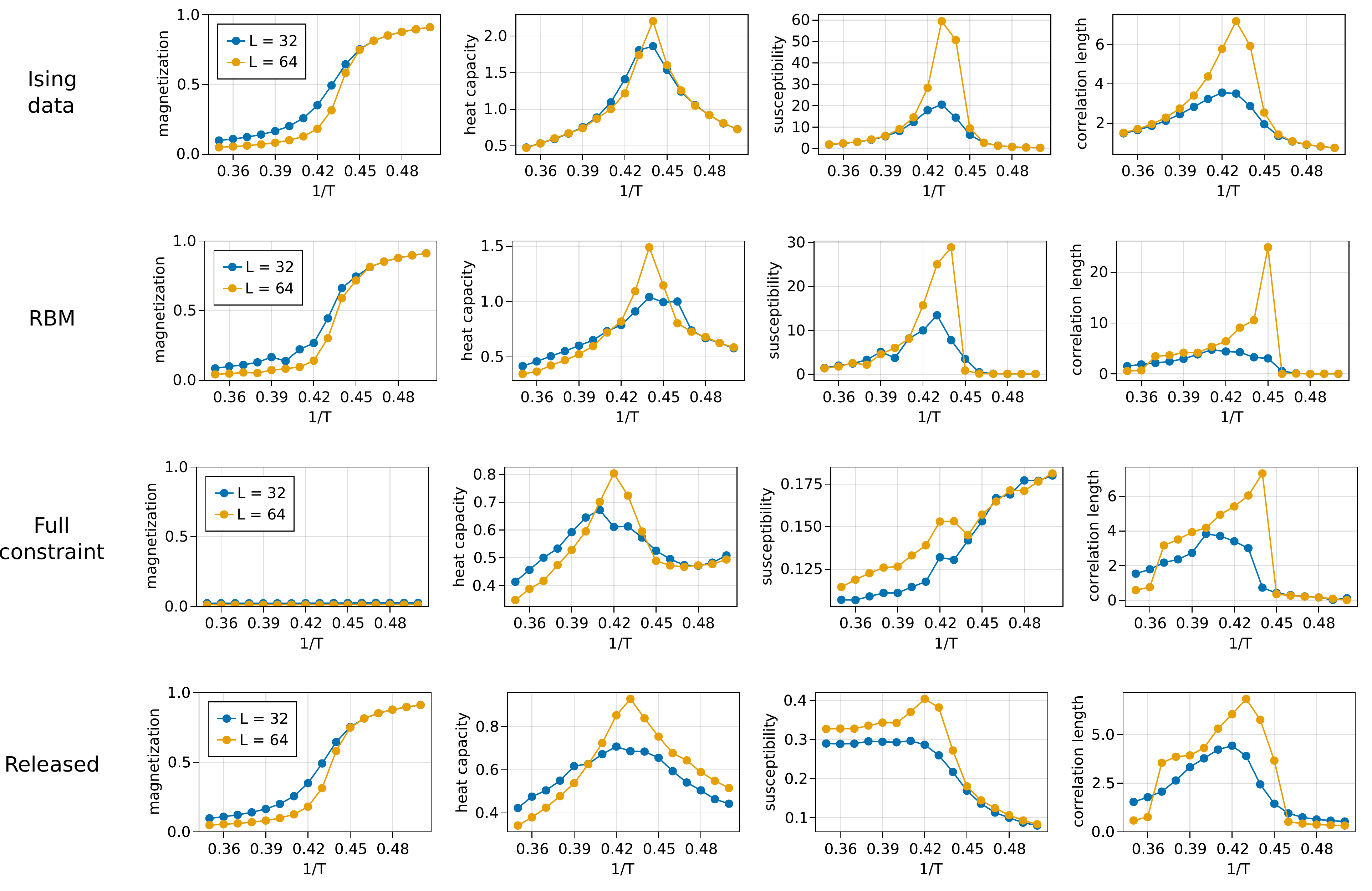}
\caption{\label{SI:fig:ising_obs}Observables for 2D-Ising model as a function of temperature. All RBMs with $M=50$ hidden units. Similar results were obtained with $M=10,100$.}
\end{figure}

\begin{figure}
\centering
\includegraphics[width=0.8\linewidth]{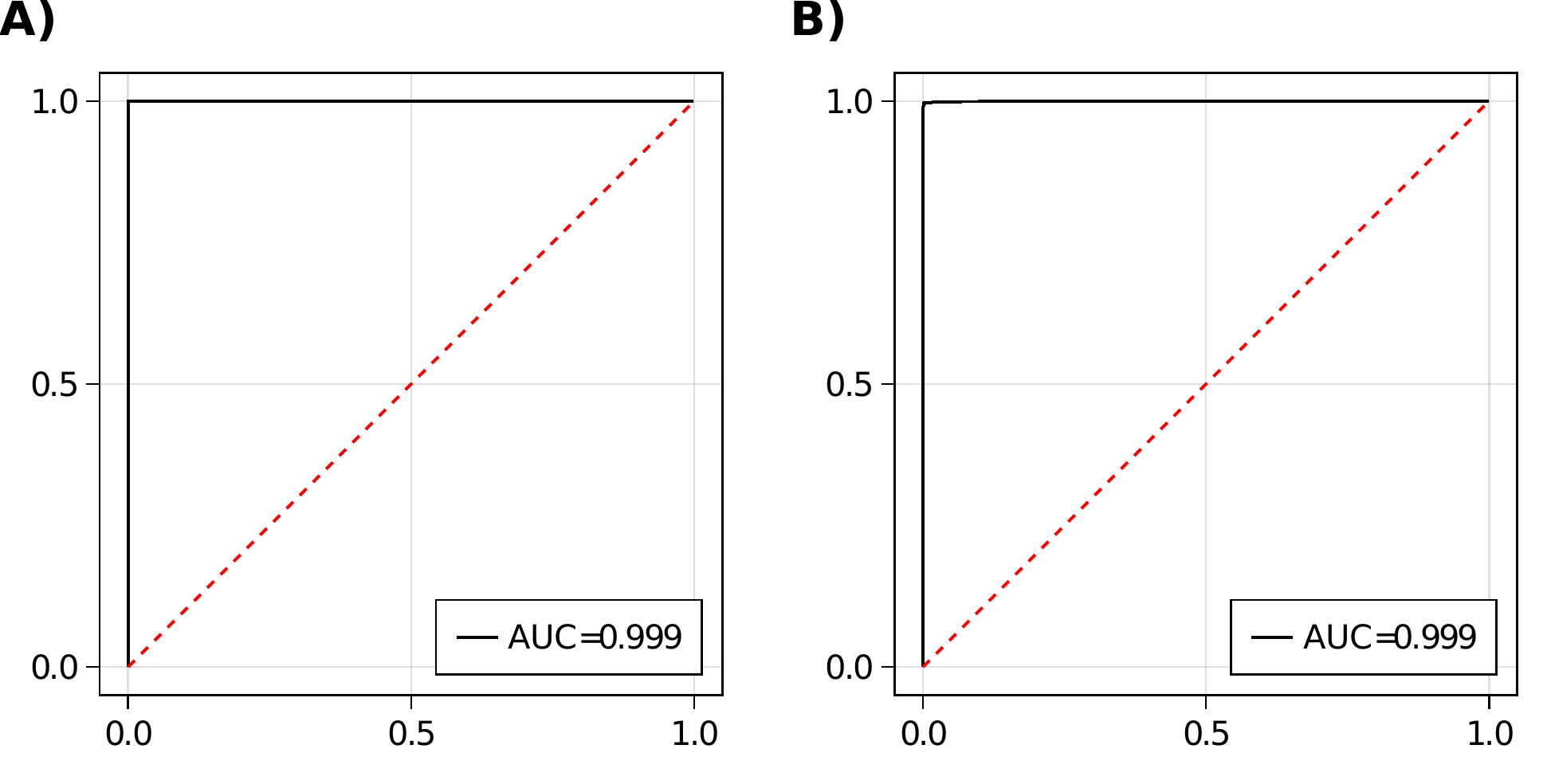}
\caption{\label{SI:fig:inputs_classifier}\textbf{A)} Receiver operating characteristic (ROC) curve of linear perceptron classifier trained on RBM inputs. The RBM was trained on MNIST0/1 data, and the classifier objective is to predict the digit class of images presented on the RBM visible layer. \textbf{B)} Like A), but for sequences from the KH domain, labeled by their taxonomic origin (bacteria or eukaryotic).}
\end{figure}

\begin{figure}
\centering
\includegraphics[width=0.8\linewidth]{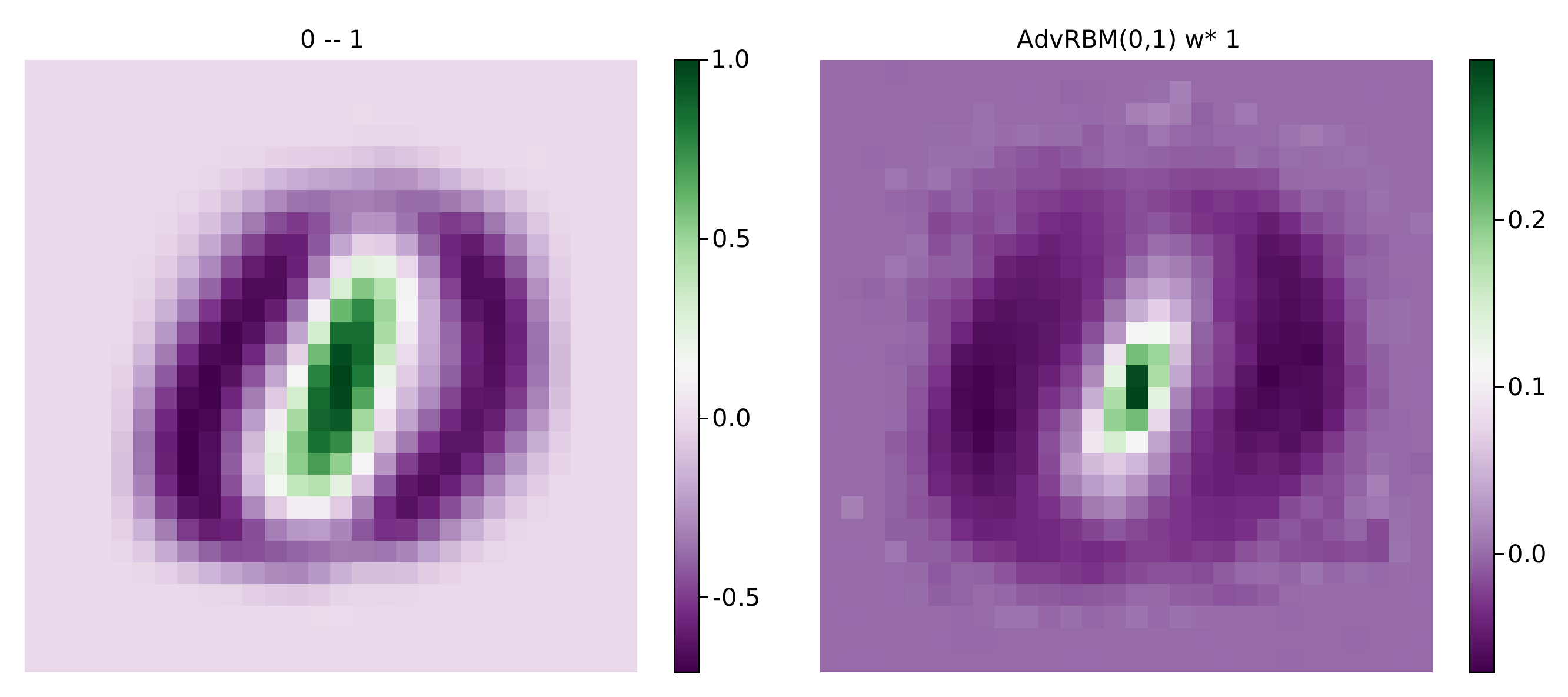}
\caption{\label{SI:fig:q01}\textbf{Left:} Vector $\mathbf q^{(1)}$ for MNIST0/1. \textbf{Right:} Weights of the released unit, $\mathbf w^\ast$, for the RBM trained with the linear constraint in the MNIST0/1 case.}
\end{figure}

\begin{figure}
\centering
\includegraphics[width=0.8\linewidth]{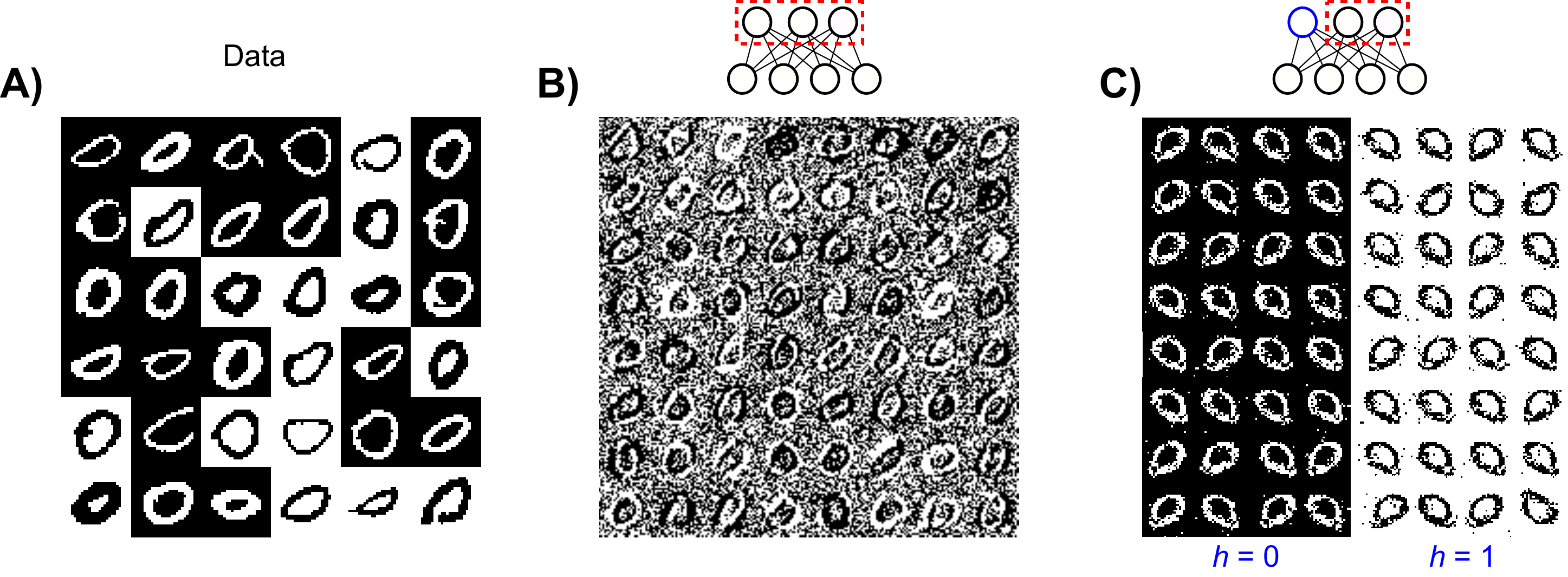}
\caption{\textbf{A)} Zero digits from MNIST, randomly presented as a white digit on black background, or (flipping the pixel values) as a black digit on white background. Labels are set to $u=0$ or $u=1$ according to whether the background is white or black, respectively. \textbf{B)} Data generated by an RBM trained on this data, with the first-order constraint acting on all hidden units. The samples are such that their original background and color of the digit are difficult to elucidate. The digit is still recognizable, through the correlations of pixels in the strokes of the zero shape. \textbf{C)} Samples generated by the partially constrained RBM, where the released hidden unit is conditioned to one of the two binary states on each side of the panel.} \label{SI:fig:mnist_black_white}
\end{figure}

\begin{figure}
\centering
\includegraphics[width=\linewidth]{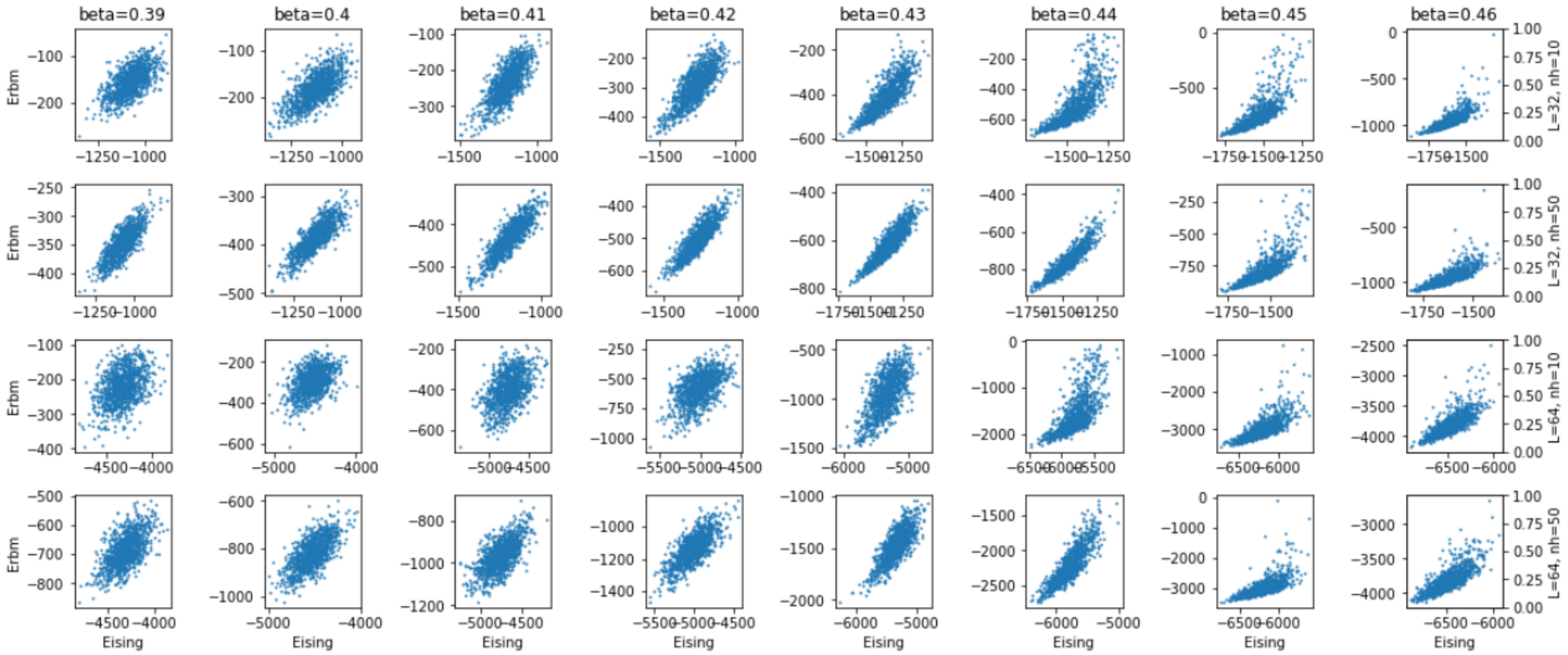}
\caption{\label{SI:fig:ising_energy_rbm}Ising model energy vs. energy of the trained RBM, for different temperatures, grid sizes, and numbers of hidden units.}
\end{figure}

\begin{figure}
\centering
\includegraphics[width=\linewidth]{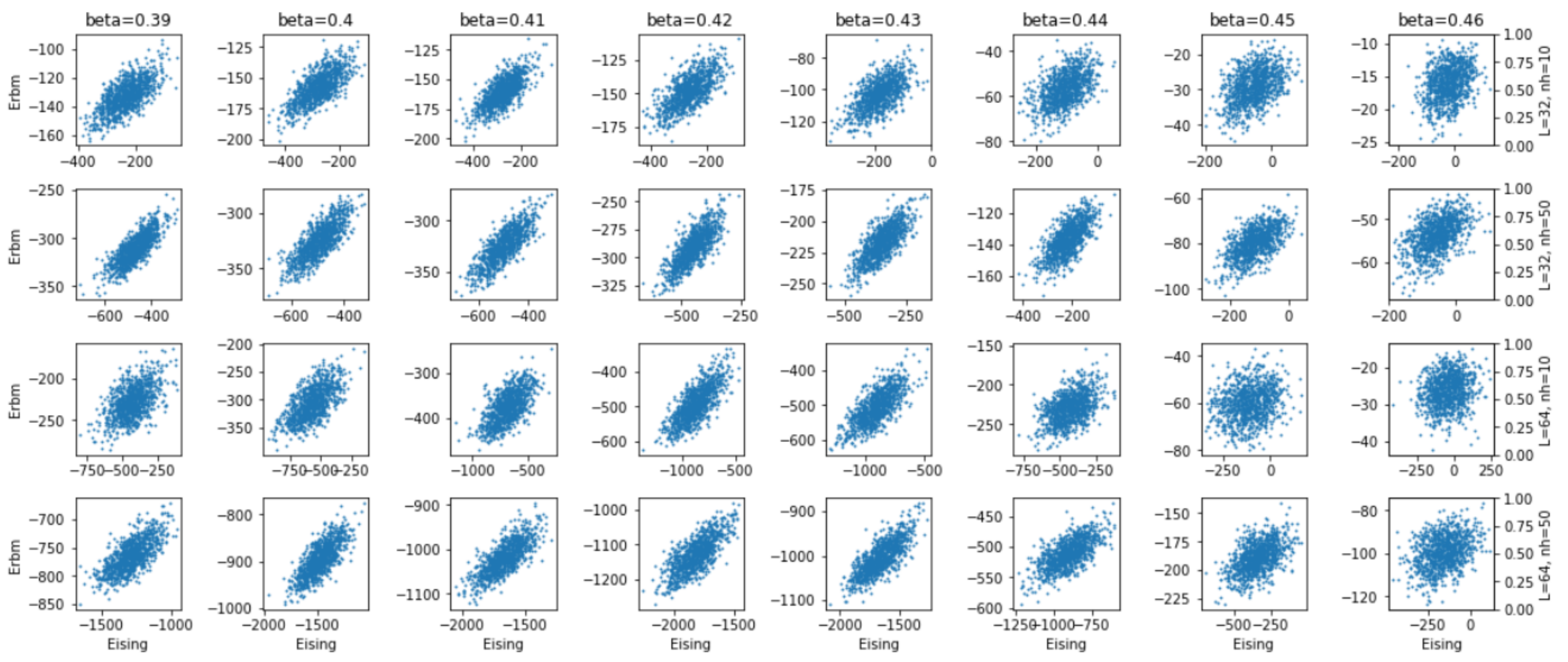}
\caption{\label{SI:fig:ising_energy_adv}Ising model energy vs. energy of the trained RBM under the linear constraint acting on all hidden units, for different temperatures, grid sizes, and numbers of hidden units.}
\end{figure}

\begin{figure}
\centering
\includegraphics[width=\linewidth]{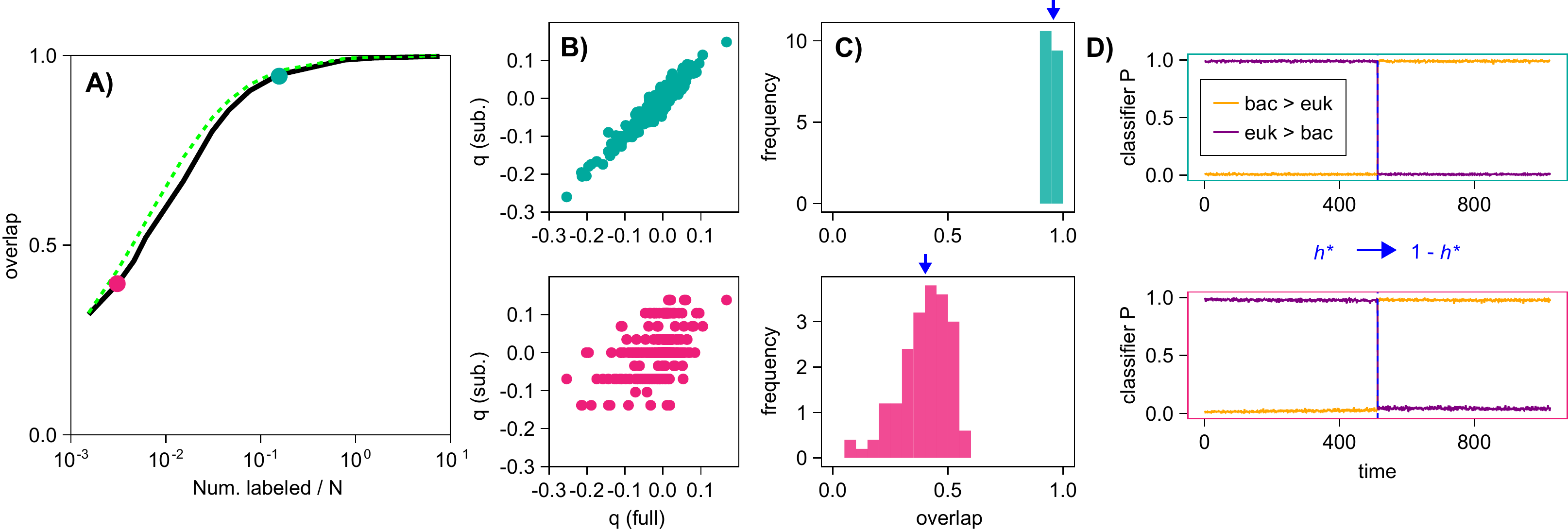}
\caption{\textbf{Sub-sampling labeled data for KH-domain protein sequences dataset}. \emph{Cf.} Figure \ref{fig:subsample} in the main-text for the MNIST0/1 counterpart of this figure.
\textbf{A)} Overlap \eqref{eq:overlap} between $\mathbf{q}^{(1)}_{sub}$ (computed on a sub-sampled labeled dataset) and $\mathbf{q}^{(1)}_{full}$ (computed on the full dataset), plotted as a function of the number of labeled examples the in sub-sampled dataset divided by the dimension. An average over 100 random sub-samples is taken. The black solid curve shows the empirical result, while the dashed green curve is the theoretical estimate \eqref{eq:overlap-bound}. \textbf{B)} For the pink and cyan dots of A), we plot an example of the obtained vectors $\mathbf{q}^{(1)}_{sub}$ in comparison to $\mathbf{q}^{(1)}_{full}$. \textbf{C)} Histogram of overlaps over the 100 realizations of the sub-sampled data, at the conditions of the cyan and pink dots of panel A). \textbf{D)} Label manipulation, using the sub-sampled $\mathbf{q}^{(1)}_{sub}$ in the two cases.}
\label{SI:fig:subsample_pfam}
\end{figure}

\end{document}